%% file: paper.tex
\documentclass[10pt]{article}

\usepackage{amsfonts,amsmath,amsthm}
\usepackage{algorithm,algorithmic}
\usepackage{bm,color}
\usepackage[dvipdfmx]{graphicx}
\usepackage[margin=1in]{geometry}

\usepackage{mcr}
\usepackage{pkg}
\usepackage{thmE}
\usepackage{algE}
\usepackage{multirow}
\usepackage{arydshln}
\title{Safe Feature Pruning for Sparse High-Order Interaction Models}

\date{June 20, 2015}

\author{
Kazuya Nakagawa \\
Nagoya Institute of Technology \\
Nagoya, Japan \\
\texttt{nakagawa.k.mllab.nit@gmail.com} \\
\and 
Shinya Suzumura \\
Nagoya Institute of Technology \\
Nagoya, Japan \\
\texttt{suzumura.mllab.nit@gmail.com} \\
\and 
Masayuki Karasuyama\\
Nagoya Institute of Technology \\
Nagoya, Japan \\
\texttt{karasuyama@nitech.ac.jp} \\
\and 
Koji Tsuda\\
University of Tokyo \\
Tokyo, Japan \\
\texttt{tsuda@k.u-tokyo.ac.jp} \\
\and 
Ichiro Takeuchi\thanks{Corresponding author} \\
Nagoya Institute of Technology \\
Nagoya, Japan \\
\texttt{takeuchi.ichiro@nitech.ac.jp} \\
}

\begin{document}

\maketitle

\clearpage
\begin{abstract}
\input{abst}

 \vspace{.1in}
\vspace{.1in}
\end{abstract}

\clearpage

\input{sec1}

\input{sec2}

\input{sec3}
\input{sec4}

\input{sec5}

\appendix

\input{appA}
\input{appB}

\end{document}

%% file: abst.tex
Taking into account high-order interactions among covariates
is valuable 
in many practical regression problems.
This is,
however, 
computationally challenging task 
because the number of high-order interaction features to be considered would be extremely large 
unless the number of covariates is sufficiently small. 
In this paper,
we propose a novel efficient algorithm
for
LASSO-based sparse learning of such high-order interaction models. 
Our basic strategy for reducing the number of features 
is to employ the idea of 
recently proposed \emph{safe feature screening (SFS) rule}.
An SFS rule has a property that, 
if a feature satisfies the rule, 
then the feature is guaranteed to be non-active in the LASSO solution,
meaning that it can be \emph{safely} screened-out prior to the LASSO training process. 
If a large number of features can be screened-out
before training the LASSO,
the computational cost and the memory requirment can be dramatically reduced. 
However,
applying such an SFS rule 
to
each of the extremely large number of high-order interaction features 
would be computationally infeasible.
Our key idea for solving this computational issue is to 
exploit the underlying tree structure among
high-order interaction features.
Specifically,
we introduce a pruning condition 
called 
\emph{safe feature pruning (SFP) rule}
which has a property that, 
if the rule is satisfied in a certain node of the tree,
then all the high-order interaction features
corresponding to its descendant nodes
can be guaranteed to be non-active at the optimal solution.
Our algorithm is extremely efficient,
making it possible to work,
e.g., 
with $3^{\rm rd}$ order interactions of 10,000 original covariates,
where the number of possible high-order interaction features is
greater than $10^{12}$.

%% file: sec1.tex
\section{Introduction}
\label{sec:intro}

Sparse learning of high-dimensional models has been actively studied in the past decades
\cite{buhlmann11a}.
Among many approaches,
LASSO \cite{Tibshirani96a} is one of the most widely used methods,
and its 
statistical and computational properties have been intensively investigated.
The main task in LASSO training is to identify the set of
\emph{active} features
whose coefficients turn out to be nonzero at the optimal solution. 
In case we know which features would be active, 
the solution trained only with those active features is guaranteed to be optimal. 
This observation suggests so-called \emph{feature screening} approaches,
where
we first screen-out a subset of features which would be non-active at the optimal solution,
and then
train a LASSO
only with the remaining features. 
The LASSO training can be highly efficient
if a majority of non-active features could be screened out a priori.

Existing screening approaches are categorized into two types.
In the first type of approaches called 
\emph{non-safe feature screening},
a subset of features
which are 
\emph{predicted}
to be non-active
at the optimal solution are first identified, 
and then 
a LASSO is trained 
after screening out those features.
Since the predictions might be incorrect, 
the obtained LASSO solution is used for checking 
if all the screened-out features are really non-active.
Unless all of them are confirmed to be truly non-active, 
some of those features must be brought back into the working feature set, 
and a LASSO is trained again 
with the updated working feature set. 
In non-safe screening approaches,
such a trial-and-error process must be repeated 
until all the optimality conditions are satisfied.

Another type of approaches 
called
\emph{safe feature screening (SFS)} 
was recently introduced by El Ghaoui et al. \cite{ElGhaoui2012}.
The advantage of SFS is that 
the screened-out features are guaranteed to be non-active
at the optimal solution,
meaning that 
the iterative trial-and-error process is not necessary. 
The safe feature screening approach has been receiving an increasing attention in the literature,
and several extensions have been recently studied
\cite{ElGhaoui2012,Xiang2011,Liu2014,Wang2014}.
SFS is especially useful
when the number of features is extremely large
and the entire data set cannot be stored in the memory. 
Once a subset of features are screened-out by SFS, 
those features can be completely removed from the memory 
because they would never be accessed during the following LASSO training process.

In this paper,
we study sparse learning problems for high-order interaction models.
Let us denote the original training set by 
$\{(\bm z^i, y^i)\}_{i \in [n]}$,
where
$n$ is the number of training instances,
$\bm z^i := [z^i_1 \ldots z^i_d]^\top$
is the $d$-dimensional original covariates,
and
$y^i$
is the scalar response. 
In high-order interaction models up to order $r$,
we have 
$D = \sum_{\kappa \in [r]} {d \choose \kappa}$
features.
Thus,
the expanded 
$n \times D$
design matrix
$\bm X$ 
has the form: 
\begin{align}
 \label{eq:X}
 \!\!
 \nonumber
 \text{\scriptsize(main effect)}
 ~~~~~
 \text{\scriptsize($2^{\rm nd}$ order interactions)}
 ~~~~~ ~~~~
 \cdots
 ~~~~~ ~~~~
 \text{\scriptsize($r^{\rm th}$ order interactions)}
 ~~~~~ ~~~~~ ~~~~~ ~~~~~
\\
 \bm X := \mtx{ccc:ccc:c:ccc}{
 z^1_{1} & \ldots & z^1_{d} & z^1_{1} z^1_{2} & \ldots & z^1_{d-1} z^1_{d} & \ldots & z^1_{1} z^1_{2} \!\cdots\! z^1_{r} & \ldots & z^1_{d-r+1} z^1_{d-r+2} \!\cdots\! z^1_{ d} \\
 \vdots & \ddots &\vdots & \vdots & \ddots & \vdots & \ddots & \vdots & \ddots & \vdots \\
 z^n_{1} & \ldots & z^n_{d} & z^n_{1} z^n_{2} & \ldots & z^n_{d-1} z^n_{d} & \ldots & z^n_{1} z^n_{2} \!\cdots\! z^n_{r} & \ldots & z^n_{d-r+1} z^n_{d-r+2} \!\cdots\! z^n_{ d} \\
 }.
~ ~~~~~
\end{align}
Then,
we consider LASSO problem
\begin{align}
\label{eq:lasso}
 \bm \beta^* :=
 \arg \min_{\bm \beta \in \RR^D}
 ~
 \frac{1}{2}
 \|\bm y - \bm X \bm \beta\|_2^2
 +
 \lambda
 \|\bm \beta\|_1, 
\end{align}
where
$\bm y := [y^1 \ldots y^n]^\top \in \RR^n$ 
and 
$\lambda > 0$
is the regularization parameter which makes a balance between the first loss term and the second regularization term. 
Unless the original input dimension
$d$
is fairly small,
the number of features
$D$
would be extremely large.
For example,
when
$d = 10,000$ and $r=3$,
we have
$D > 10^{12}$
features.
Although a variety of LASSO training algorithms have been proposed, 
it is still computationally infeasible to solve such a high-dimensional LASSO problem. 
Furthermore, 
it would also be difficult to load the entire (expanded) data set in the memory, 
making it hard to use existing LASSO solvers.

\begin{figure}[t]
 \begin{center}
  \includegraphics[scale=0.4]{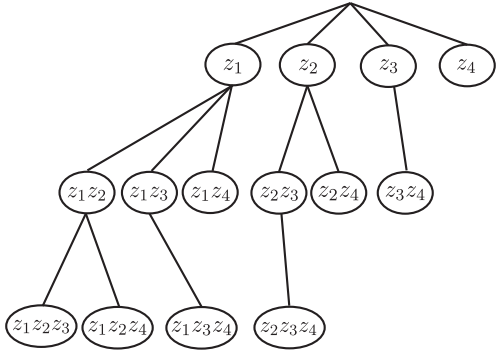}
   \caption{The underlying tree structure among high-order interaction features ($d = 4$ and $r = 3$).}
\label{fig:tree}
 \end{center}
\end{figure}

Our basic strategy is to develop 
an SFS method 
particularly suited for high-order interaction models,
and screen-out majority of non-active features 
before actually training the model by the LASSO. 
Unfortunately,
any existing SFS methods cannot be used for our problem 
because 
it is impossible to evaluate each SFS rule for each of the extremely large $D$ features. 
Our key idea
for overcoming this computational difficulty 
is to exploit 
the underlying tree structure among high-order interaction features 
as depicted in
\figurename~\ref{fig:tree}.
We propose,
what we call,
\emph{safe feature pruning (SFP)},
a novel safe feature screening method for high-order interaction models. 
Our SFP rule is a condition 
defined in each node of the tree such that, 
if the condition is satisfied in a certain node, 
then all the high-order interaction features
corresponding to its descendant nodes are guaranteed to be non-active 
at the optimal solution. 
It allows us to safely screen-out
a large number of high-order interaction features. 

%% file: sec2.tex
\section{LASSO for high-order interaction models}
\label{sec:LASSO_with_Combinatorial_Interactions}
\subsection{Problem setup}
In this paper
we study sparse learning of high-order interaction models. 
Throughout the paper,
we assume that 
$z^i_j, (i, j) \in [n] \times [d]$
is standardized to 
$[0, 1]$.
As mentioned in the previous section,
the training set in the original covariate domain is denoted as 
$\{(\bm z^i, y^i)\}_{i \in [n]}$
where
$\bm z^i \in [0, 1]^d$
and
$y^i \in \RR$. 
In addition,
we also denote the training set in the \emph{expanded} feature domain
as
$\{(\bm x^i, y^i)\}_{i \in [n]}$,
where
$\bm x^i \in [0, 1]^D$
is the expanded feature vector of the $i^{\rm th}$ training instance,
i.e.,
the $i^{\rm th}$ row of the design matrix
$\bm X$
in \eq{eq:X}. 
Note that high-order interaction features
$x^i_j, (i, j) \in [n] \times [D]$ 
is also defined in
$[0, 1]$.
Furthermore,
we denote
the
$j^{\rm th}$
column of the wide design matrix
$\bm X$
as 
$\bm x_j, j \in [D]$. 
The problem we consider here is to find a sparse solution
$\bm \beta^* = [\beta^*_1, \ldots, \beta^*_D]^\top$
by solving the LASSO problem
in \eq{eq:lasso}.

\subsection{Preliminaries and basic idea}
When the penalty parameter
$\lambda$
is sufficiently large,
only a small portion of the $D$ coefficients
$\{\beta^{*}_j\}_{j \in D}$
would be non-zero.
We denote the index set of the active features as 
\begin{align*}
 \cA^* := \{j \in [D] ~:~ \beta^{*}_j \neq 0\}. 
\end{align*}
In convex optimization literature,
it is well-known
that
the optimal solution
does not depend on non-active variables,
which is formally stated as follows.
\begin{lemm}
 \label{lemm:active}
 Let
 $\cA$
 be an index set 
 such that 
 $\cA^* \subseteq \cA  \subseteq [D]$.
 Then,
 the solution of the LASSO problem
 \eq{eq:lasso}
 is given as 
 \begin{align}
  \label{eq:expanded-reduced-lasso}
   \bm \beta^*_{\cA}
   = 
  \arg \min_{\bm \beta \in \RR^{|\cA|}}
  \frac{1}{2}
  \| \bm y - \bm X_\cA \bm \beta \|_2^2
   +
   \lambda
   \| \bm \beta \|_1,
   ~~~
   \bm \beta^*_{\bar{\cA}}
   =
   \bm 0,
 \end{align}
 where
 $\bm \beta^*_{\cA}$
 and 
 $\bm \beta^*_{\bar{\cA}}$
 are the subvectors of
 $\bm \beta^*$
 with the components in
 $\cA$
 and
 $\bar{\cA} := [D] \setminus \cA$,
 respectively,
 and
 $\bm X_{\cA} \in [0, 1]^{n \times |\cA|}$
 is a submatrix of
 $\bm X$
 which only has columns indexed by
 $\cA$. 
\end{lemm}

Lemma~\ref{lemm:active}
indicates that, 
if we have an index set 
$\cA \supseteq \cA^*$, 
the optimal solution
$\bm \beta^{*}$
in
\eq{eq:lasso}
could be efficiently obtained
by solving a smaller optimization problem
that does not depend on 
all the $D$ features
but only on a subset of features in 
$|\cA|$.

\paragraph{Safe screening}
In order to find an index set
$\cA \supseteq \cA^*$,
we employ recently introduced technique called
\emph{safe feature screening (SFS)}
\cite{ElGhaoui2012}.
SFS enables us to find
a subset of non-active features 
without actually solving the optimization problem. 
Roughly speaking, 
SFS algorithm for LASSO problem
is based on its primal-dual relationship. 
The dual problem of the LASSO problem in 
\eq{eq:lasso}
is written as
\begin{align}
 \label{eq:lasso-dual}
 \min_{\{\theta_i\}_{i \in [n]}}
 ~
 \frac{1}{2}
 \sum_{i \in [n]}
 (\theta_i - \frac{1}{\lambda} y^i)^2
 ~~~
 \text{subject to}
 ~
 \left|
 \sum_{i \in [n]}
 x^i_j \theta_i
 \right|
 \le 1
 ~~~~~\forall~
 j \in [D],
\end{align}
where
$\{\theta_i\}_{i \in [n]}$
are the dual variables. 
Then,
using the standard convex optimization theory
(e.g., see \cite{Boyd04a}), 
we have the following lemma
(see \cite{ElGhaoui2012} for the proof). 
\begin{lemm}
 \label{lemm:safe-screening}
 Let
 $\{\theta^*_i\}_{i \in [n]}$
 be the optimal dual solutions of the LASSO dual problem in 
 \eq{eq:lasso-dual}.
 Then,
 \begin{align}
  \label{eq:safe-rule-a}
  \left|
  \sum_{i \in [n]} x^i_j \theta^*_i
  \right|
  < 1
  ~\Rightarrow~
  \beta^{*}_j = 0,
  ~~~
  j \in [D].
 \end{align}
\end{lemm}
The key idea of SFS is to efficiently compute
an upper bound
$U_j$
for
each 
$j \in [D]$
without actually solving the dual problem 
such that 
\begin{align}
 \label{eq:UB}
 U_j
 \ge
\left| \sum_{i \in [n]} x^i_j \theta^*_i \right|
\end{align}
Lemma~\ref{lemm:safe-screening}
indicates 
\begin{align*}
 U_j < 1
\Rightarrow
\beta^*_j = 0, 
\end{align*}
meaning that a non-active coefficient 
$\beta^*_j$
might be identified
before solving the optimization problem. 
After the seminal work of El Ghaoui et al~\cite{ElGhaoui2012},
several approaches for efficiently computing
such an upper bound
have been proposed \cite{Xiang2011,Liu2014,Wang2014,Ogawa2013}. 
In this paper,
we particularly use an idea of using 
\emph{variational inequality} 
for computing upper bounds,
which was recently proposed by Liu et al.\cite{Liu2014}. 

Safe screening has been used
when a sequence of LASSO solutions 
for various values of $\lambda$
are computed
(c.f., \emph{regularization path} \cite{Efron2004}). 
When we compute a sequence of solutions, 
we start from 
$\lambda_{\rm max}$ 
where a LASSO solution for any
$\lambda > \lambda_{\rm max}$
is $\bm \beta^* = \bm 0$. 
Then,
we compute a sequence of LASSO solutions
for a decreasing sequence of $\lambda$
by using the previous solution
as the initial warm-start solution. 
Upper bounds 
$\{U_j\}_{j \in [D]}$
in \eq{eq:UB}
can be constructed
by using the optimal solution of the LASSO
for another regularization parameter
$\tilde{\lambda} > \lambda$,
which should have been already obtained in the above context.

\paragraph{Tree structure and pruning rule}
Unfortunately,
SFS alone is not sufficient for our problem 
because
it is intractable to compute 
an upper bound
$U_j$
for each of the exponentially large number of features. 
For handling such extremely large number of features in the expanded feature domain,
we exploit the underlying tree structure. 
We consider a simple tree structure
as depicted in
\figurename~\ref{fig:tree}.
We denote 
each node of the tree
by an index
$j \in [D]$. 
For any node
$j$
in the tree, 
let 
$De(j)$
be a set of its descendant nodes. 
Our main contribution in this paper
is to develop 
a novel SFS method
particularly designed for fitting sparse high-order interaction models. 
Specifically,
in each node of the tree, 
we derive a condition
called 
\emph{safe feature pruning (SFP) rule}. 
Our SFP rule has the following nice property: 
\begin{align}
 \text{SFP rule for a node $j$ is satisfied}
 ~\Rightarrow~
 \left| \sum_{i \in [n]} x^i_{j^\prime} \theta^*_i \right| < 1
 \text{ for all $j^\prime \in De(j)$}.
\end{align}
This property indicates that,
if the SFP rule in a certain node $j$ of the tree
is satisfied, 
then
we can guarantee that all the high-order interaction terms
corresponding to its descendant nodes
can be safely screened-out.

\subsection{Related works}
Before presenting our main contribution,
let us briefly review related works in the literature. 
Fitting high-order interaction models 
has long been desired in many regression problems. 
In biomedical studies,
for example,
many complex diseases 
such as cancer
are known to be the consequences of 
high-order interaction effects of multiple genetic factors 
\cite{evans2006two, manolio2006genes, kooperberg2008increasing, cordell2009detecting}. 
In the past decade, 
several authors proposed extensions of the LASSO for incorporating interaction effects,
and studied their statistical properties 
\cite{Choi10a, hao2014interaction, Bien13a}. 
However, 
none of these works have sufficient computational mechanisms 
for handling exponentially large number of high-order interaction features.
Most of these works thus focus only on $2^{\rm nd}$ order interaction features
for moderate number of original covariates $d$. 
One commonly used heuristic
for reducing the number of interaction features
is to introduce 
so-called 
\emph{strong heredity assumption} 
\cite{Choi10a, hao2014interaction, Bien13a},
where, e.g.,
an interaction term
$z_1 z_2$ 
would be selected
only when both of
$z_1$
and
$z_2$
are selected. 
However,
such a heuristic assumption eliminates a chance to find out 
novel high-order interaction features
when there are no strong associations in their marginal main effects. 
To the best of our knowledge,
an only exceptional approach that can be applied to high-order interaction modeling 
for sufficiently large data sets 
is
\emph{itemset boosting (IB)} algorithm
presented in 
\cite{Saigo06a}.
IB algorithm is a boosting-type algorithm,
where a single feature is added and the model is updated in each step. 
In a nutshell,
IB algorithm manages its working feature set 
by exploiting the underlying tree structure
as we do. 
However, 
since the working feature set in IB algorithm is non-safe
(no guarantee to be active or non-active), 
aforementioned trial-and-error process is necessary. 
We compare our approach with IB algorithm in \S\ref{sec:exp},
and demonstrate that the former is computationally more efficient than the latter.
Another line of related studies is about 
the statistical issue 
such as
feature selection consistency
\cite{Wainwright09a,Meinshausen10a}
for high-order interaction models.
We have been studying how to apply
\emph{post-selection inference}
framework
recently introduced in
\cite{Lee14a}
to statistical inferences on high-order interaction models in \cite{anonymous15b}.

%% file: sec3.tex
\section{Safe feature pruning for high-order interaction models}
\label{sec:main}
In this section we present our main result.
The proposed safe feature pruning (SFP) 
is used 
when we compute a sequence of LASSO solutions
for a decreasing sequence of the regularization parameter 
$\lambda$.
Let
$\lambda_{\rm max} > \lambda_1 > \ldots > \lambda_T$
be a sequence of regularization parameter values
at each of which we want to compute the LASSO solution,
where
$\lambda_{\rm max} := \max_{j \in [D]} |\bm x_j^\top \bm y|$
\footnote{
The largest regularization parameter
$\lambda_{\rm max}$
can be efficiently computed
again by exploiting the tree structure
among features.
Specifically,
for any node
$j$
and
its arbitrary descendant node
$j^\prime$, 
by noting that 
$0 \le x^i_{j^\prime} \le x^i_j$
for any $i \in [n]$,
we have
$
|\bm x_{j^\prime}^\top \bm y|
\le
\max\{
|\sum_{i \in [n] : y_i > 0} x^i_{j^\prime} y_i|,
|\sum_{i \in [n] : y_i < 0} x^i_{j^\prime} y_i|\}
\le 
\max\{
|\sum_{i \in [n] : y_i > 0} x^i_j y_i|,
|\sum_{i \in [n] : y_i < 0} x^i_j y_i|
\}.
$
%
Using this relationship, 
we can efficiently find the maximum 
$|\bm x_j^\top \bm y|$
by searching over the tree
with pruning. 
}.
Furthermore,
we denote the optimal LASSO solutions for those sequence of regularization parameters as
$\bm \beta^*(\lambda_t) :=
[\beta^*_1(\lambda_t),
\ldots, 
\beta^*_D(\lambda_t)]^\top
\in \RR^D
$
for
$t \in [T]$. 

The outline of the algorithm for computing the sequence of solutions
is summarized in
Algorithm~\ref{alg:outline}.
In line 1,
we initialize
$\lambda_0$
and 
$\bm \beta^*(\lambda_0)$
by
$\lambda_{\rm max}$
and its corresponding solution 
$\bm \beta^*(\lambda_{\rm max}) = \zero$.
Line 3 is the core of our algorithm,
where we find a superset
$\cA(\lambda_t)$
of
$\cA^*(\lambda_t) := \{j \in [D] : \beta^*_j(\lambda_t) \neq 0\}$
by the proposed SFP method
(see Theorem~\ref{theo:main}). 
In line 4,
the LASSO problem is solved for obtaining
$\bm \beta^*(\lambda_t)$
by using any LASSO solver
only with a set of features in
$\cA(\lambda_t)$.
%
\begin{algorithm}[h]
 \caption{An outline for computing a sequence of LASSO solutions with safe feature pruning}
 \label{alg:outline}
 \begin{algorithmic}[1]
  \REQUIRE
  $\{(z^i, y^i)\}_{i \in [n]}$,
  $r$, 
  $\{\lambda_t\}_{t \in [T]}$
  \STATE
  $\lambda_0 \lA \max_{j \in [D]} |\bm x_j^\top \bm y|$
  and
  $\bm \beta^*(\lambda_0) \lA \zero$
  \FOR{$t = 1, \ldots, T$}
  \STATE
  Find 
  $\cA(\lambda_t) \supseteq \cA^*(\lambda_t)$
  by 
  {\bf safe feature pruning (SFP)} method
  with
  $(\lambda_{t-1}, \bm \beta^*(\lambda_{t-1}))$
  \STATE
  Compute the LASSO solution 
  $\bm \beta^*(\lambda_t)$
  by using only a subset of features in
  $\cA(\lambda_t)$
  \ENDFOR
  \ENSURE
  $\{\bm \beta^*(\lambda_t)\}_{t \in [T]}$
 \end{algorithmic}
\end{algorithm}

The following theorem is used in line 3 of the Algorithm~\ref{alg:outline}.
Given the optimal LASSO solution
$\bm \beta^*(\lambda_{t-1})$
for a regularization parameter
$\lambda_{t-1}$,
for
any
$\lambda_t \in (0, \lambda_{t-1})$, 
we can develop a SFP rule 
such that, 
if 
the rule is satisfied, 
then all the features corresponding to in its descendant nodes
$De(j)$
are guaranteed to be non-active at the optimal LASSO solution
for the regularization parameter
$\lambda_t$. 
\begin{theo}[{\bf safe feature pruning} with $(\lambda_{t-1}, \bm \beta^*(\lambda_{t-1}))$]
 \label{theo:main}
 Suppose that
 the optimal LASSO solution 
 $\bm \beta^*(\lambda_{t-1})$
 is available
 for a regularization parameter
 $\lambda_{t-1}$.
 In addition, 
 for any $\lambda_t \in (0, \lambda_{t-1})$, 
 define 
 \begin{align*}
  &
  P_1
  ~
  :=
  \frac{1}{2} \left(
  \|\bm{x}_j\|_2 \|\bm{b}\|_2 
  +
  \sum_{i : c_i > 0} c_i x^i_j 
  \right),
  ~~~
  P_2
  ~
  :=
  \frac{1}{2} \left(
  \|\bm{x}_j\|_2
  \|\bm{b} - \frac{\bm a^\top \bm b}{\|\bm a\|_2^2}\bm{a}\|_2 
  +
  \sum_{i : d_i > 0} d_i x^i_j
  \right),
  \\
  &
  M_1 :=
  \frac{1}{2} \left(
  \|\bm{x}_j\|_2 \|\bm{b}\|_2 
  -
  \sum_{i : c_i < 0} c_i x^i_j
  \right),
  ~~~
  M_2 :=
  \frac{1}{2} \left(
  \|\bm{x}_j\|_2
  \|\bm{b} - \frac{\bm a^\top \bm b}{\|\bm a\|_2^2}\bm{a}\|_2   
  -
  \sum_{i : d_i < 0} d_i x^i_j
  \right),
 \end{align*}
where 
 \begin{align*}
  &
  \bm a
  :=
  \frac{1}{\lambda_{t-1}}
  \sum_{j \in [D] : \beta^*_j(\lambda_{t-1}) \neq 0}
  \bm x_j \beta^*_j(\lambda_{t-1}),
  ~~~~
  \bm b
  :=
  \left(
  \frac{1}{\lambda_{t}} - \frac{1}{\lambda_{t-1}}
  \right)
  \bm y
  +
  \bm a,
  \\
  &
  \bm c
  :=
  \left(
  \frac{1}{\lambda_t}
  +
  \frac{1}{\lambda_{t-1}}
  \right)
  \bm y
  -
  \frac{\lambda_{t-1}}{\lambda_t}
  \bm a,
  ~
  ~~~~~ ~~~~~ ~~~~~ ~
  \bm d
  :=
  \bm{c} - \frac{\bm a^\top \bm b}{\|\bm{a}\|_2^2} \bm{a}.
 \end{align*}
 Then,
 for any node $j$ in the tree such that
 $\bm x_j \neq \zero$, 
 \begin{align}
  \label{eq:SFP-rule}
  \max \{U_j^+, U_j^-\} < 1\
  ~~~\Rightarrow~~~
  \beta^*_{j^\prime}(\lambda_t) = 0
  ~~~
  \text{ for all }
  j^\prime \in De(j),
 \end{align}
 where, 
 when
 $\lambda_{t-1} = \lambda_{\rm max} = \max_{j \in [D]}|\bm x_j^\top \bm y|$, 
 \begin{align*}
  U_j^+ := P_1, 
  ~~~
  U_j^- := M_1, 
 \end{align*}
 while, 
 when
 $\lambda_{t-1} < \lambda_{\rm max}$, 
 \begin{align*}
  U_j^+
  :=
  \max\{P_1, P_2\},
  ~~~
  U_j^-
  :=
  \max\{M_1, M_2\}.
 \end{align*}
 Furthermore,
 when
 the original covariates
 $z^i_j \in \{0, 1\}$
 for $i \in [n]$ and $j \in [d]$,
 then 
 \begin{align*}
  U_j^+
  &:=
  \mycase{
  P_2
  &
  ~~
  \text{if }
  - \sum_{i \in [n] : a_i < 0}
  a_i x^i_j
  <
  \frac{\bm a^\top \bm b}{\|\bm b\|_2^2},
  \\
  \max\{P_1, P_2\}
  &
  ~~
  \text{otherwise},
  }
  \\
  U_j^-
  &:=
  \mycase{
  M_2
  &
  \text{if }
  \phantom{-}
  \sum_{i \in [n] : a_i  > 0}
  a_i x^i_j
  <
  \frac{\bm a^\top \bm b}{\|\bm b\|_2^2},
  \\
  \max\{M_1, M_2\}
  &
  \text{otherwise}.
  }
 \end{align*}
\end{theo}
The proof of Theorem~\ref{theo:main} is presented
in supplementary appendix~\ref{app:proof}.
In the depth-first search in the tree,
if we encounter a node with
$\bm x_j = \zero$,
then we can, of-course, guarantee that its descendant nodes would be non-active. 
Note that the SFP condition
in
\eq{eq:SFP-rule}
can be computed 
by using the sparse previous solution
$\bm \beta^*(\lambda_{t-1})$ 
and a set of information
available in the node $j$. 
It means that,
if the SFP rule 
is satisfied
at a certain node $j$
in the tree,
we can stop searching over the tree and
all its descendant nodes can be screened-out. 

Our pruning approach relies on the fact that 
original covariates are defined in $[0, 1]$. 
For example, 
it is easy to note that 
\begin{align*}
 x^i_{j^\prime} \le x^i_{j}
 ~
 \forall
 ~
 (j^\prime, j) \in [D]^2
 ~
 \text{such that}
 ~
 j^\prime \in De(j).
\end{align*}
Such diminishing monotonicity properties on the high-order interaction features 
indicate that 
higher-order interaction features
which correspond to deep nodes in the tree
are more likely to be non-active
than those corresponding to shallow nodes. 
For example,
if the original covariates are defined in binary domain,
i.e.,
$x^i_j \in \{0,1\}$,
then the features would be more sparse 
as we consider higher-order interactions.
As we see in the following experiment section,
when the original covariates are sparse, 
our pruning approach works quite well.
%

For covariates defined in binary domain, 
where values 1 and 0 respectively indicate the existence and the non-existence of a certain property,
it is easy to interpret interaction effects
because
they simply indicate co-existence of multiple properties. 
On the other hand,
for continuous covariates, 
the interpretation of an interaction effect would depend on its coding. 
If each covariate is defined in $[0, 1]$ domain,
and the value represents the ``degree'' of an existence of a certain property,
then an interaction effect can be similarly understood as the ``degree'' of co-existence
of multiple properties. 

%% file: sec4.tex
\section{Experiments}
\label{sec:exp}
In this section,
we demonstrate the effectiveness of the proposed safe feature pruning (SFP) approach
through numerical experiments.

\subsection{Experimental setup}
In the experiments,
we computed a sequence of LASSO solutions
at a decreasing sequence of regularization parameters
$\lambda$.
Specifically,
we started from
$\lambda_0 = \lambda_{\rm max} := \max_{j \in [D]} |\bm x_j^\top \bm y|$,
and considered a sequence
$\lambda_t = (1 - 0.1/\sqrt{t}) \lambda_{t-1}$
for
$t = 1, 2, \ldots$
until
$\lambda_t / \lambda_{\rm max} < 0.01$. 
We considered interaction model up to $r = 3^{\rm rd}$ order.
As the LASSO solver,
we used shooting algorithm. 
All the codes were implemented by ourselves in C++,
and all the experiments were conducted 
by HP workstation Z800 (Xeon(R) CPU X5675 (3.07GHz), 48GB MEM). 

\subsection{Synthetic data experiments}
First,
we investigated the advantage of the proposed pruning scheme 
by comparing the computational costs with and without the
safe feature pruning (SFP)
on small synthetic data sets.
The synthetic data were generated from
\begin{align*}
 \bm y = \bm X \bm \beta + \bm \veps,
 \bm \veps \sim N(\bm 0, \sigma^2 \bm I),
\end{align*}
where
$\bm y \in \RR^n$
is the response vector,
$\bm X \in \RR^{n \times D}$
is the input design matrix,
$\bm \beta \in \RR^D$
is the coefficient vector,
and
$\bm \veps \in \RR^n$
is the Gaussian noise vector. 
Here,
we did not actually compute the extremely wide design matrix $\bm X$
because it has exponentially large number of columns. 
Instead, 
we generated a
random binary matrix 
$\bm Z \in \{0, 1\}^{n \times d}$
and each expanded high-order interaction feature 
$x^i_j$
was generated
from the $i^{\rm th}$ row of $\bm Z$
only when it was needed. 
For simplicity and computational efficiency,
we assumed that the covariates (hence interaction features as well) are binary,
and the sparsity $\eta \in [0, 1]$
(the fraction of 0s in the entries of $\bm Z$)
was changed among
$95\%, 90\%, 85\%, 80\%$ 
to see how sparsity can be exploited for efficient computation.
We set
$\bm \beta = \zero$,
$\sigma = 0.1$,
$n = 1,000$
and
$d = 1,000$. 
The total computation time in seconds and the average pruning rates 
are shown in 
Table~\ref{tab:jinkou}.
The results indicate that
SFP is fairly effective
for computational efficiency. 
Furthermore,
as the data set gets sparse,
the advantage of SFP increases. 
\begin{table}[t]
\begin{center}
\caption{Computation time in seconds for computing a sequence of LASSO solutions.}
 \vspace*{0.5mm}
 \begin{tabular}{rrrr}
\hline
sparsity(\%) & without SFP  & with SFP & pruning rate \\ \hline
95 & 12811.66 & 170.93 & 99.63\\
90 & 28816.94 & 417.07 & 99.42\\
85 & 52343.61 & 1248.41 & 98.47\\
80 & 80224.89 & 4460.59 & 95.63\\
\hline
 \end{tabular}
\label{tab:jinkou}
\end{center}
 \vspace*{-5mm}
\end{table}

\subsection{Benchmark data experiments}
Next,
we compared the proposed SFP approach
with
itemset boosting (IB)
algorithm presented in \cite{Saigo06a}. 
IB algorithm is a variant of working set method,
where
a set of working features 
and
a LASSO solution trained only with the working feature set 
are maintained in each step. 
IB algorithm updates the working set
by adding a feature
that most violates the current optimality condition.
The core of IB algorithm is that 
it can efficiently find the most violating feature
by exploiting
the underlying tree structure and anti-monotonicity property. 
Since there is no guarantee that all the features not in the working set are truly non-active in IB algorithm,
one must repeat 
trial-and-error process 
until all the optimality conditions are satisfied.

\begin{figure*}[t]
  \begin{center}
   \begin{tabular}{lll}
    \includegraphics[width=0.3\textwidth]{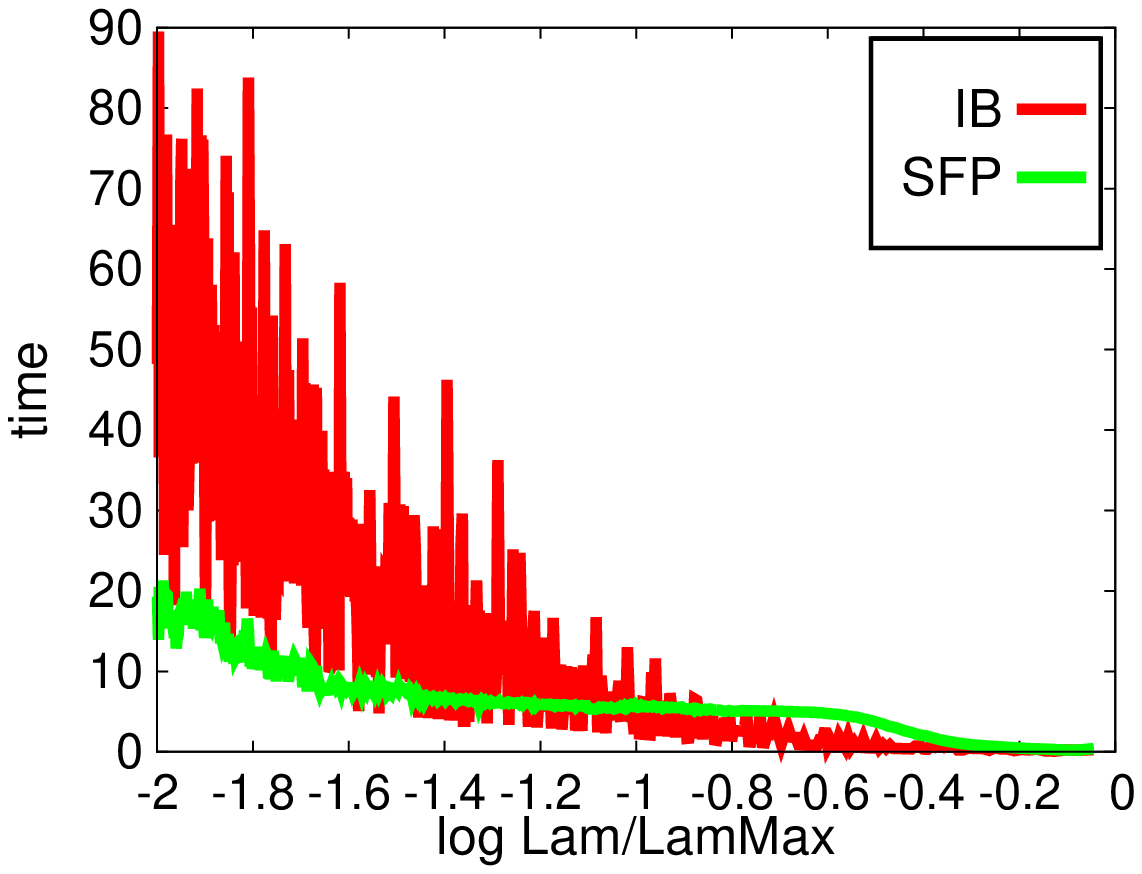}
    \hspace*{-0.5mm} & \hspace*{-0.5mm}
    \includegraphics[width=0.3\textwidth]{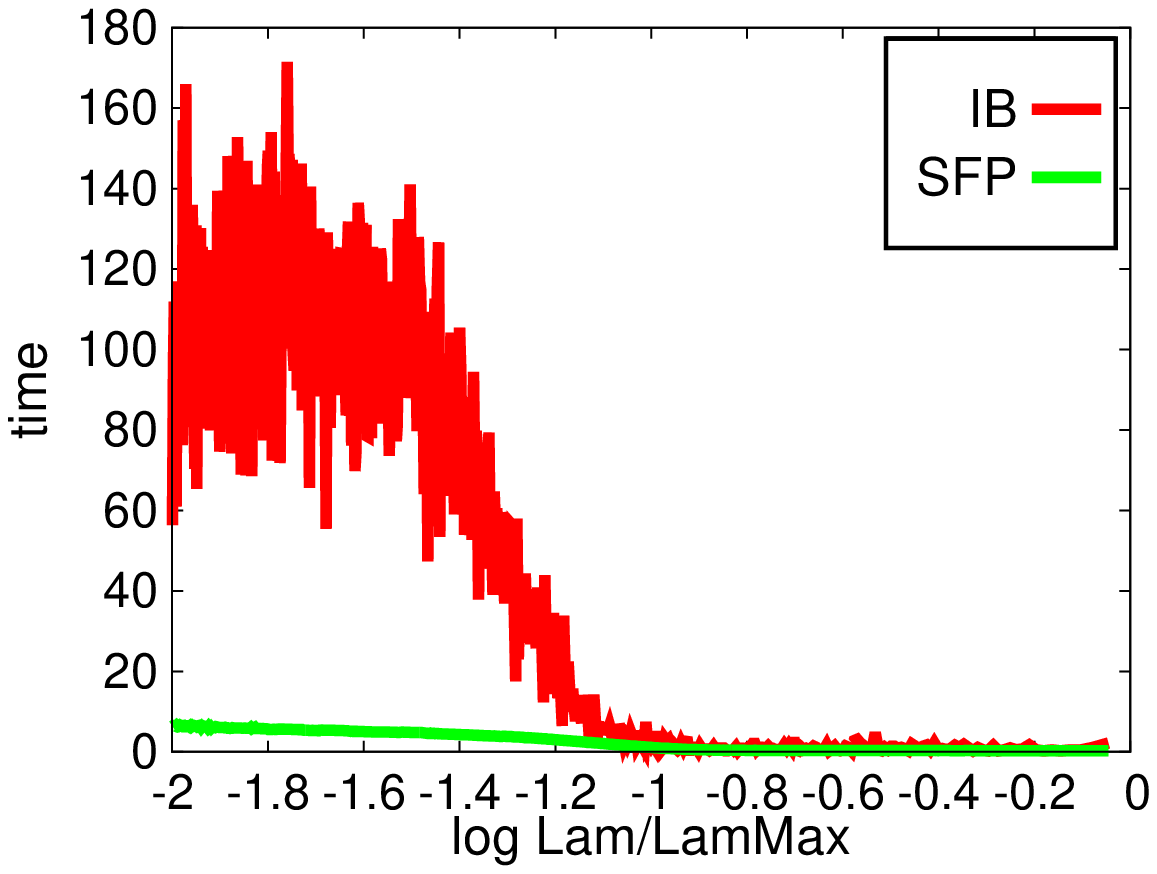}
    \hspace*{-0.5mm} & \hspace*{-0.5mm}
    \includegraphics[width=0.3\textwidth]{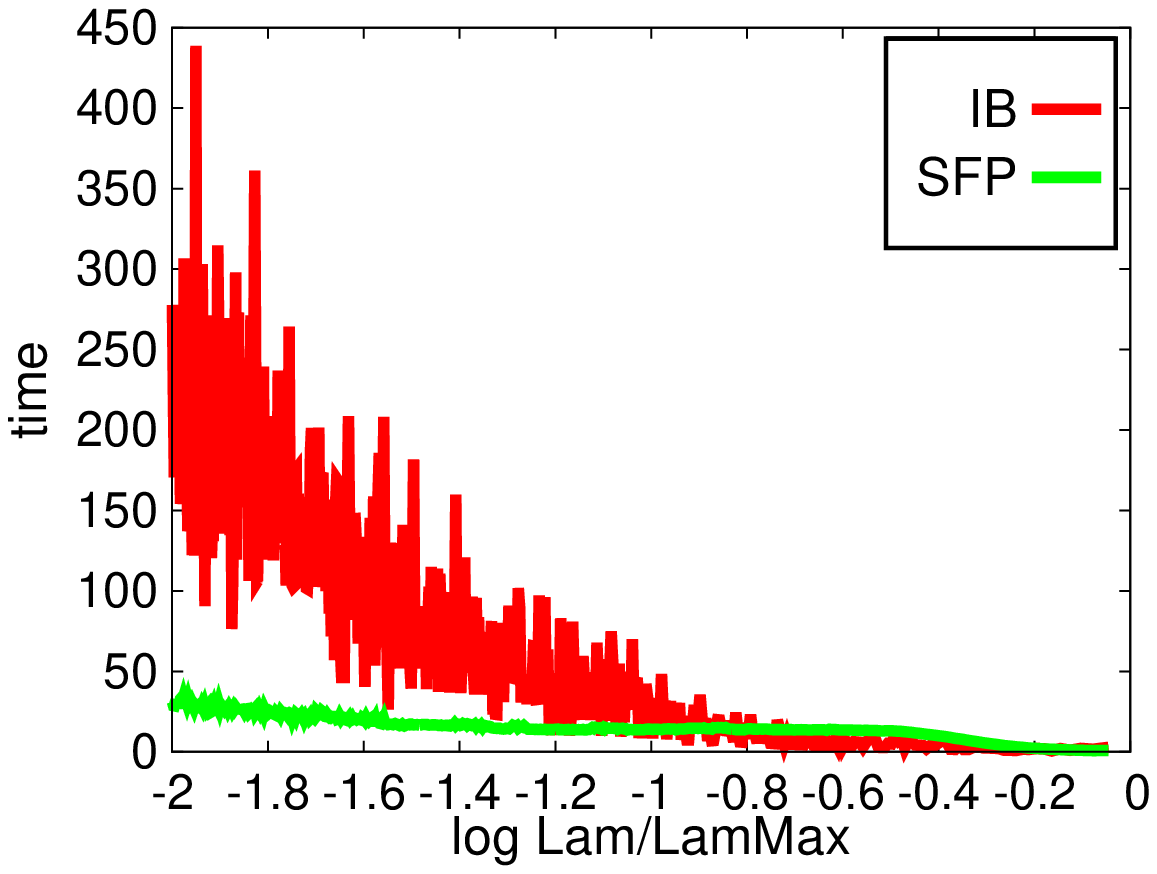} \\
    \multicolumn{3}{c}
    {\small (a) Computation time in seconds}\\
    \includegraphics[width=0.3\textwidth]{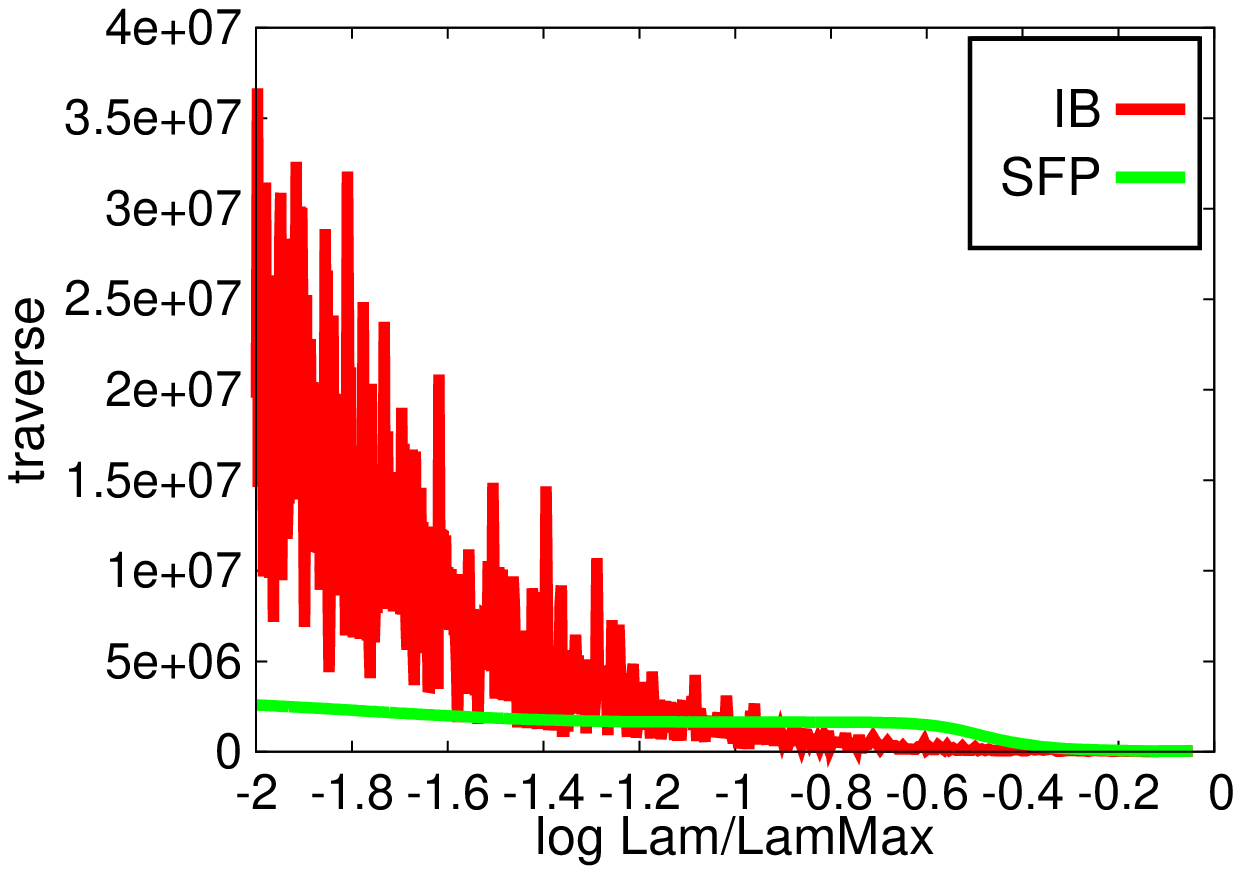}
    \hspace*{-0.5mm} & \hspace*{-0.5mm}
    \includegraphics[width=0.3\textwidth]{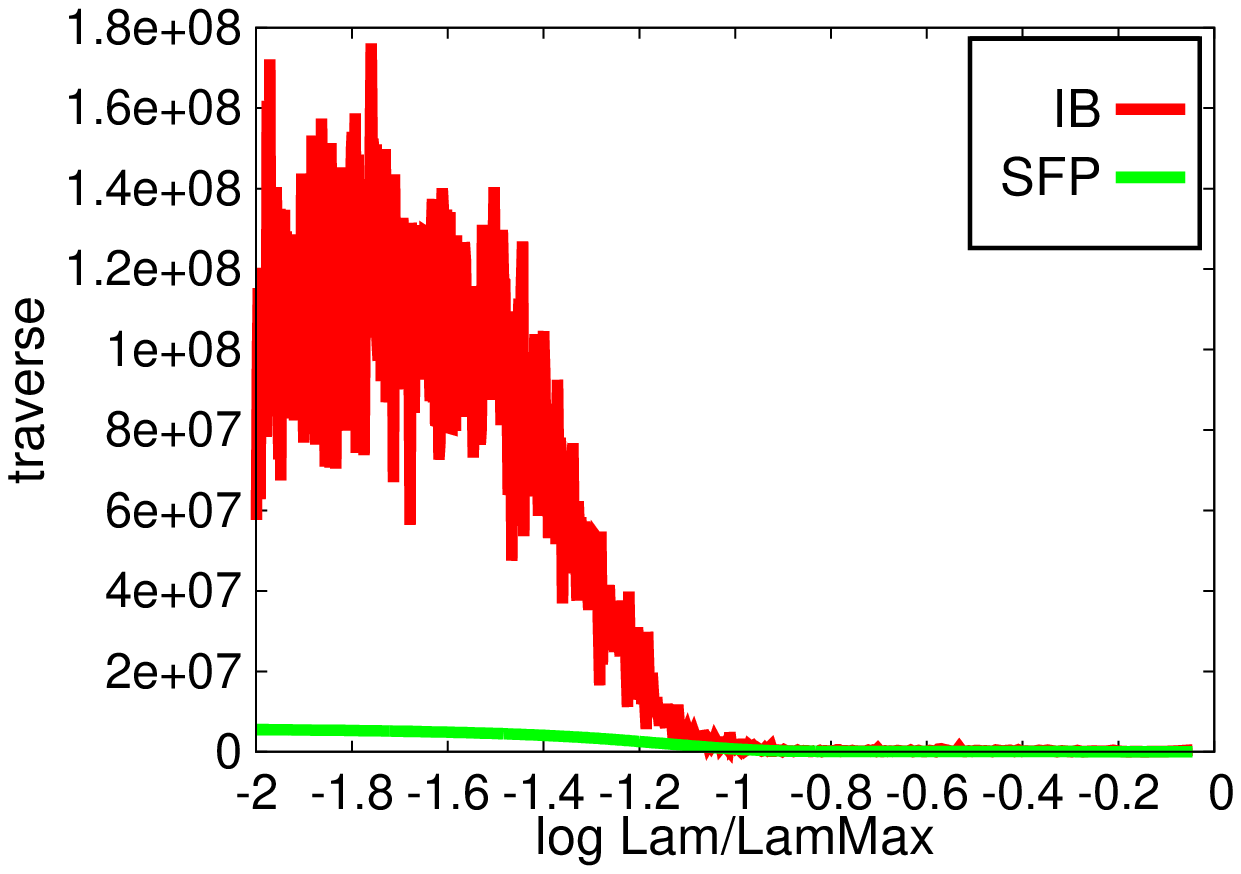}
    \hspace*{-0.5mm} & \hspace*{-0.5mm}
    \includegraphics[width=0.3\textwidth]{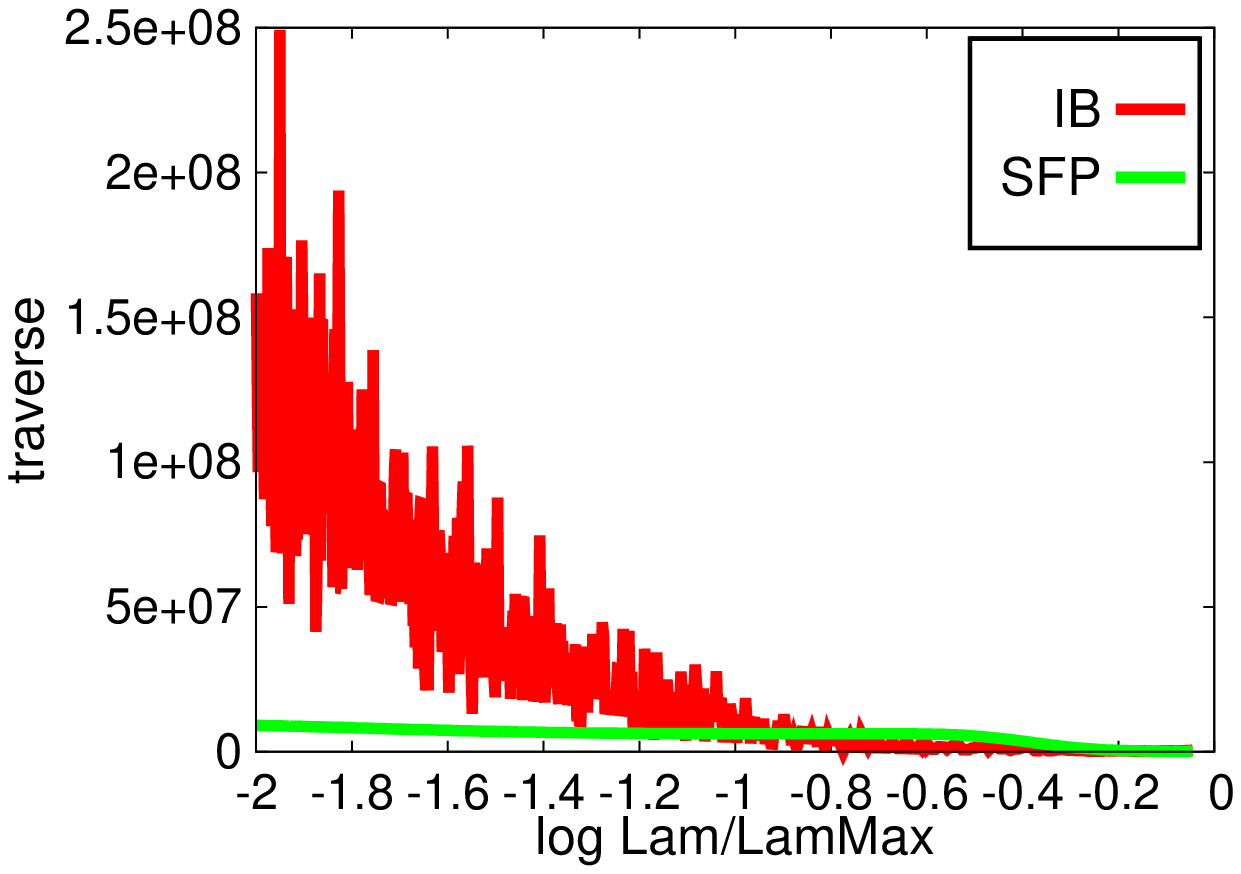} \\
    \multicolumn{3}{c}
    {\small (b) The number of traverse nodes}\\
    \includegraphics[width=0.3\textwidth]{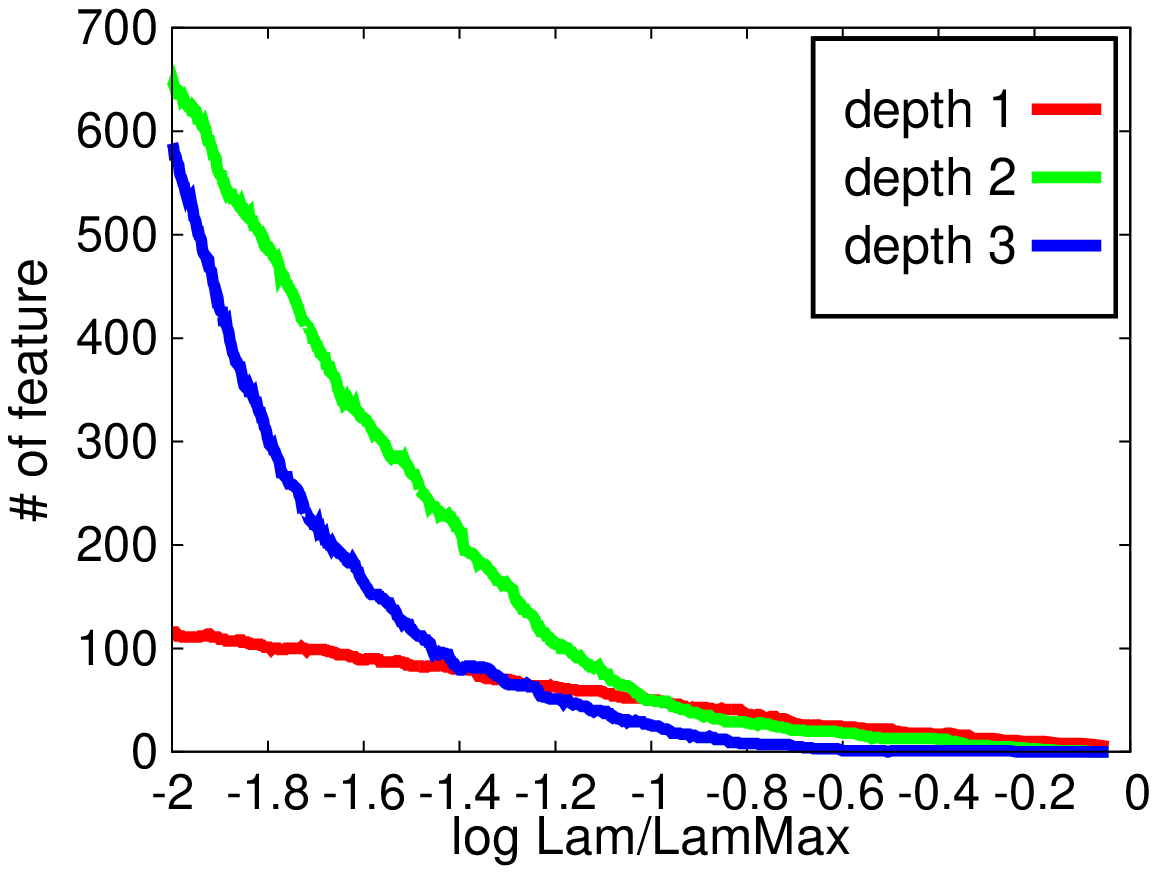}
    \hspace*{-0.5mm} & \hspace*{-0.5mm}
    \includegraphics[width=0.3\textwidth]{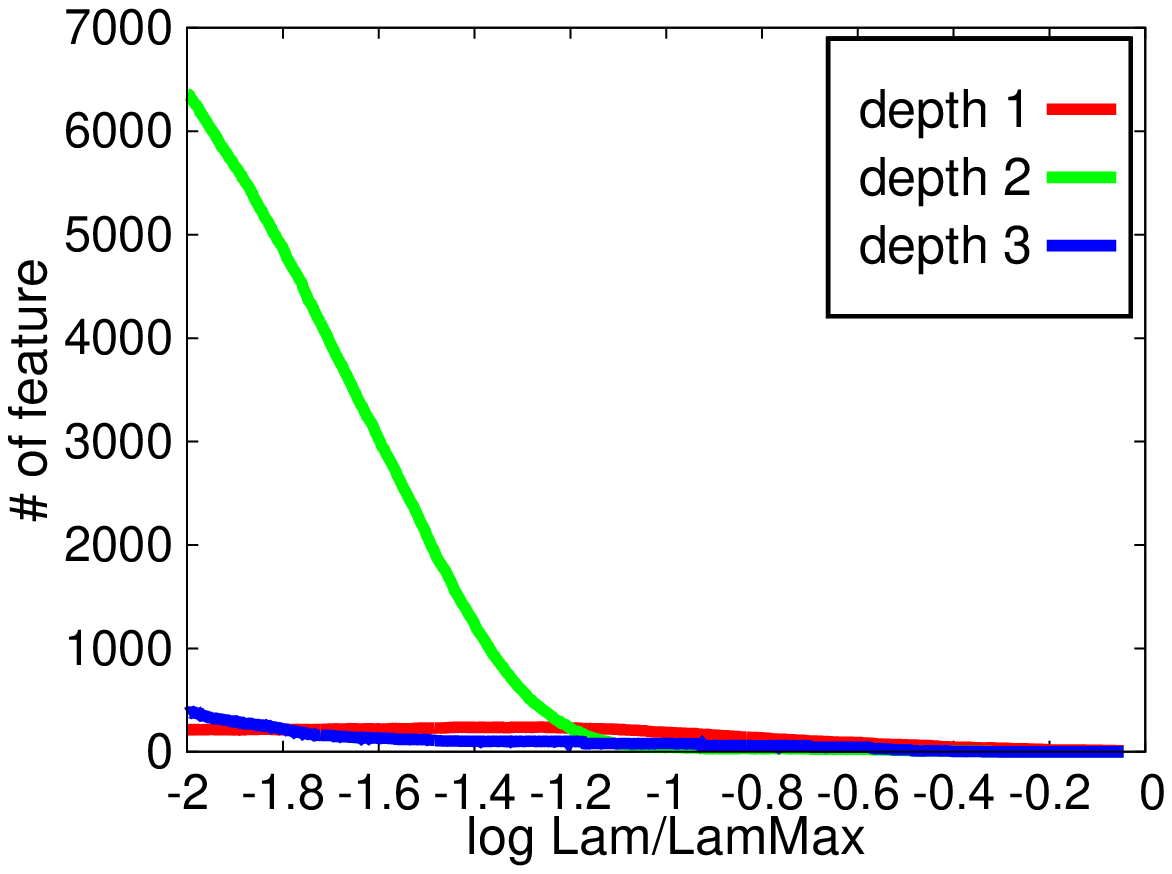}
    \hspace*{-0.5mm} & \hspace*{-0.5mm}
    \includegraphics[width=0.3\textwidth]{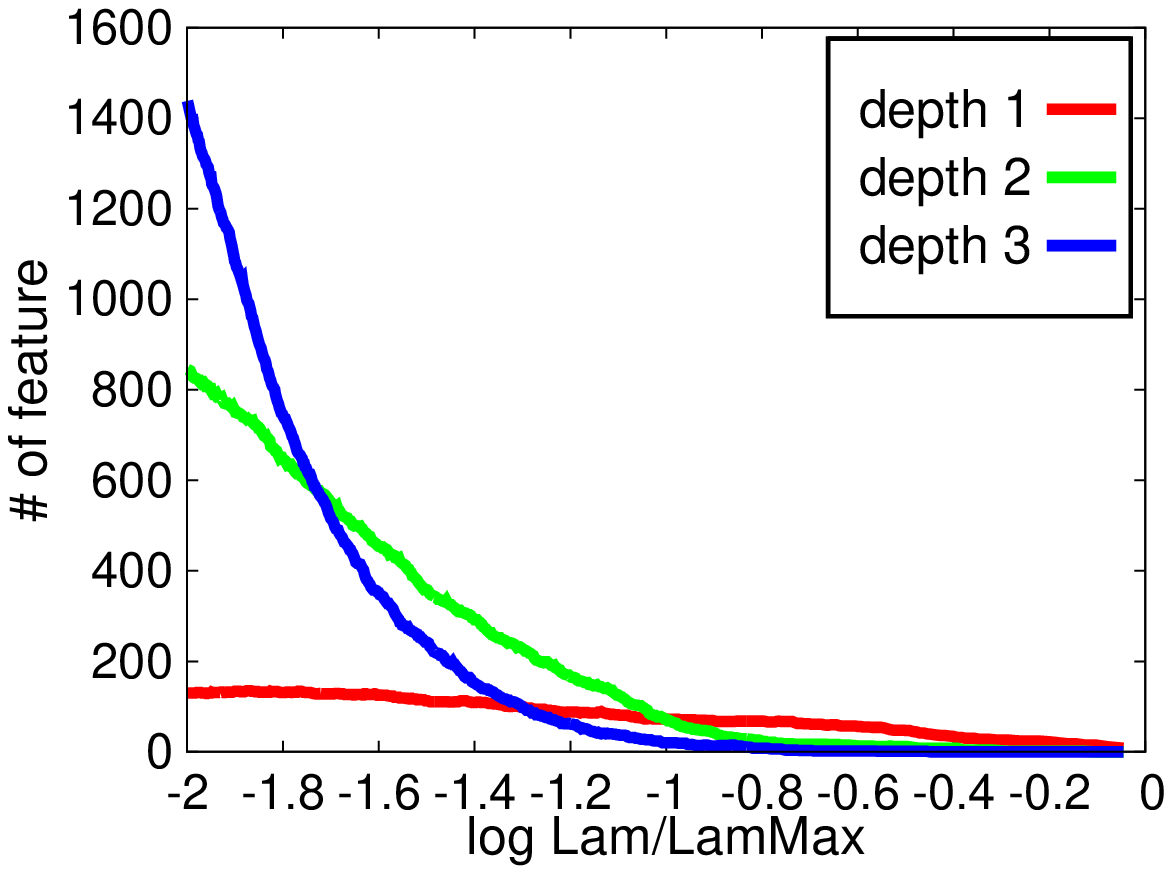} \\
    \multicolumn{3}{c}
    {\small (c) The number of active features}\\
    \multicolumn{3}{c}
    {\small main effect features (red), 2nd order interaction effect (green), and 3rd order interaction effect (blue) features}
   \end{tabular}
   \caption{Results on usps (left), protein (center), mnist (right) data sets.}
 \label{fig:benchmark}
  \end{center}
\end{figure*}

We used seven benchmark datasets in libsvm dataset repository
\cite{Chang2011a}
as listed in Table~\ref{tab:benchmark}.
In each dataset,
we restrict
the maximum numbers of instances $n$ and covariates $d$ to be 10,000.
%
Although some of these datasets are for binary classifications,
we regarded the response
$\{y_i\}_{i \in [n]}$
be real variables,
and standardized them so that they have the mean zero and the variance one. 
As we discussed in \S\ref{sec:main},
we only considered binary original covariates.
For a continuous covariate in the original data set, 
we first standardized it to have the mean zero and the variance one,
and then represented the covariate by two binary variables, 
each of which indicates whether
the value is greater than $\delta$ 
or
the value is smaller than $-\delta$,
where 
$\delta \in \{1.5, 2.0\}$.
The computation time in seconds are shown in
Table~\ref{tab:benchmark}.
We see in the table that 
the proposed SFP is faster than IB algorithm in almost all cases.
Figure 
\ref{fig:benchmark} 
shows the results on
usps,
protein
and
mnist
data sets (same figures for other datasets are presented in supplementary Appendix~\ref{app:results})
For each data set,
(a) computation time in seconds,
(b) the number of traverse nodes, 
and
(c) the number of active features
are plotted for each regularization parameter values.
We first see in (a) and (b) that the computation time is roughly proportional to the number of traverse nodes. 
Comparing SFP and IB algorithm in (a) and (b), 
the former is faster than the latter 
especially when
$\lambda$
is small. 
Furthermore,
the computational costs of IB algorithm in (a) and (b)
seems to be positively correlated with the number of active features in (c). 
A possible explanation of these observations is that, 
when $\lambda$ is small, 
IB algorithm must repeatedly search over the tree
for finding out which feature would be coming into the working set 
since there are many 
active features
that newly enters to the working set. 
On the other hand,
the computational cost of the proposed SFP did not increase as IB algorithm 
because SFP approach can screen-out large number of features at the same time. 
The plots in (c) suggests that,
when
$\lambda$
is small,
many high-order interaction features
(2nd and 3rd order interaction features are shown in green and blue, respectively)
become active,
indicating the potential advantage of considering high-order interaction features. 

\begin{table}[t]
\begin{center}
 \caption{Total computation time for computing a sequence of LASSO solutions in high order interaction models (seconds) for various benchmark datasets.}
 \vspace*{0.5mm}
 \label{tab:benchmark}
\begin{tabular}{rrr|rr|rr}
\hline
\multirow{2}{*}{Data}  & \multirow{2}{*}{$n$} & \multirow{2}{*}{$d$} 
&\multicolumn{2}{c|}{$\delta$=1.5} & \multicolumn{2}{c}{$\delta$=2.0} \\
\cline{4-7}
& & & IB & SFP & IB & SFP \\ \hline \hline
usps & 7,291 & 256 & 10514.81 & 4506.82 & 1373.21 & 1907.21 \\
madelon & 2,000 & 500 & 6685.99 & 1865.42 & 982.69 & 358.08 \\
protein & 10,000 & 357 & 33320.01 & 1975.00 & 20593.55 & 1297.30 \\
mnist & 10,000 & 780 & 47985.89 & 10047.49 & 7418.59 & 4009.85 \\
rcv1\_binary & 10,000 & 10,000 & 1707.81 & 734.60 & 2893.67 & 826.56 \\
real-sim & 10,000 & 10,000 & 11122.02 & 3583.31 & 15101.93 & 3473.46 \\
news20 & 10,000 & 10,000 & $>100K$ & 34988.51 & $>100K$ & 28568.94 \\ \hline
\end{tabular}
\end{center}
\end{table}
 \vspace*{0.5mm}

%% file: sec5.tex
\section{Conclusion}
We proposed a safe feature screening rules
called
\emph{safe feature pruning (SFP)}
for
high-order interaction models.
A key advantage of SFP is that, 
by exploiting the underlying tree structure among high-order interaction features and its anti-monotonicity property, 
a large number of high-order interaction features can be simultaneously 
screened out
by evaluating a few simple conditions on the tree. 
As long as the original covariates are sufficiently sparse,
our algorithm can be used for
high-order interaction models
with the number of features 
$D > 10^{12}$. 
Our next future work would be to extend this idea to classification setups.

%% file: appA.tex
\section{Proof of Theorem~\ref{theo:main}}
\label{app:proof}
In this section we prove Theorem~\ref{theo:main}.
To prove the theorem, 
we use a recent results on safe feature screening 
developed in \cite{Liu2014}. 
Our technical contribution in the following proof
is
in bounding the screening condition of a feature in a node
based only on the information available
in its ancestor node,
which is the crucial property of the proposed safe feature pruning. 

Although the notations and formulations are different, 
the following proposition is essentially identical with
the recent result in \cite{Liu2014}.
\begin{prop}[Liu et al.\cite{Liu2014}]
 \label{prop:liu}
 Consider a pair of regularization parameters
 $\lambda_1$
 and
 $\lambda_2$
 for the LASSO problem in \eq{eq:lasso}
 such that 
 $\max_{j \in [D]} |\bm x_j^\top \bm y| \ge \lambda_1 > \lambda_2 > 0$,
 and denote 
 the optimal dual variable vectors
 for these two problems 
 as 
 $\bm \theta_1^*$
 and 
 $\bm \theta_2^*$,
 respectively.
 In addition, 
 define
 \begin{align*}
  \bm a := \frac{\bm y}{\lambda_1} - \bm \theta^*_1,
  ~
  \bm b := \frac{\bm y}{\lambda_2} - \bm \theta^*_1,
  ~
  \bm c := \frac{\bm y}{\lambda_2} + \bm \theta^*_1.
 \end{align*}
 Furthermore, 
 for each
 $j \in [D]$, 
 define
 \begin{align}
  \label{eq:prf-x}
  u_j^+
  &:=
  \mycase{
  \frac{1}{2}
  (
  \bm x_j^\top \bm c + \|\bm x_j\|_2 \|\bm b\|_2
  )
  &
  \text{ if }
  \frac{
  \bm b^\top \bm a
  }{
  \|\bm b\|_2
  }
  \le
  -
  \frac{
  \bm x_j^\top \bm a
  }{
  \|\bm x_j\|_2
  }, 
  \\
  \frac{1}{2}
  (
  \bm x_j^\top \bm c
  +
  \|
  \bm x_j - \frac{\bm x_j^\top \bm a}{\|\bm a\|_2^2} \bm a
  \|_2
  \|
  \bm b - \frac{\bm b^\top \bm a}{\|\bm a\|_2^2} \bm a
  \|_2
  -
  \frac{
  \bm a^\top \bm b
  }{
  \|\bm a\|_2
  }
  \frac{
  \bm x_j^\top \bm a
  }{
  \|\bm a\|_2
  }
  )
  &
  \text{ otherwise}.
  }
  \\
  \label{eq:prf-y}
  u_j^-
  &:=
  \mycase{
  \frac{1}{2}
  (
  - \bm x_j^\top \bm c + \|\bm x_j\|_2 \|\bm b\|_2
  )
  &
  \text{ if }
  \frac{
  \bm b^\top \bm a
  }{
  \|\bm b\|_2
  }
  \le
  \frac{
  \bm x_j^\top \bm a
  }{
  \|\bm x_j\|_2
  }, 
  \\
  \frac{1}{2}
  (
  - \bm x_j^\top \bm c
  +
  \|
  \bm x_j - \frac{\bm x_j^\top \bm a}{\|\bm a\|_2^2} \bm a
  \|_2
  \|
  \bm b - \frac{\bm b^\top \bm a}{\|\bm a\|_2^2} \bm a
  \|_2
  +
  \frac{
  \bm a^\top \bm b
  }{
  \|\bm a\|_2
  }
  \frac{
  \bm x_j^\top \bm a
  }{
  \|\bm a\|_2
  }
  )
  &
  \text{ otherwise}.
  }
 \end{align}
 Then,
 for $j \in [D]$, 
 \begin{align*}
  \max\{u_j^+, u_j^-\} < 1
  ~~~
  \Rightarrow
  ~~~
  |\bm x_j^\top \bm \theta_2^*| < 1, 
 \end{align*}
 i.e.,
 the $j^{\rm th}$ feature is non-active in the optimal solution
 of LASSO with the regularization parameter $\lambda_2$. 
\end{prop}
For the proof of this proposition, see \cite{Liu2014}. 

\begin{proof}[Proof of Theorem~\ref{theo:main}]
 Remember that,
 for a feature indexed by 
 $j$,
 a feature indexed by 
 $j^\prime$
 represent a feature corresponding to one of its descendant nodes,
 i.e.,
 $j^\prime \in De(j)$.
 To prove the theorem,
 using the result of Proposition~\ref{prop:liu},
 it is suffice to show that
 \begin{align}
  \label{eq:what-we-show}
  U_j^+ < 1
  ~~~
  \Rightarrow
  ~~~
  u_{j^\prime}^+ < 1
  ~~~~~
  \text{ and }
  ~~~~~
  U_j^- < 1
  ~~~
  \Rightarrow
  ~~~
  u_{j^\prime}^- < 1.
 \end{align}
 
 First,
 we prove the case with
 $\lambda_{t-1} = \lambda_{\rm max} = \max_{j \in [D]} |\bm x_j^\top \bm y|$.
 In this case,
 from the primal-dual relationship of the optimal LASSO solution,
 the optimal dual solution at $\lambda_{t-1}$ is written as
 $\bm \theta^*_{t-1} = \frac{\bm y}{\lambda_{t-1}}$. 
 Since it means 
 $\bm a = \zero$,
 from \eq{eq:prf-x} and \eq{eq:prf-y},
 \begin{align*}
  u_{j^\prime}^+ = \frac{1}{2} (\bm x_{j^\prime}^\top \bm c + \|\bm x_{j^\prime}\|_2 \|\bm b\|_2),
  ~~~
  u_{j^\prime}^- = \frac{1}{2} (- \bm x_{j^\prime}^\top \bm c + \|\bm x_{j^\prime}\|_2 \|\bm b\|_2).
 \end{align*}
 Using the fact that
 \begin{align}
  \bm x_{j^\prime}^\top \bm c
  =
  \sum_{i: c_i > 0}
  c_i x^i_{j^\prime}
  +
  \sum_{i: c_i < 0}
  c_i x^i_{j^\prime}
  \le 
  \sum_{i: c_i > 0}
  c_i x^i_{j^\prime}
  \le 
  \sum_{i: c_i > 0}
  c_i x^i_j
 \end{align}
 and
 \begin{align*}
  \|\bm x_{j^\prime}\|_2
  \le
  \|\bm x_j\|_2,
 \end{align*}
 we have
 \begin{align}
  \label{eq:prf-s}
  u_{j^\prime}^+
  &=
  \frac{1}{2} (\bm x_{j^\prime}^\top \bm c + \|\bm x_{j^\prime}\|_2 \|\bm b\|_2)
  \le
  \frac{1}{2} (\sum_{i: c_i > 0} c_i x^i_j + \|\bm x_j\|_2 \|\bm b\|_2)
  = P_1,
  \\
  \label{eq:prf-t}
  u_{j^\prime}^-
  &=
  \frac{1}{2} (- \bm x_{j^\prime}^\top \bm c + \|\bm x_{j^\prime}\|_2 \|\bm b\|_2)
  \le
  \frac{1}{2} (- \sum_{i: c_i < 0} c_i x^i_j + \|\bm x_j\|_2 \|\bm b\|_2)
  = M_1,
 \end{align}
 which proves the theorem when $\lambda_{t-1} = \lambda_{\rm max}$.

 Next,
 we consider
 $\lambda_{t-1} < \lambda_{\rm max}$. 
 In this case,
 we cannot decide which of the two cases in
 \eq{eq:prf-x}
 and 
 \eq{eq:prf-y}
 are applied.
 In the first case
 of 
 \eq{eq:prf-x} 
 and 
 \eq{eq:prf-y}
 are applied,
 we can show that
 \begin{align*}
  u_{j^\prime}^+ \le P_1
  ~~~~~
  \text{ and }
  ~~~~~
  u_{j^\prime}^- \le M_1
 \end{align*}
 in the same way as 
 \eq{eq:prf-s}
 and 
 \eq{eq:prf-t}.
 On the other hand,
 when the second case
 of 
 \eq{eq:prf-x} 
 is applied,
 we can show that
 \begin{align*}
  u_{j^\prime}^+
  &=
  \frac{1}{2}
  (\bm x_{j^\prime}^\top \bm c
  +
  \|\bm x_{j^\prime} - \frac{\bm x_{j^\prime}^\top \bm a}{\|\bm a\|_2^2} \bm a\|_2
  \|\bm b - \frac{\bm b^\top \bm a}{\|\bm a\|_2^2} \bm a\|_2
  -
  \frac{\bm a^\top \bm b}{\|\bm a\|_2}
  \frac{\bm x_{j^\prime}^\top \bm a}{\|\bm a\|_2})
  \\
  &=
  \frac{1}{2}
  (\bm x_{j^\prime}^\top \bm d
  +
  \|\bm x_{j^\prime} - \frac{\bm x_{j^\prime}^\top \bm a}{\|\bm a\|_2^2} \bm a\|_2
  \|\bm b - \frac{\bm b^\top \bm a}{\|\bm a\|_2^2} \bm a\|_2)
  \\
  &\le
  \frac{1}{2}
  (\sum_{i:d_i>0} d_i x^i_j
  +
  \|\bm x_{j} \|_2
  \|\bm b - \frac{\bm b^\top \bm a}{\|\bm a\|_2^2} \bm a\|_2) = P_2,
 \end{align*}
 where
 we used
 \begin{align*}
  \bm x_{j^\prime}^\top \bm d
  =
  \sum_{i: d_i > 0}
  d_i x^i_{j^\prime}
  +
  \sum_{i: d_i < 0}
  d_i x^i_{j^\prime}
  \le 
  \sum_{i: d_i > 0}
  d_i x^i_{j^\prime}
  \le 
  \sum_{i: d_i > 0}
  d_i x^i_j
  \end{align*}
and
\begin{align*}
 \|\bm x_{j^\prime} - \frac{\bm x_{j^\prime}^\top \bm a}{\|\bm a\|_2^2} \bm a\|_2
 =
 \sqrt{
 \|\bm x_{j^\prime}\|_2^2 - \frac{(\bm x_{j^\prime}^\top \bm a)^2}{\|\bm a\|_2^2}
 }
 \le
 \|\bm x_{j^\prime}\|_2
 \le
 \|\bm x_{j}\|_2
\end{align*}
 When
 the second case of 
 \eq{eq:prf-t}
 is applied, 
 we can show the following  in the same way as above:
 \begin{align*}
  u_{j^\prime}^-
  &=
   \frac{1}{2}
  (  - \bm x_{j^\prime}^\top \bm c
  +
  \|
  \bm x_{j^\prime} - \frac{\bm x_{j^\prime}^\top \bm a}{\|\bm a\|_2^2} \bm a
  \|_2
  \|
  \bm b - \frac{\bm b^\top \bm a}{\|\bm a\|_2^2} \bm a
  \|_2
  +
  \frac{
  \bm a^\top \bm b
  }{
  \|\bm a\|_2
  }
  \frac{
  \bm x_{j^\prime}^\top \bm a
  }{
  \|\bm a\|_2
  } )
  \\
  &=
   \frac{1}{2}
  (  - \bm x_{j^\prime}^\top \bm d
  +
  \|
  \bm x_{j^\prime} - \frac{\bm x_{j^\prime}^\top \bm a}{\|\bm a\|_2^2} \bm a
  \|_2
  \|
  \bm b - \frac{\bm b^\top \bm a}{\|\bm a\|_2^2} \bm a
  \|_2)
  \\
  &\le
   \frac{1}{2}
  ( - \sum_{i:d_i < 0} d_i x^i_j
  +
  \|\bm x_j\|_2
  \|
  \bm b - \frac{\bm b^\top \bm a}{\|\bm a\|_2^2} \bm a
  \|_2) = M_2. 
 \end{align*}

 Finally,
 when the original covariates $z^i_j$ is binary,
 we can judge which of the two cases in
 \eq{eq:prf-s}
 and
 \eq{eq:prf-t}
 are applied 
 by using the information available at the node $j$,
 and a slightly tighter bounds can be obtained
 (the proof of how one can make this judgment is omitted, but it can be shown in a similar manner as above). 
\end{proof}

%% file: appB.tex
\section{Additional Experimental Results}
\label{app:results}
In this section, we show the results on several benchmark datasets.
For each dataset, 
(a) Computation time in seconds, 
(b) The number of traverse nodes, 
(c) The number of active features, 
(d) The number of solving LASSO in IB, 
(e) The number of non-screened out features and total active features, 
(f) Computation total time in seconds
are plotted in the following figures with $\delta$ = 1.5
(left) and $\delta$ = 2.0 (right).
\subsection{Results on usps}
\begin{figure}[ht]
\begin{center}
\begin{tabular}{cc}
\includegraphics[scale=0.5]{./usps_time_15.eps}
&
\includegraphics[scale=0.5]{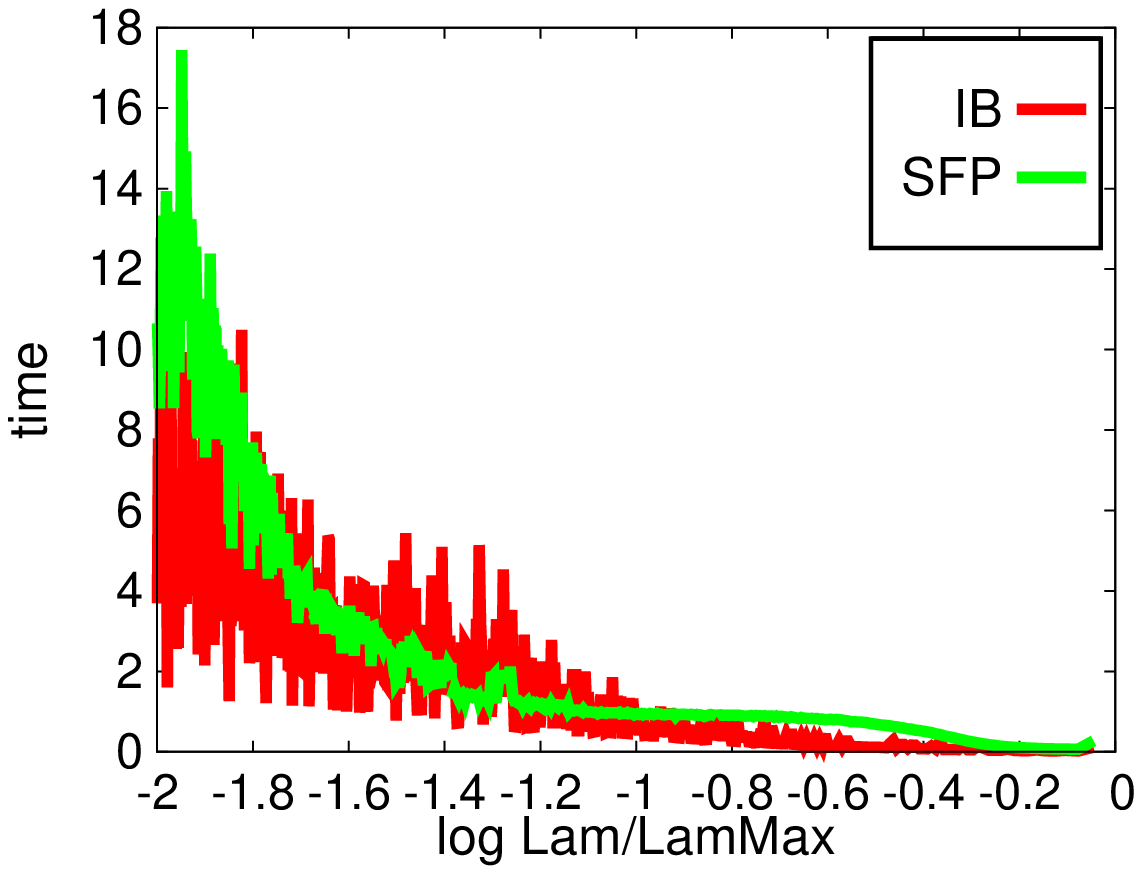}
\\
\multicolumn{2}{c}{(a) Computation time in seconds}
\\
\includegraphics[scale=0.5]{./usps_trav_15.eps}
&
\includegraphics[scale=0.5]{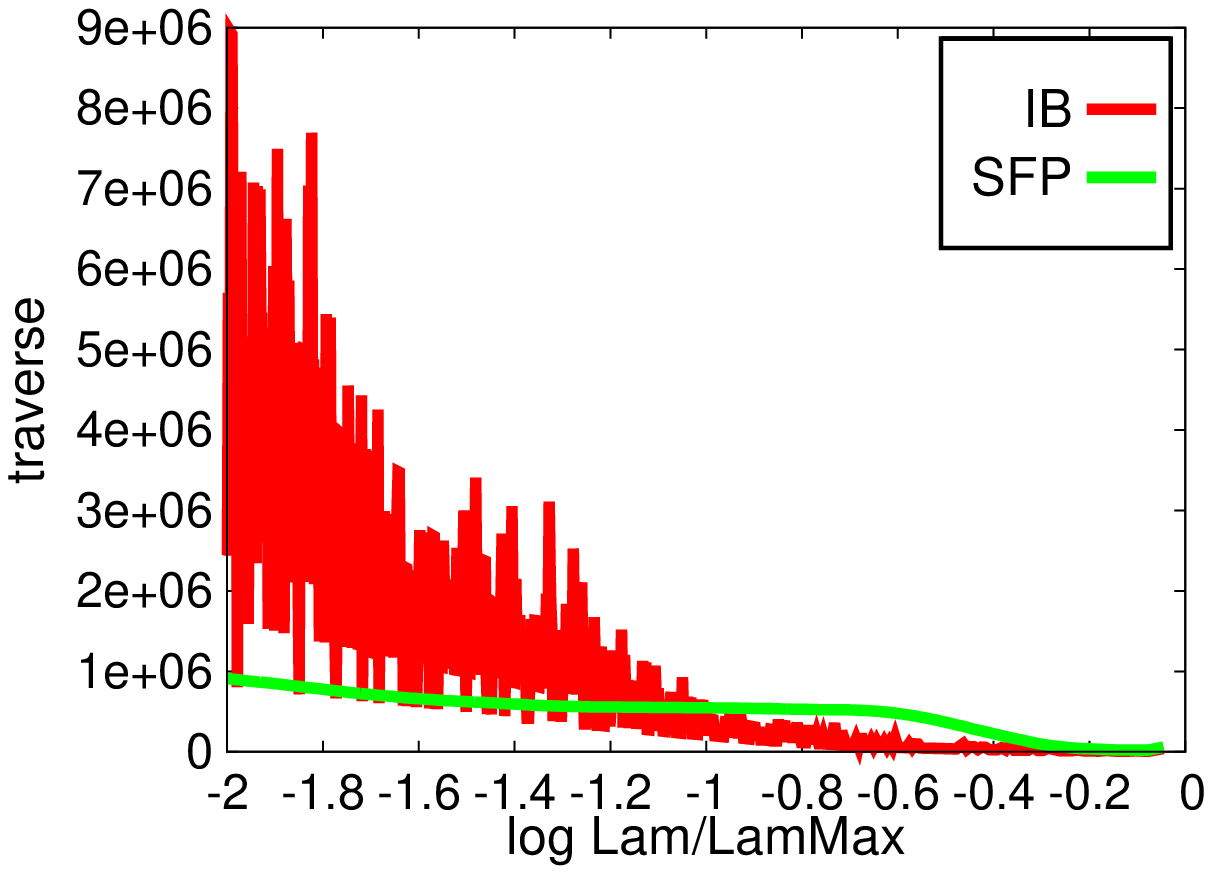}
\\
\multicolumn{2}{c}{(b) The number of traverse nodes}
\\
\includegraphics[scale=0.5]{./usps_feature_15.eps}
&
\includegraphics[scale=0.5]{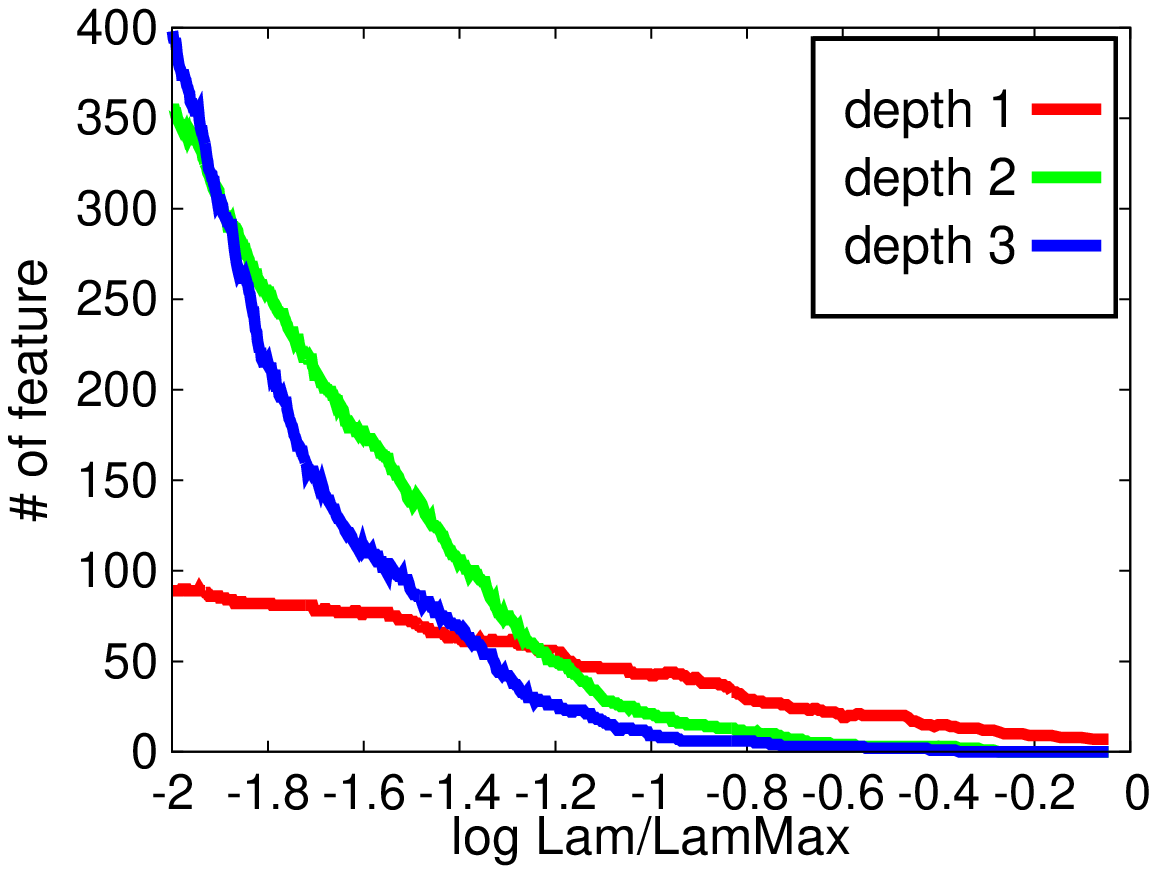}
\\
\multicolumn{2}{c}{(c) The number of active features}
\end{tabular}
\end{center}
\end{figure}

\begin{figure}[ht]
\begin{center}
\begin{tabular}{cc}
\includegraphics[scale=0.5]{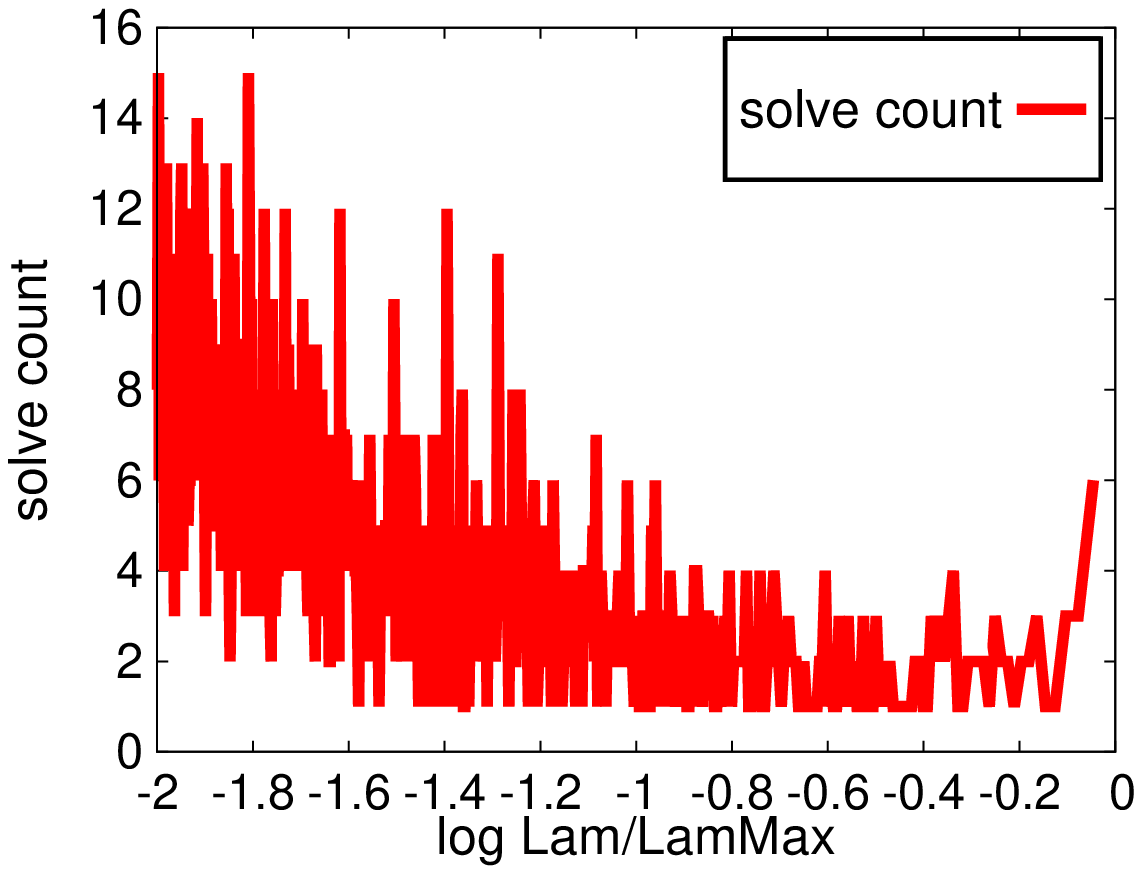}
&
\includegraphics[scale=0.5]{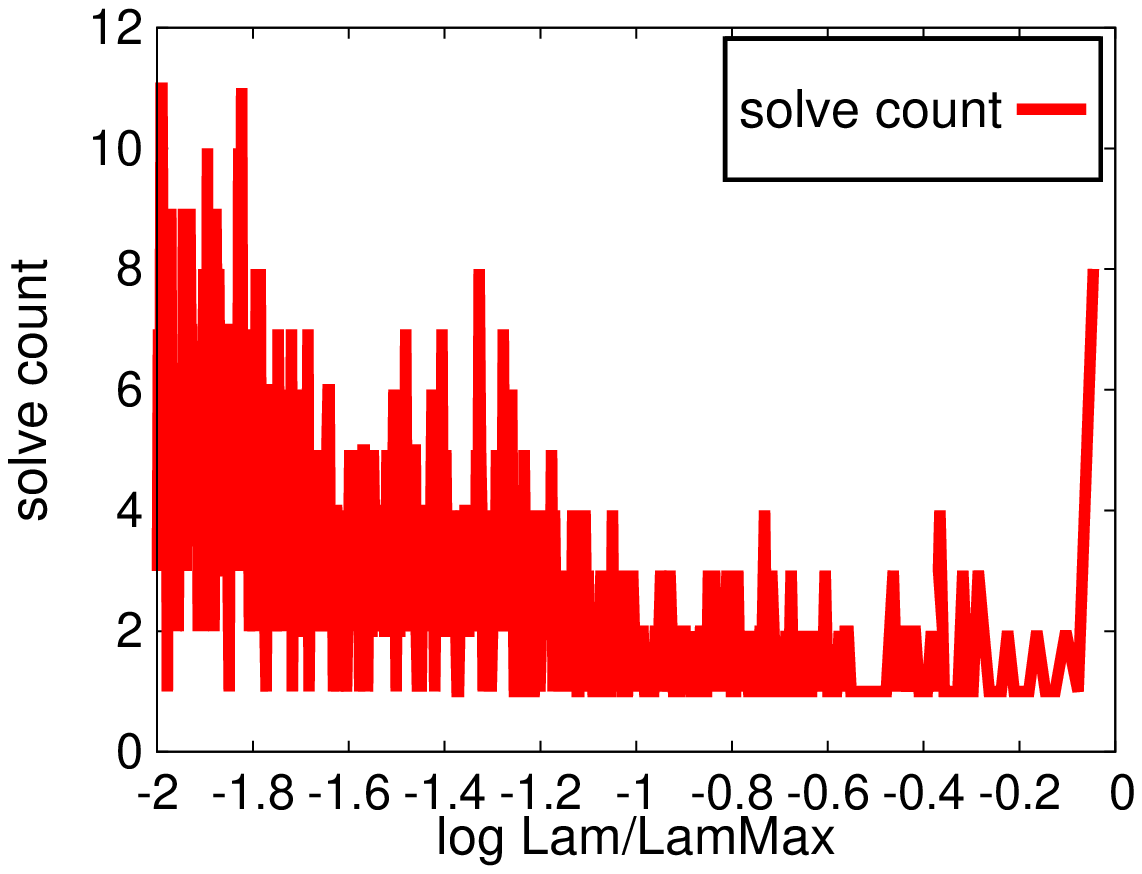}
\\
\multicolumn{2}{c}{(d) The number of solving LASSO in IB}
\\
\includegraphics[scale=0.5]{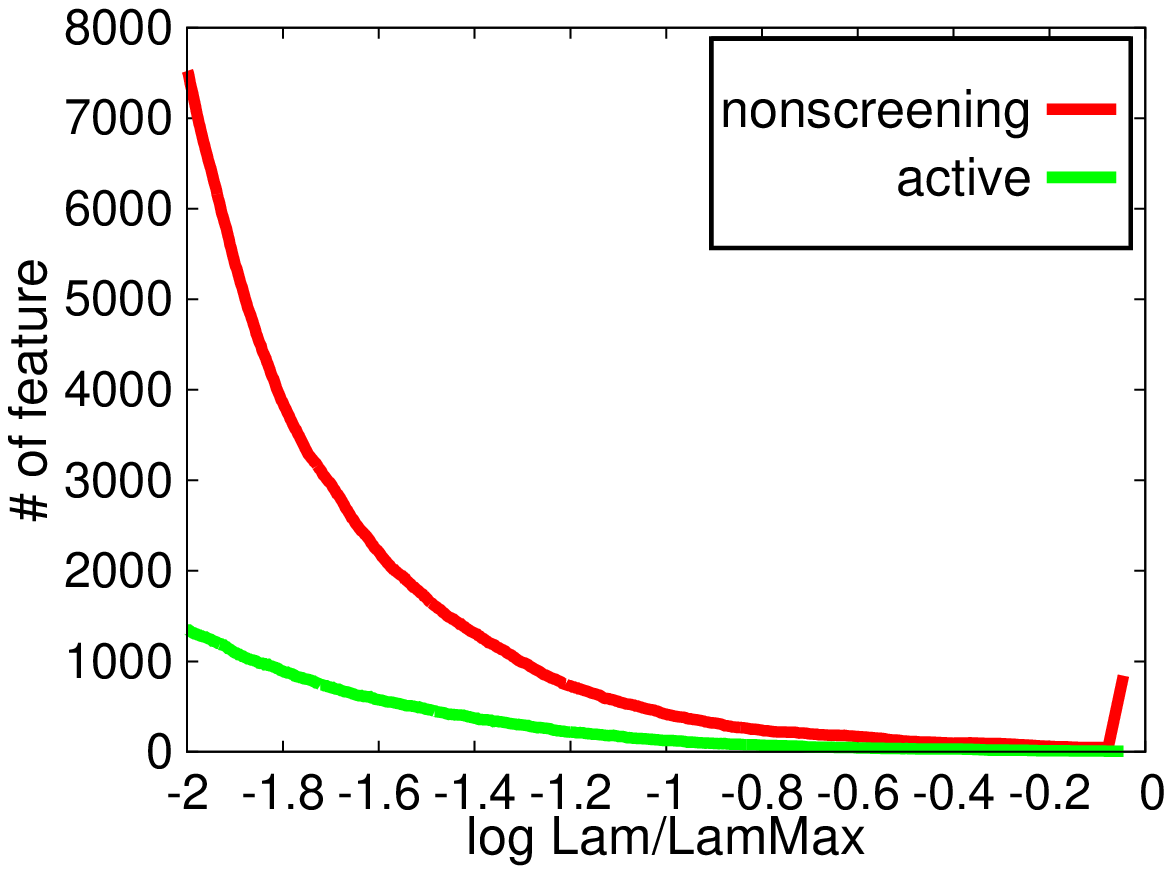}
&
\includegraphics[scale=0.5]{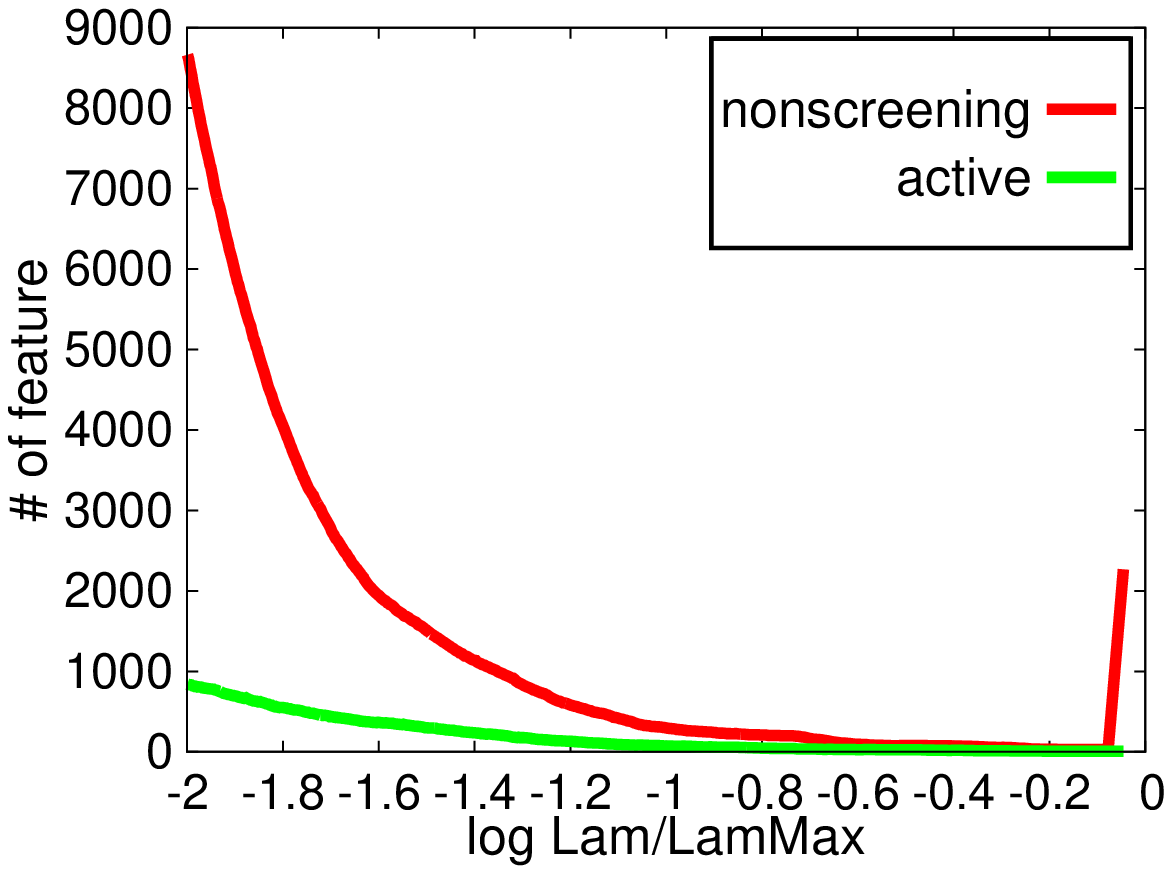}
\\
\multicolumn{2}{c}{(e) The number of non-screened out features and total active features}
\\
\includegraphics[scale=0.5]{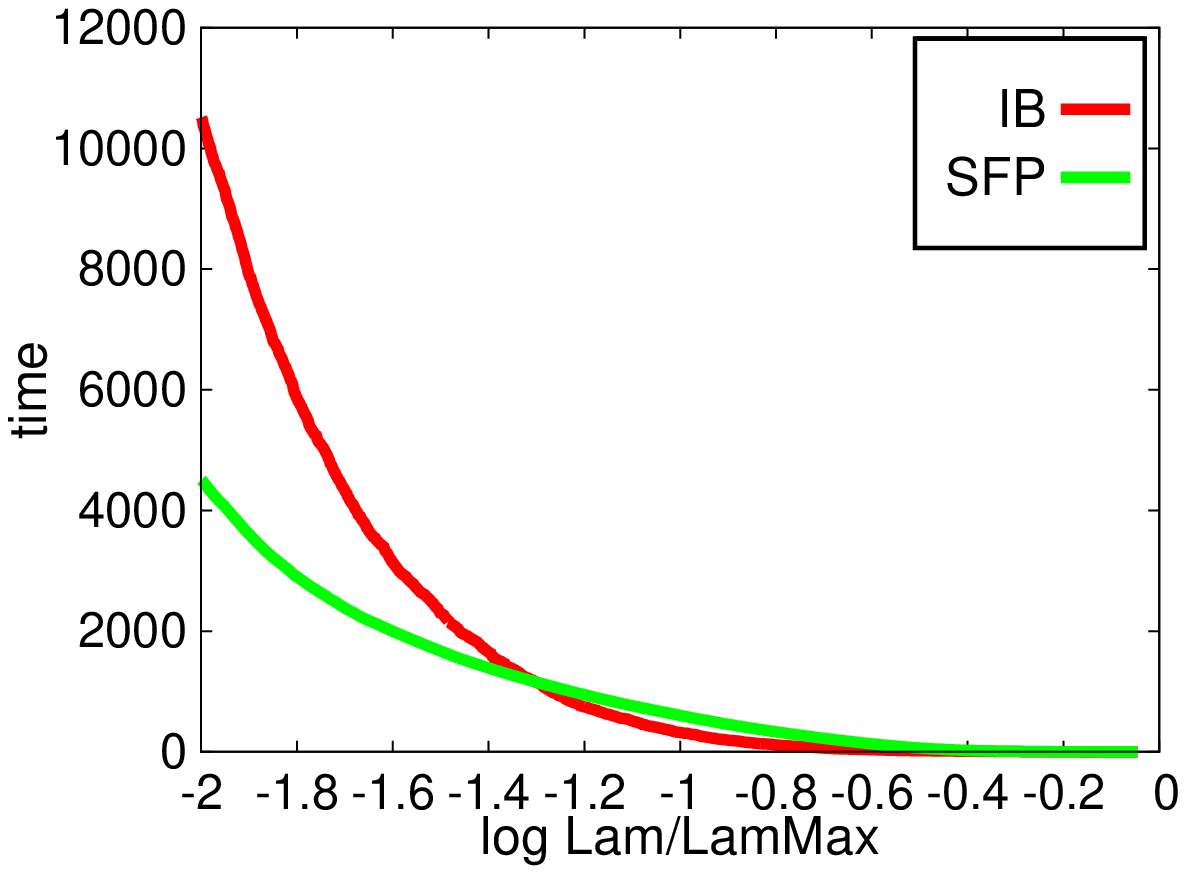}
&
\includegraphics[scale=0.5]{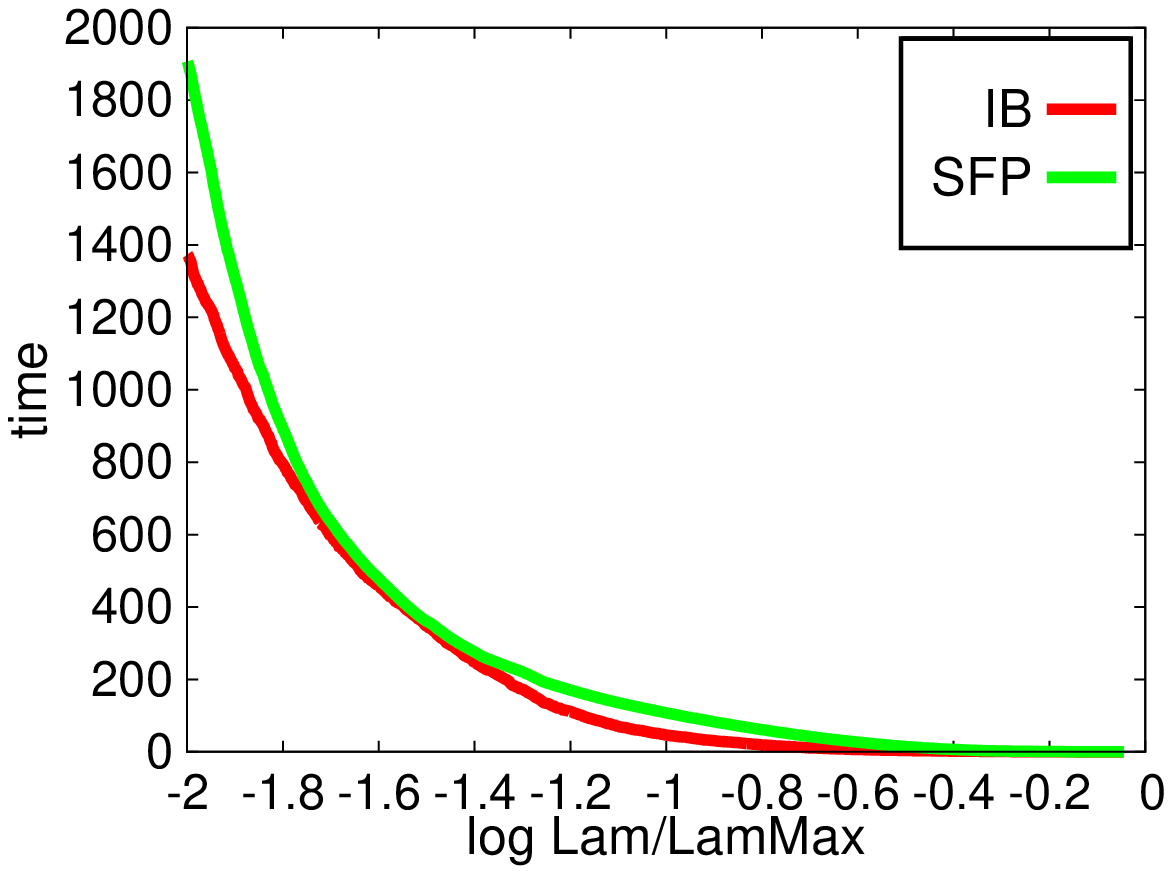}
\\
\multicolumn{2}{c}{(f) Computation total time in seconds}
\end{tabular}
\end{center}
\end{figure}

\clearpage

\subsection{Results on madelon}
\begin{figure}[ht]
\begin{center}
\begin{tabular}{cc}
\includegraphics[scale=0.5]{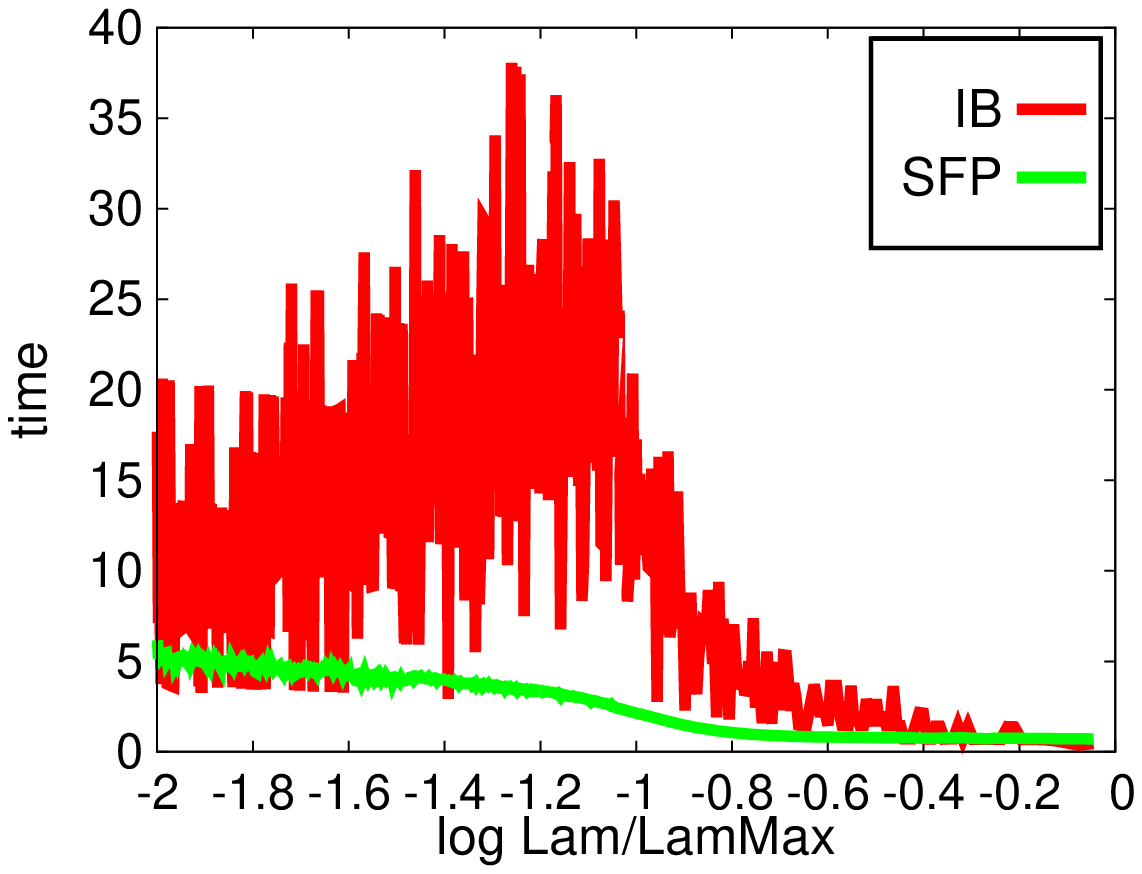}
&
\includegraphics[scale=0.5]{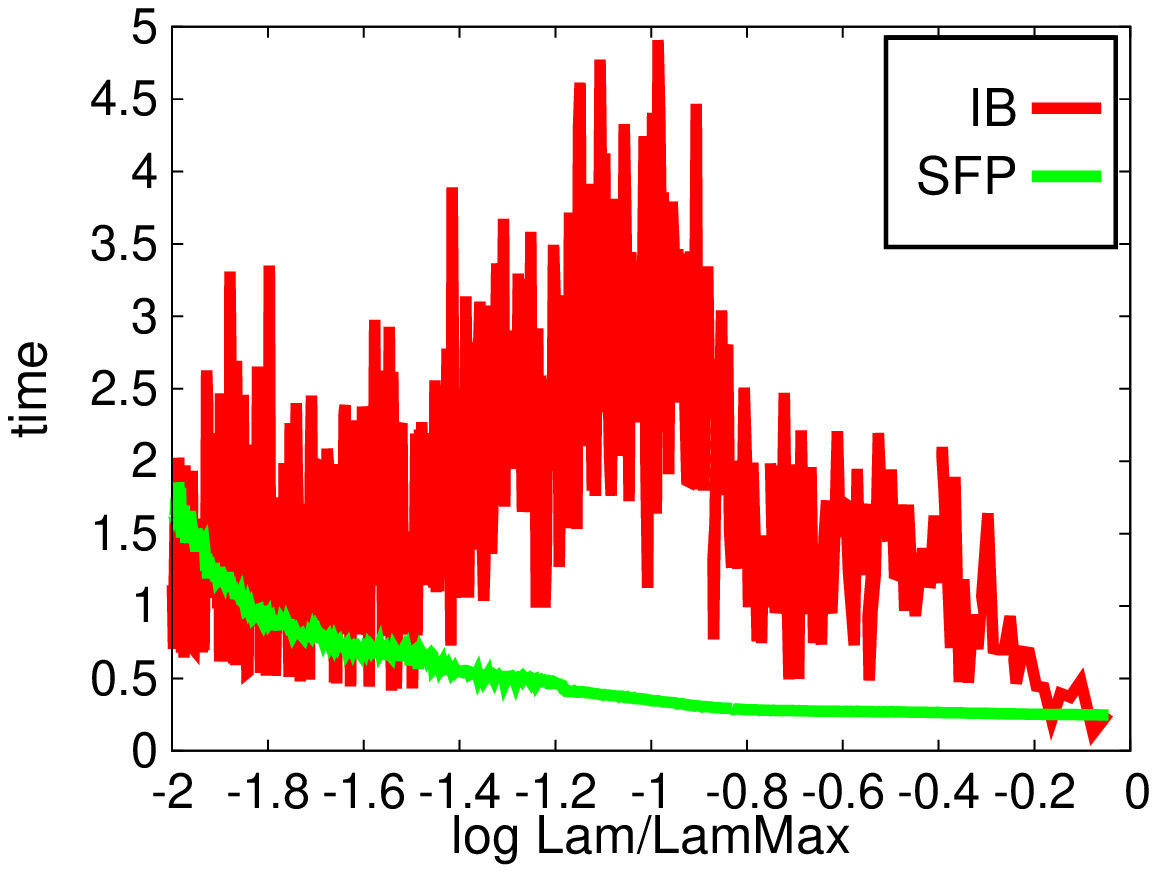}
\\
\multicolumn{2}{c}{(a) Computation time in seconds}
\\
\includegraphics[scale=0.5]{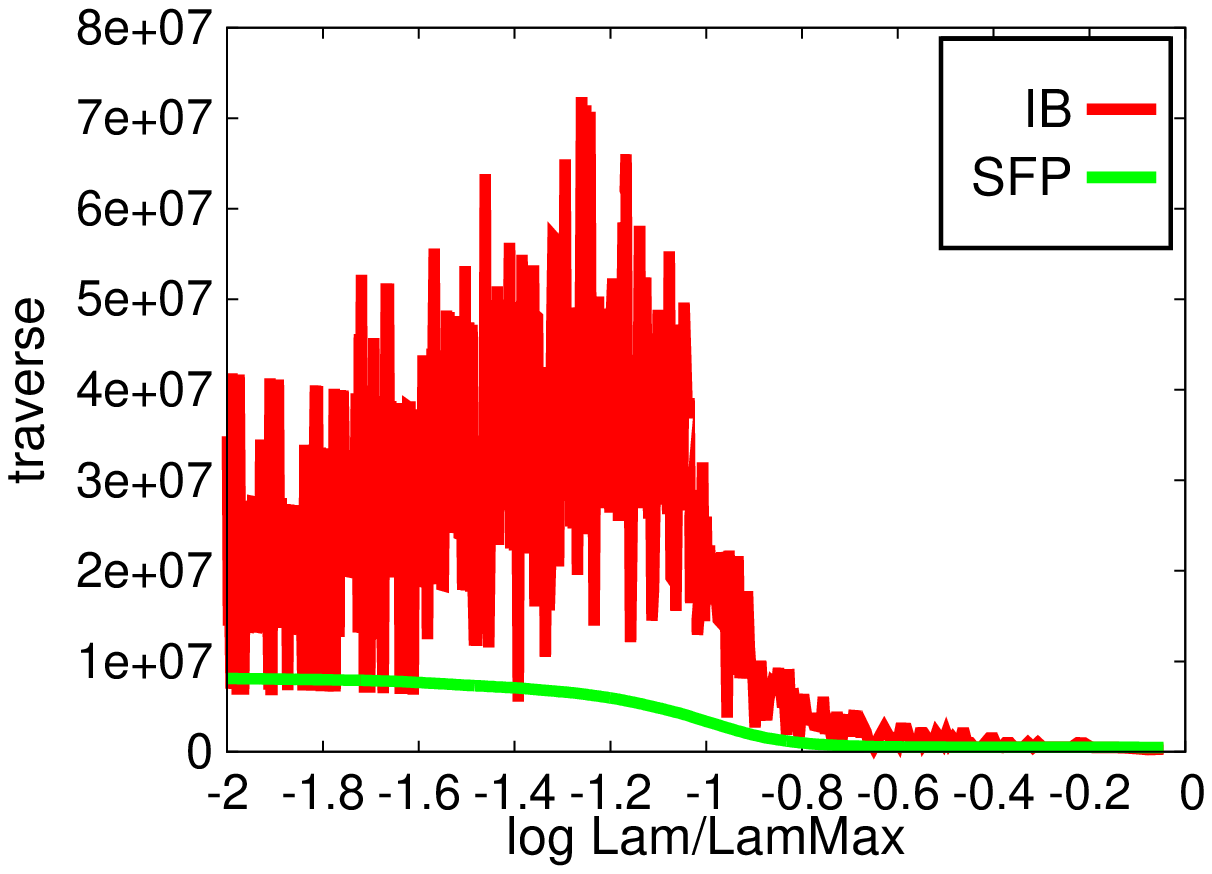}
&
\includegraphics[scale=0.5]{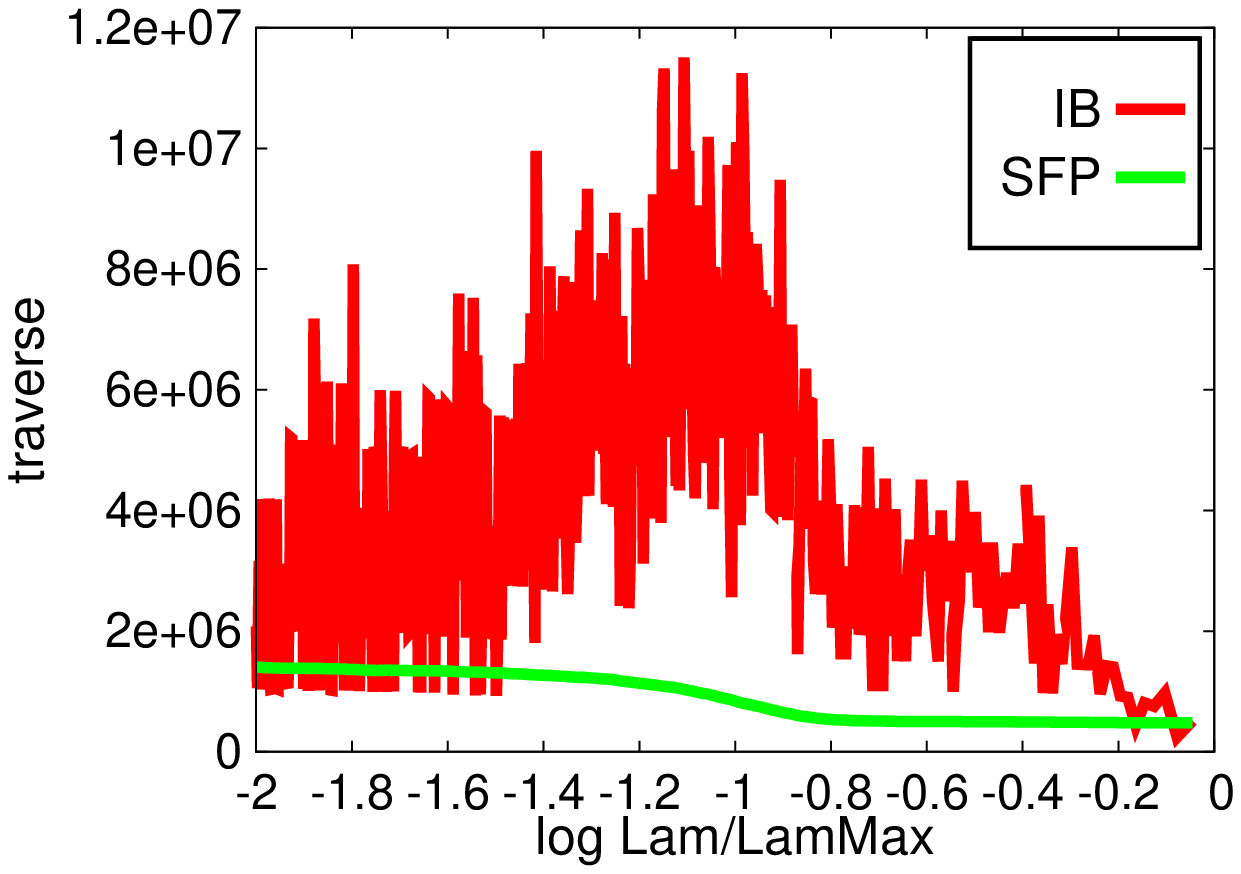}
\\
\multicolumn{2}{c}{(b) The number of traverse nodes}
\\
\includegraphics[scale=0.5]{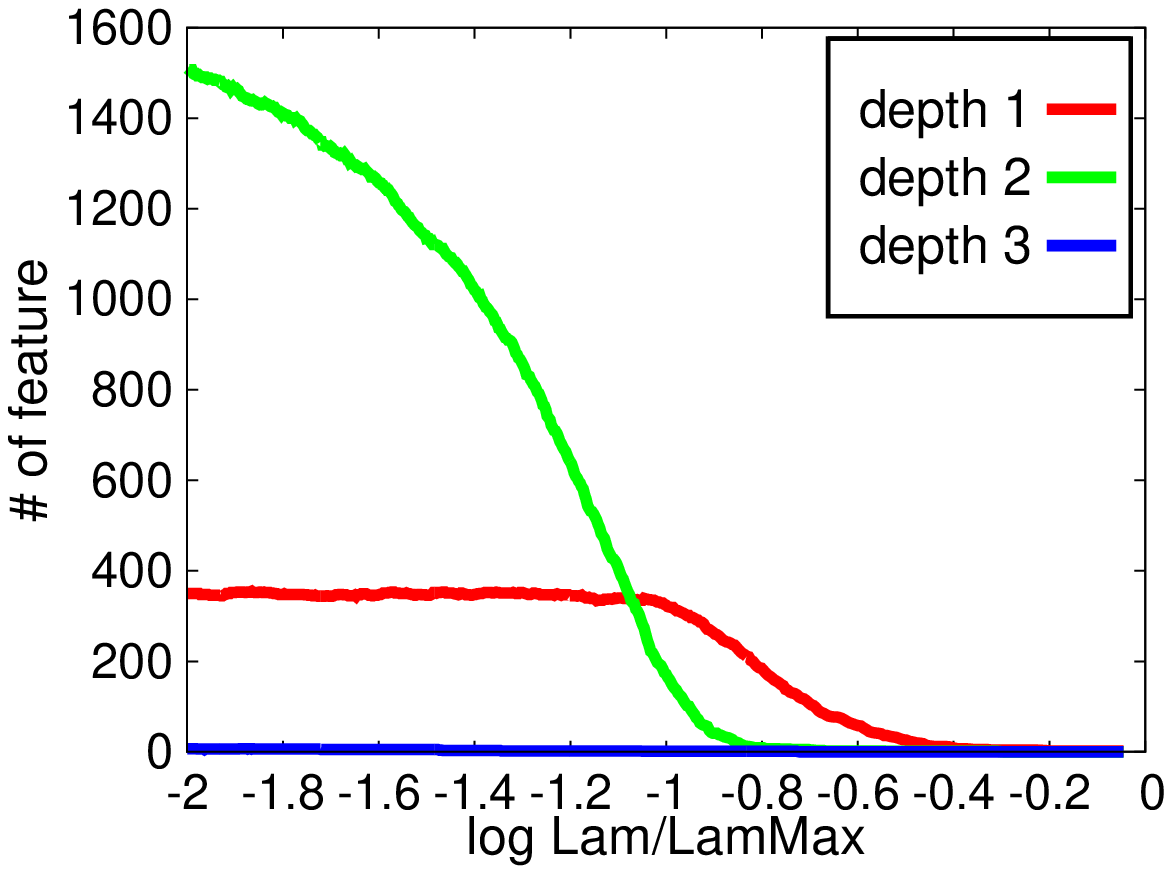}
&
\includegraphics[scale=0.5]{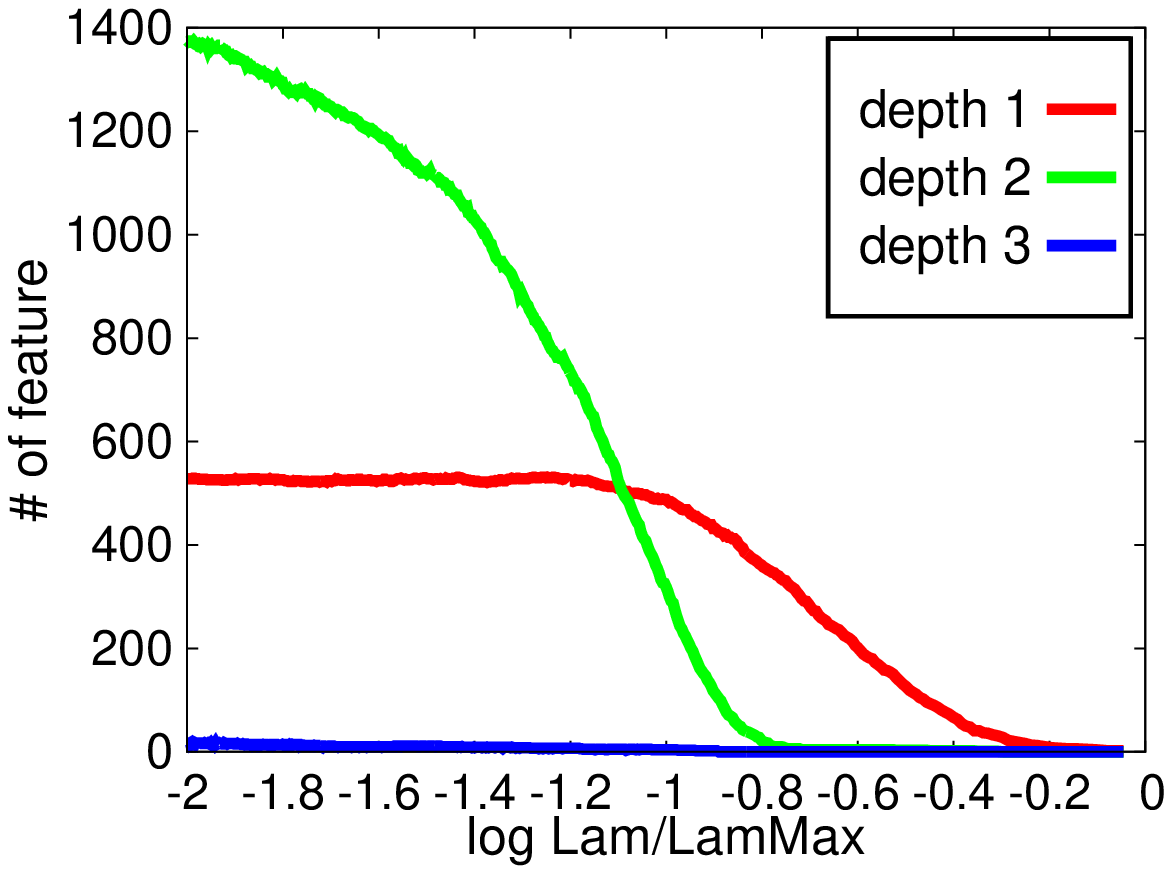}
\\
\multicolumn{2}{c}{(c) The number of active features}
\end{tabular}
\end{center}
\end{figure}

\begin{figure}[ht]
\begin{center}
\begin{tabular}{cc}
\includegraphics[scale=0.5]{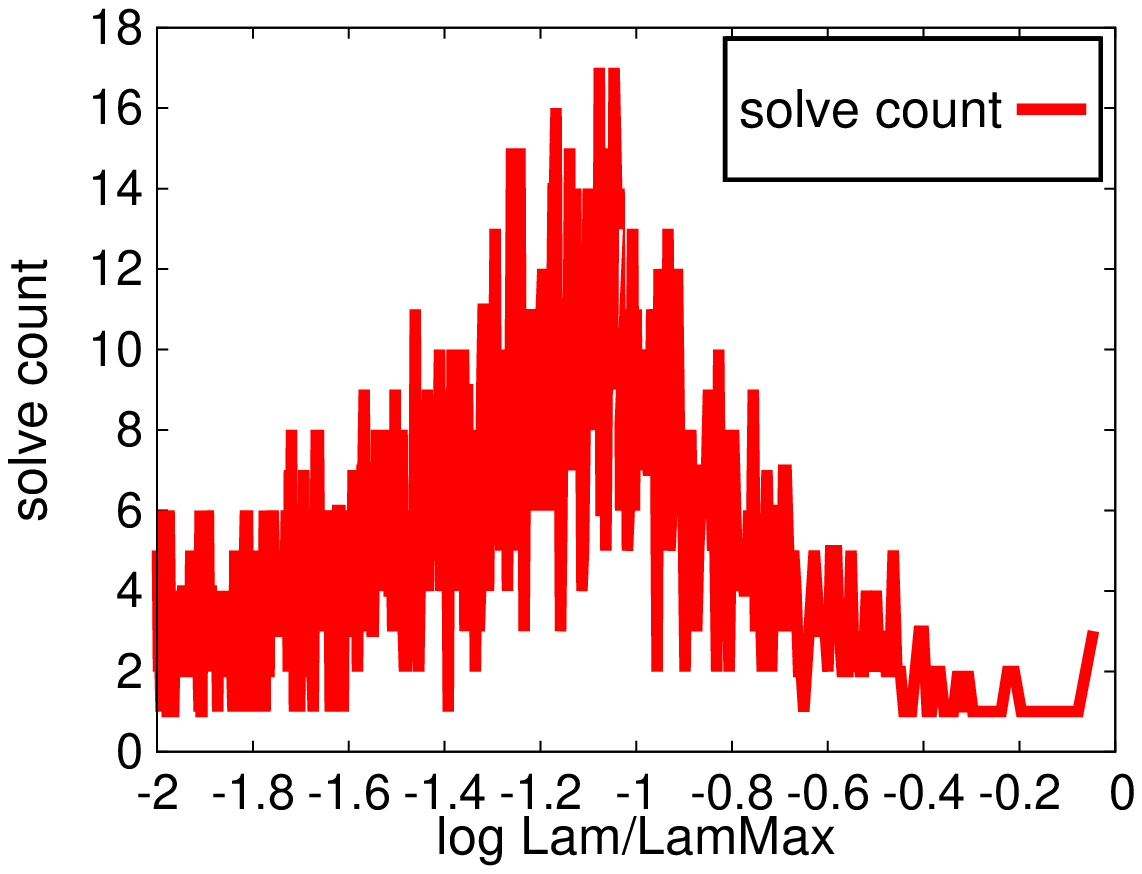}
&
\includegraphics[scale=0.5]{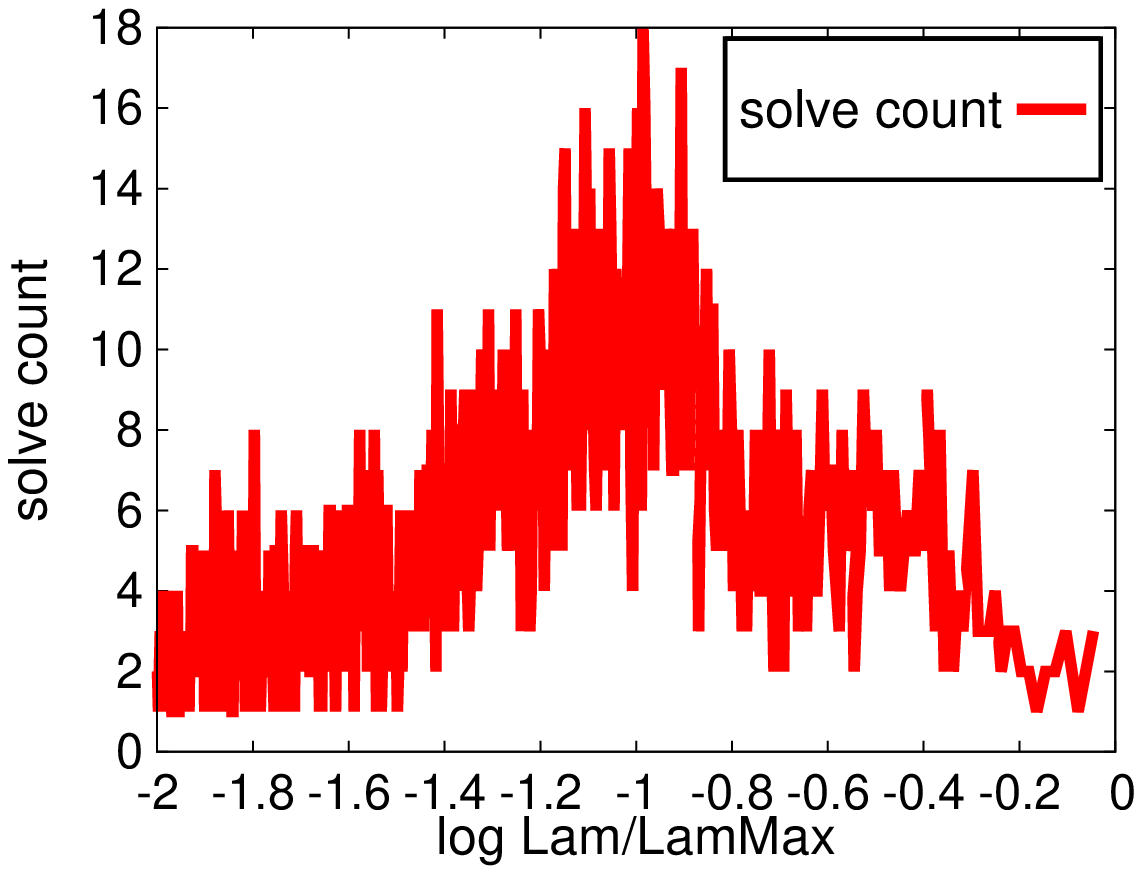}
\\
\multicolumn{2}{c}{(d) The number of solving LASSO in IB}
\\
\includegraphics[scale=0.5]{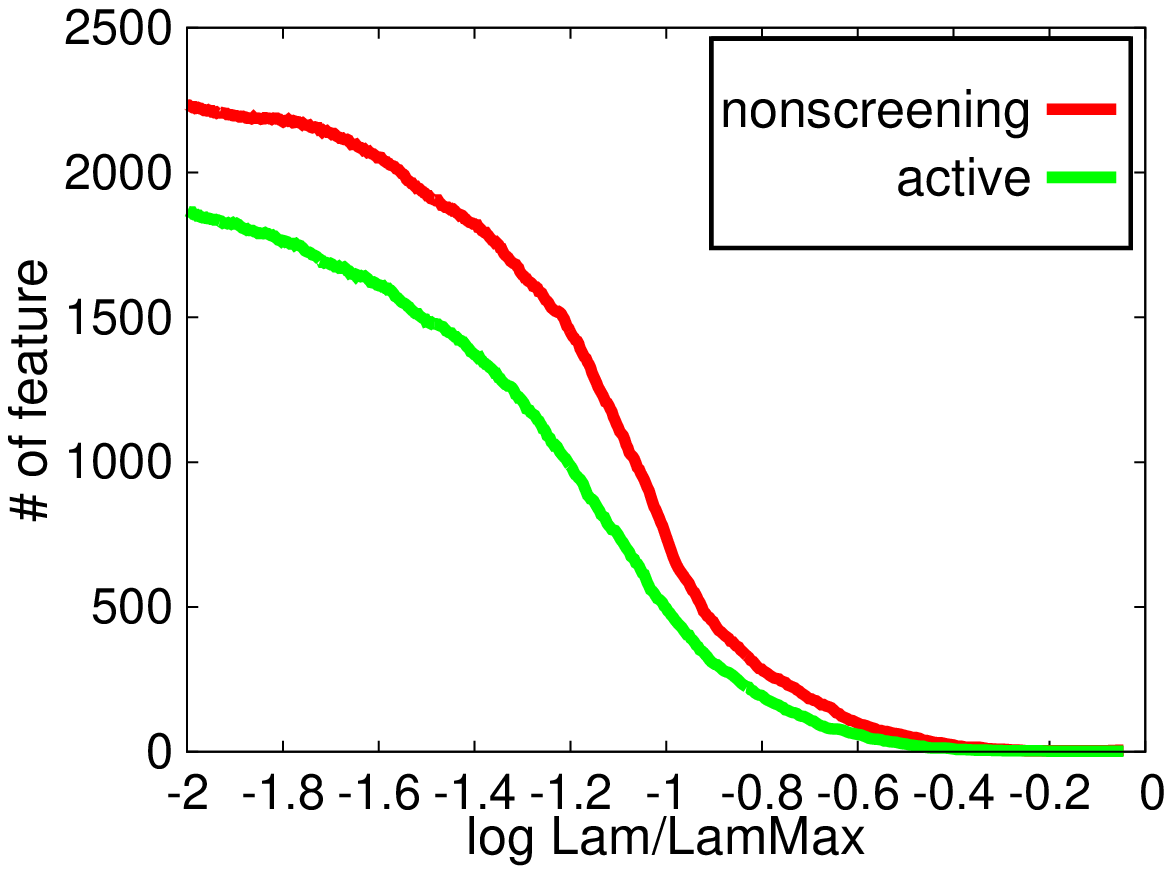}
&
\includegraphics[scale=0.5]{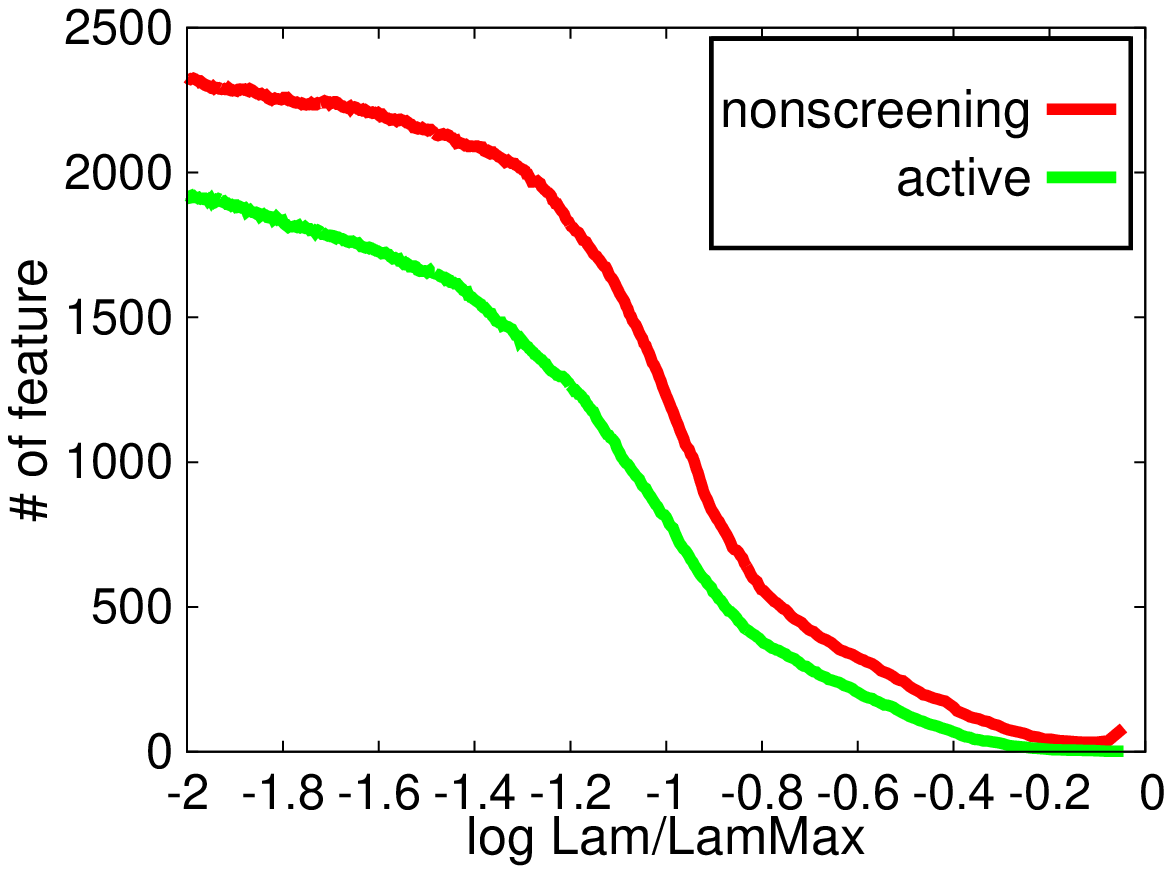}
\\
\multicolumn{2}{c}{(e) The number of non-screened out features and total active features}
\\
\includegraphics[scale=0.5]{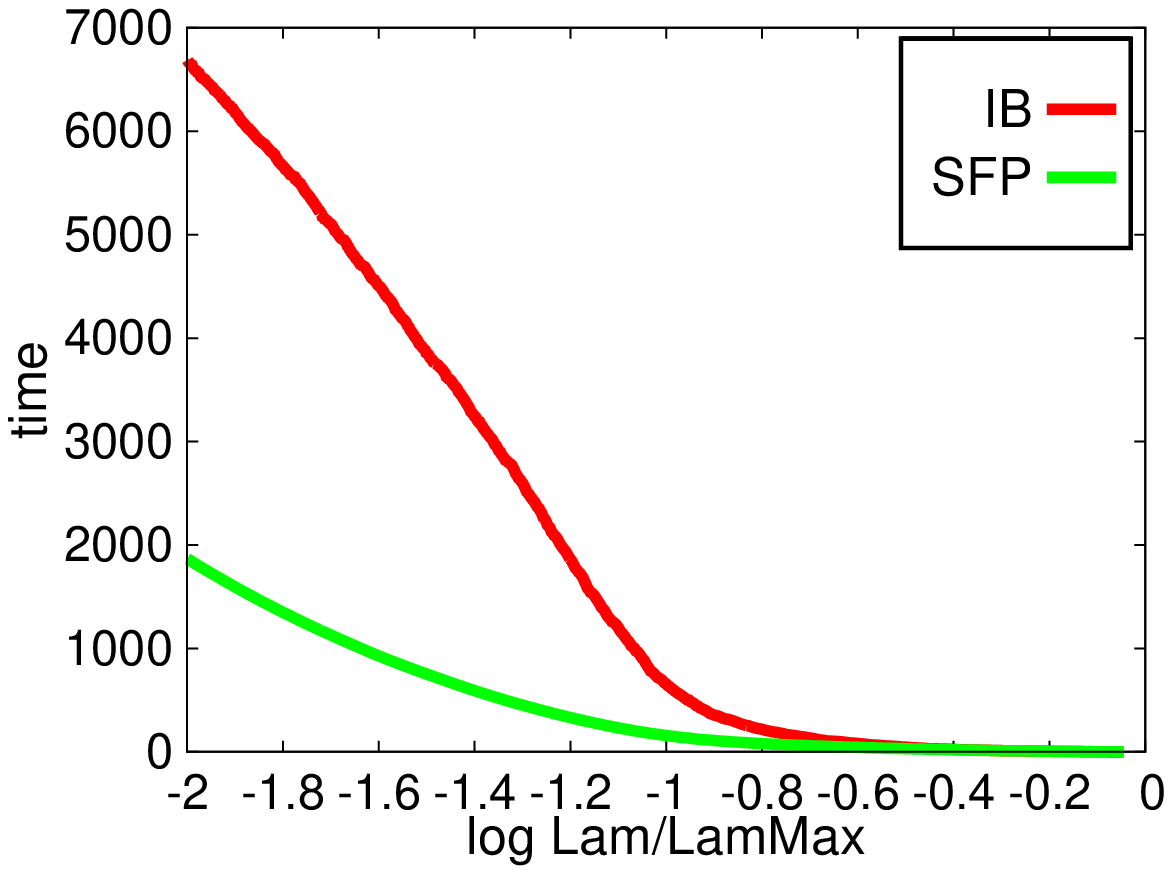}
&
\includegraphics[scale=0.5]{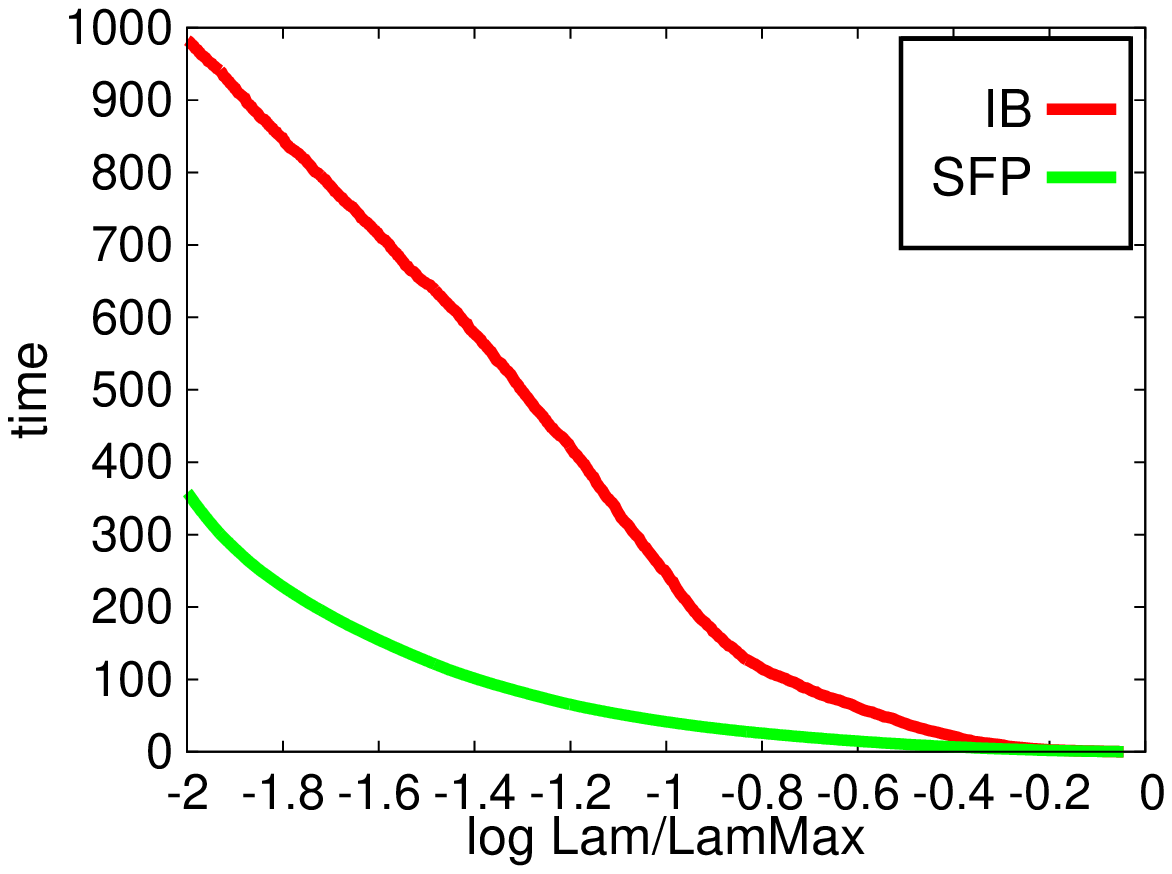}
\\
\multicolumn{2}{c}{(f) Computation total time in seconds}
\end{tabular}
\end{center}
\end{figure}

\clearpage

\subsection{Results on protein}
\begin{figure}[ht]
\begin{center}
\begin{tabular}{cc}
\includegraphics[scale=0.5]{./protein_time_15.eps}
&
\includegraphics[scale=0.5]{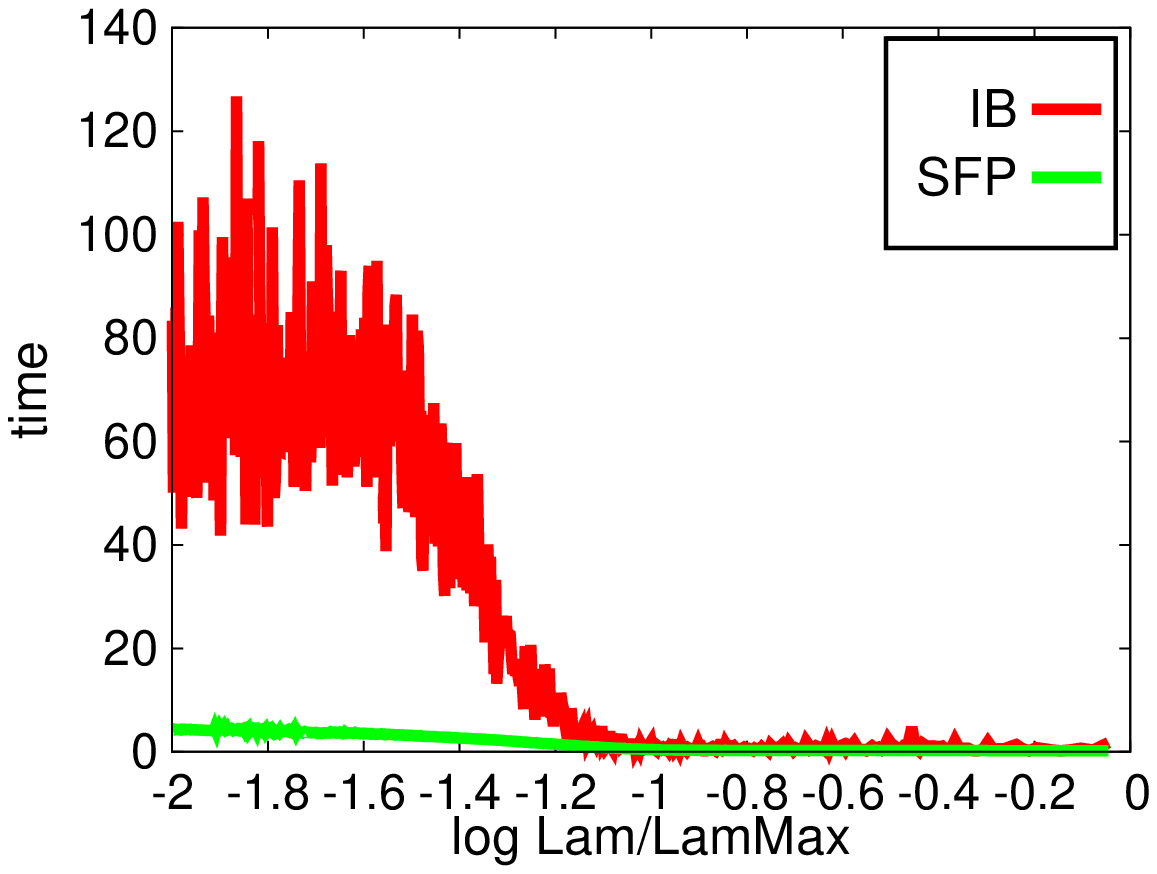}
\\
\multicolumn{2}{c}{(a) Computation time in seconds}
\\
\includegraphics[scale=0.5]{./protein_trav_15.eps}
&
\includegraphics[scale=0.5]{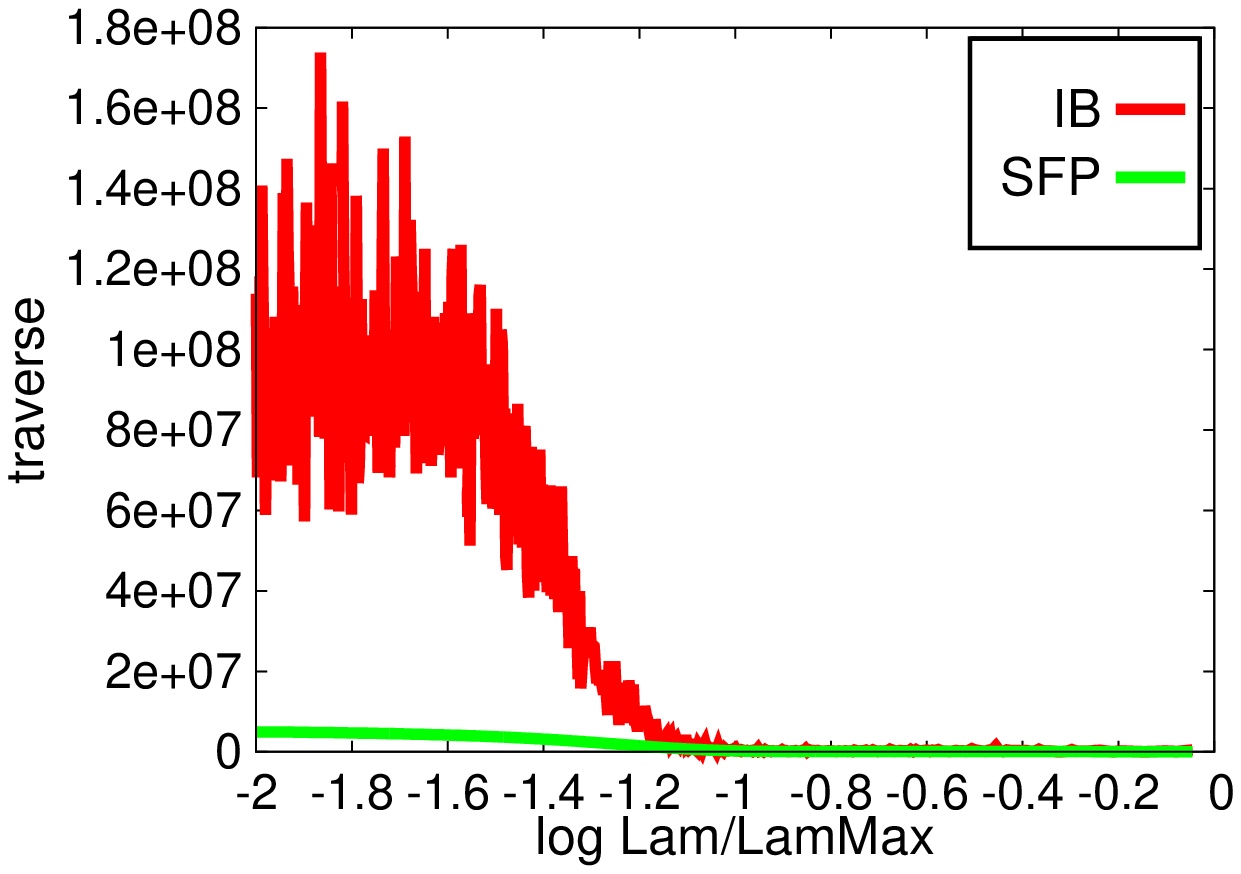}
\\
\multicolumn{2}{c}{(b) The number of traverse nodes}
\\
\includegraphics[scale=0.5]{./protein_feature_15.eps}
&
\includegraphics[scale=0.5]{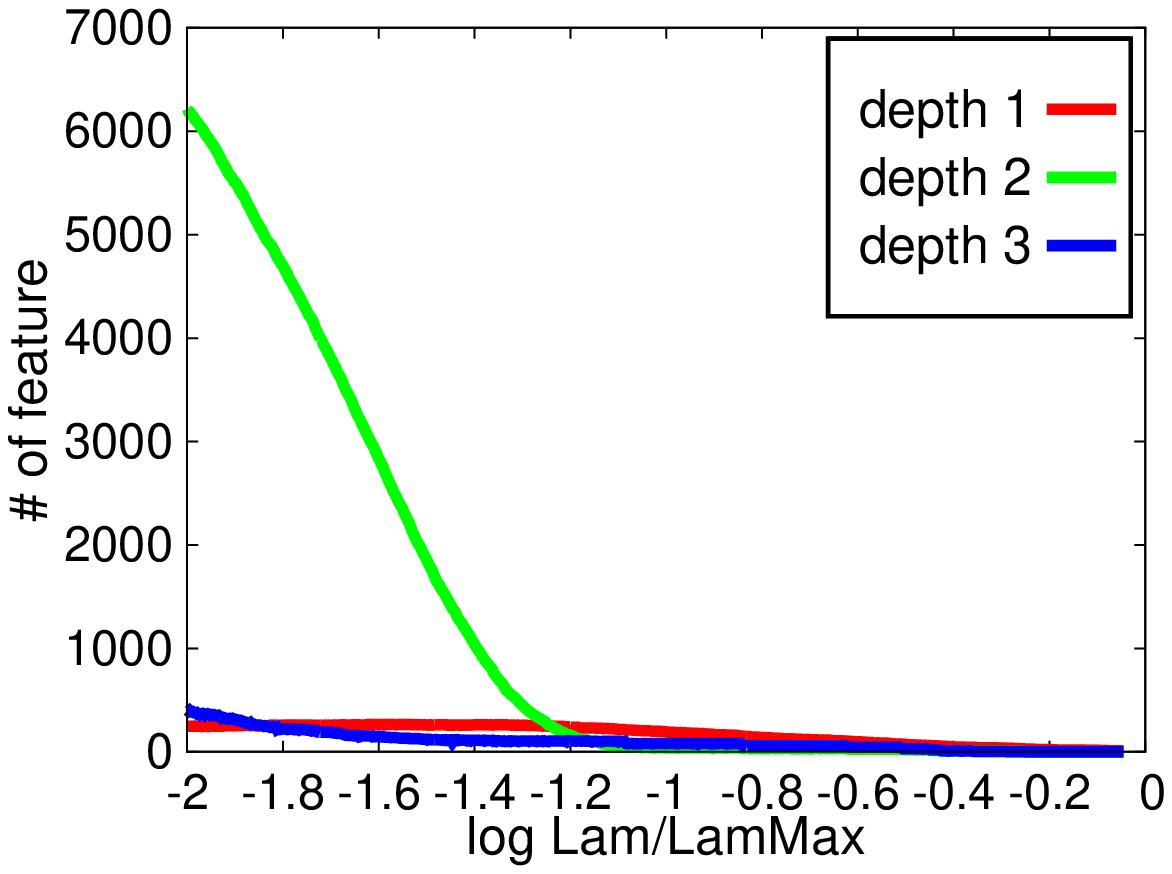}
\\
\multicolumn{2}{c}{(c) The number of active features}
\end{tabular}
\end{center}
\end{figure}

\begin{figure}[ht]
\begin{center}
\begin{tabular}{cc}
\includegraphics[scale=0.5]{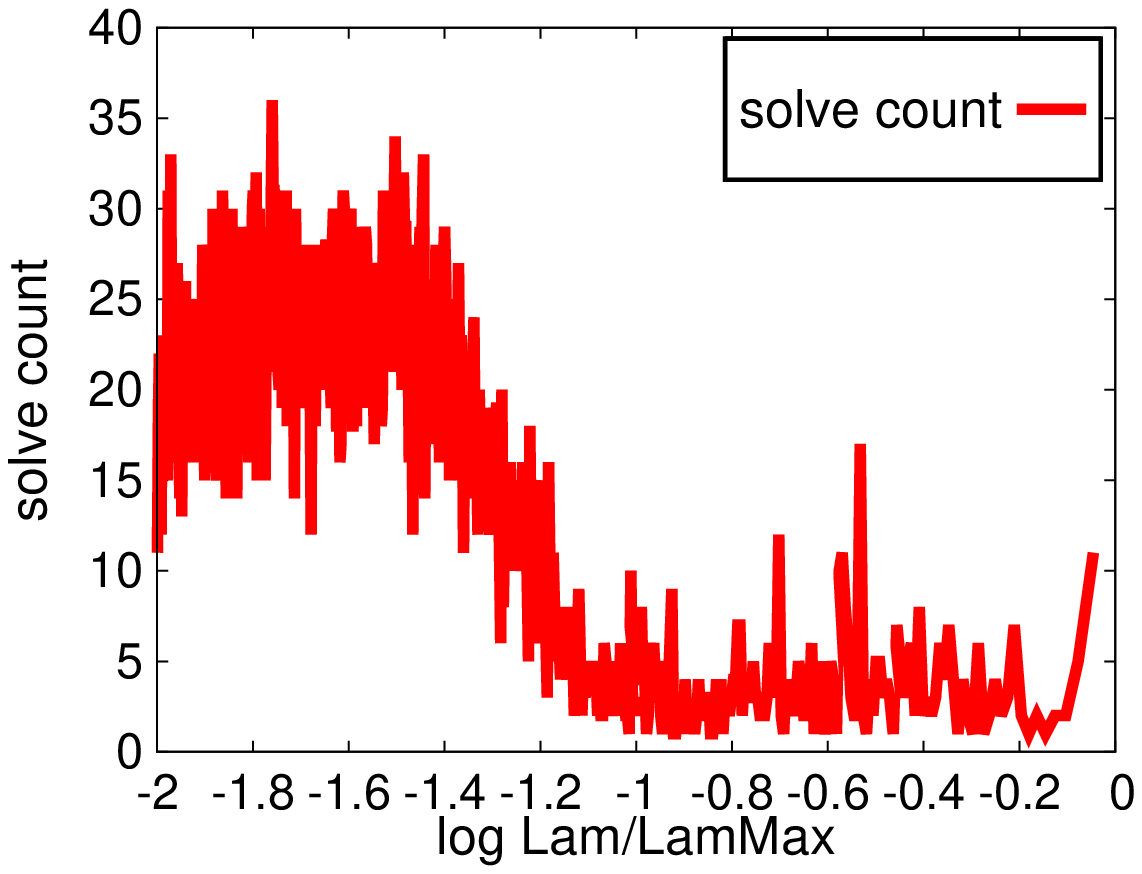}
&
\includegraphics[scale=0.5]{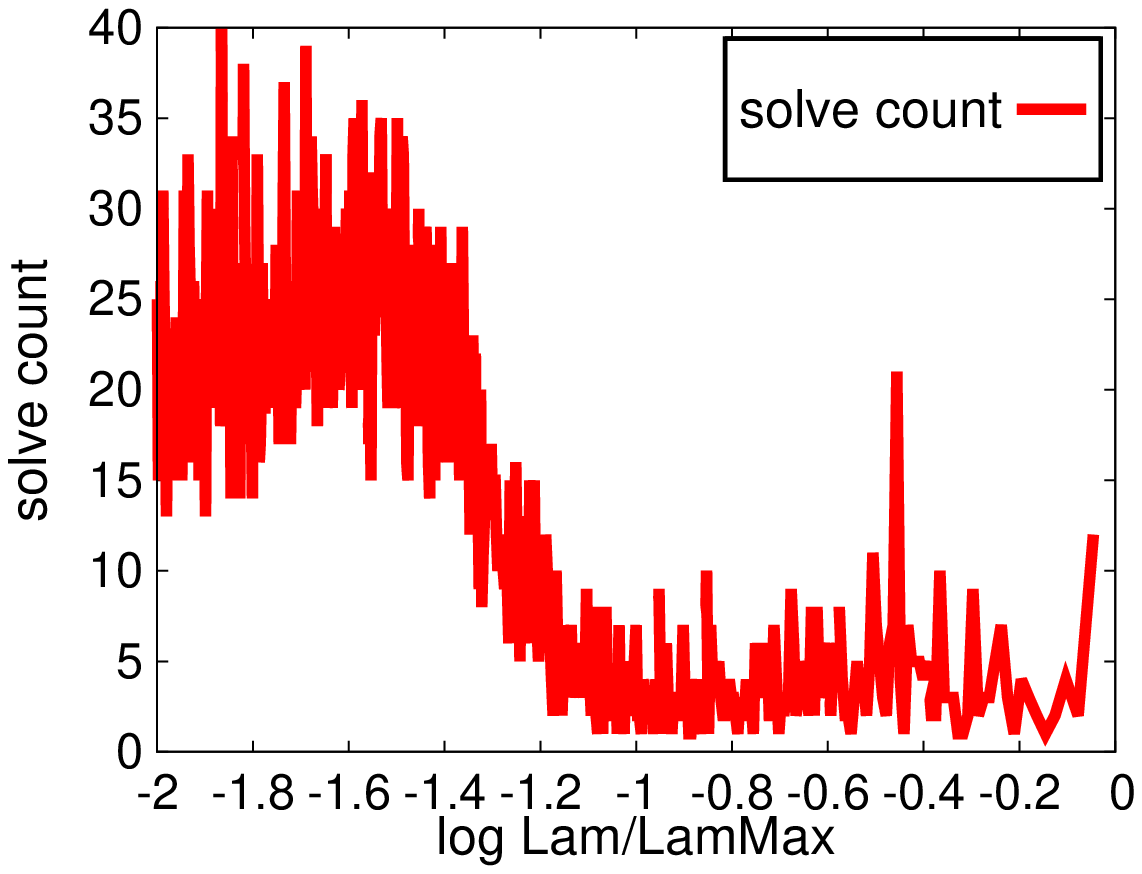}
\\
\multicolumn{2}{c}{(d) The number of solving LASSO in IB}
\\
\includegraphics[scale=0.5]{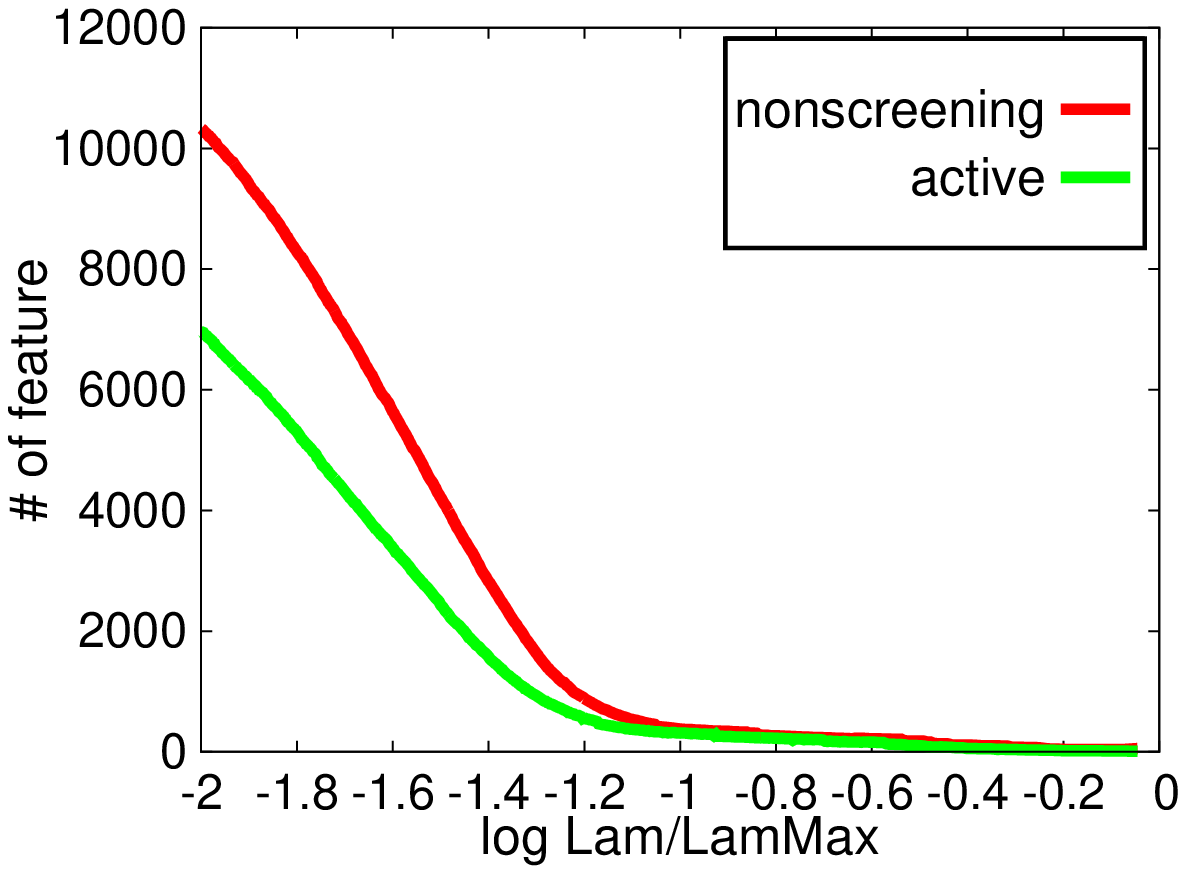}
&
\includegraphics[scale=0.5]{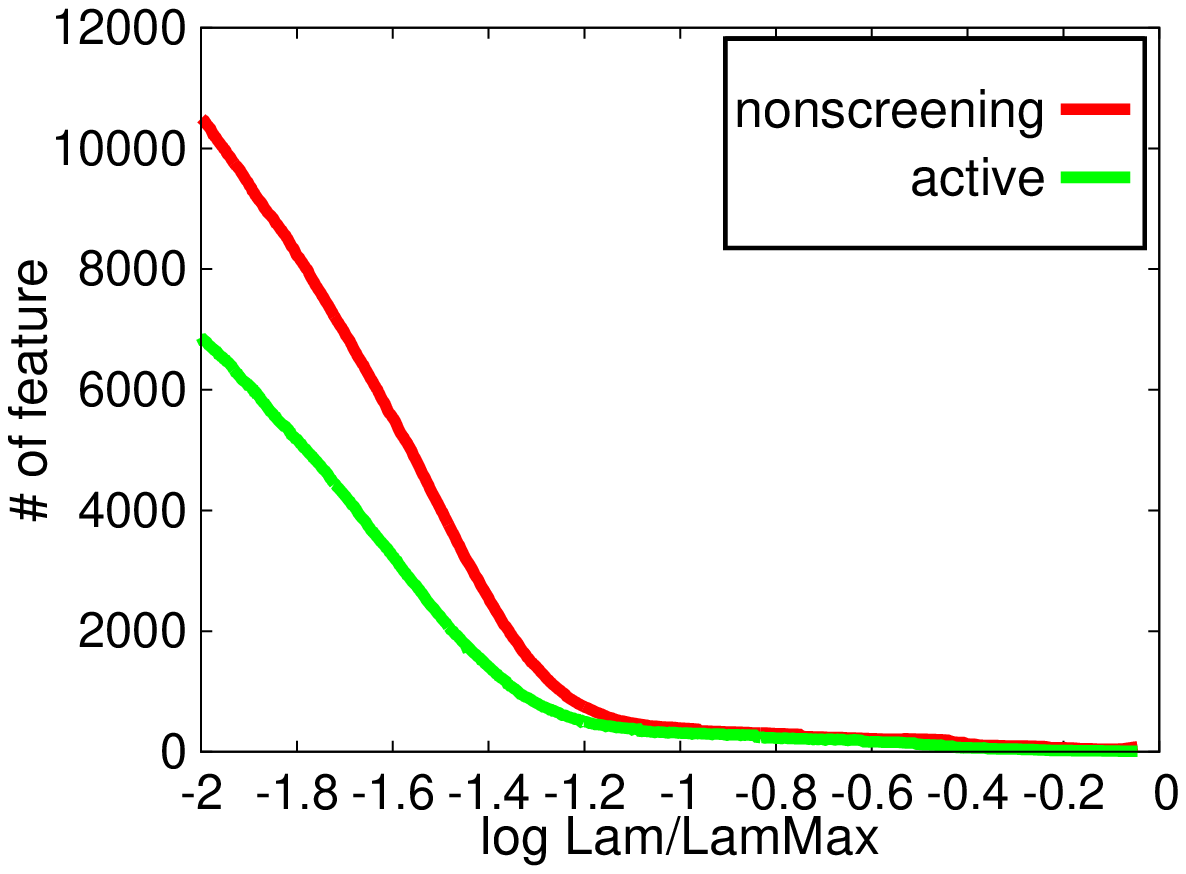}
\\
\multicolumn{2}{c}{(e) The number of non-screened out features and total active features}
\\
\includegraphics[scale=0.5]{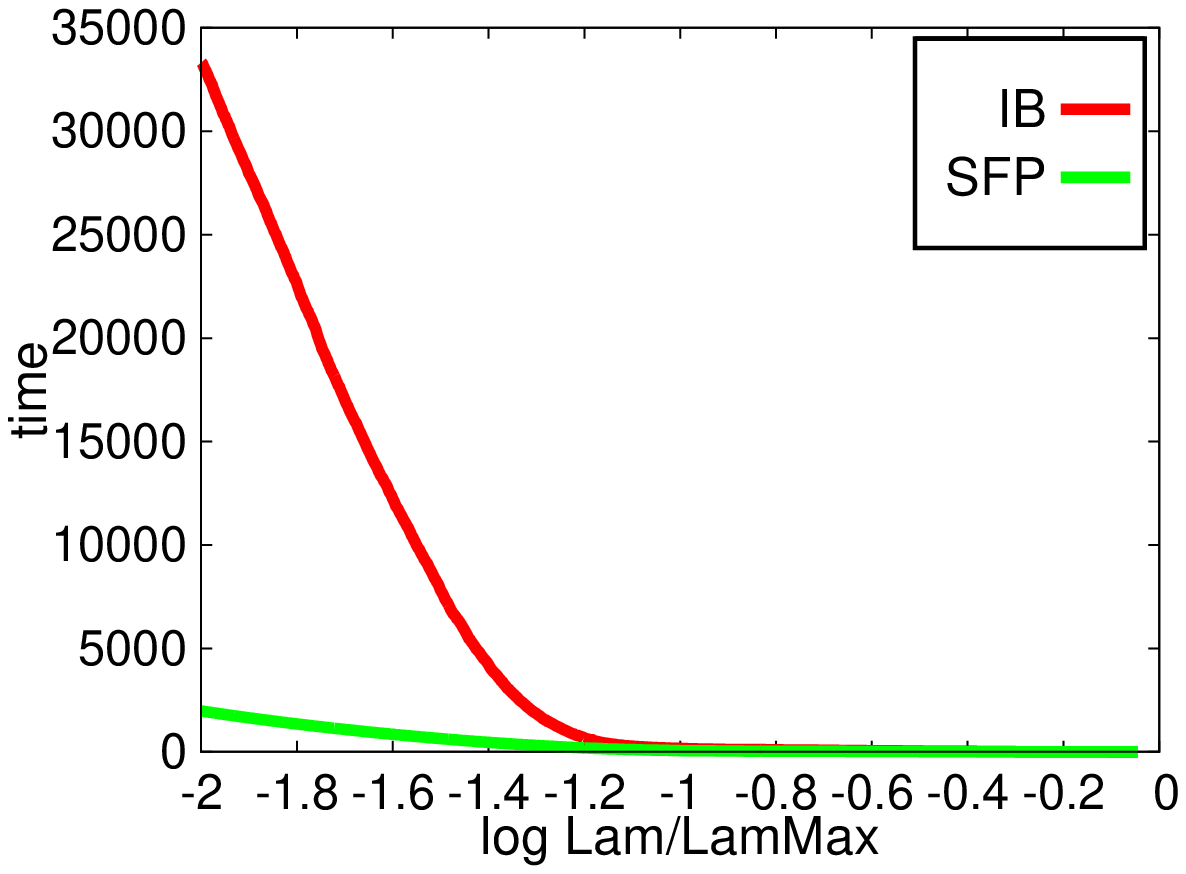}
&
\includegraphics[scale=0.5]{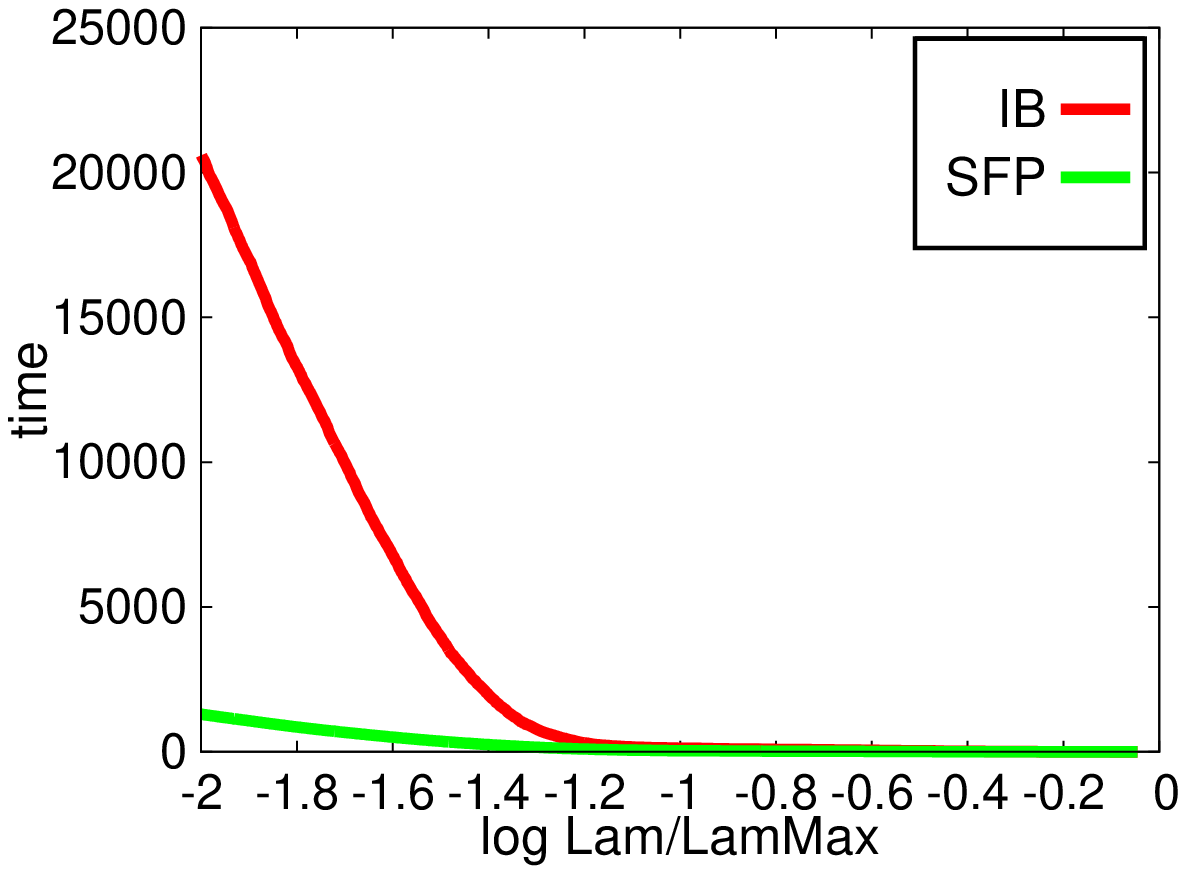}
\\
\multicolumn{2}{c}{(f) Computation total time in seconds}
\end{tabular}
\end{center}
\end{figure}

\clearpage

\subsection{Results on mnist}
\begin{figure}[ht]
\begin{center}
\begin{tabular}{cc}
\includegraphics[scale=0.5]{./mnist_time_15.eps}
&
\includegraphics[scale=0.5]{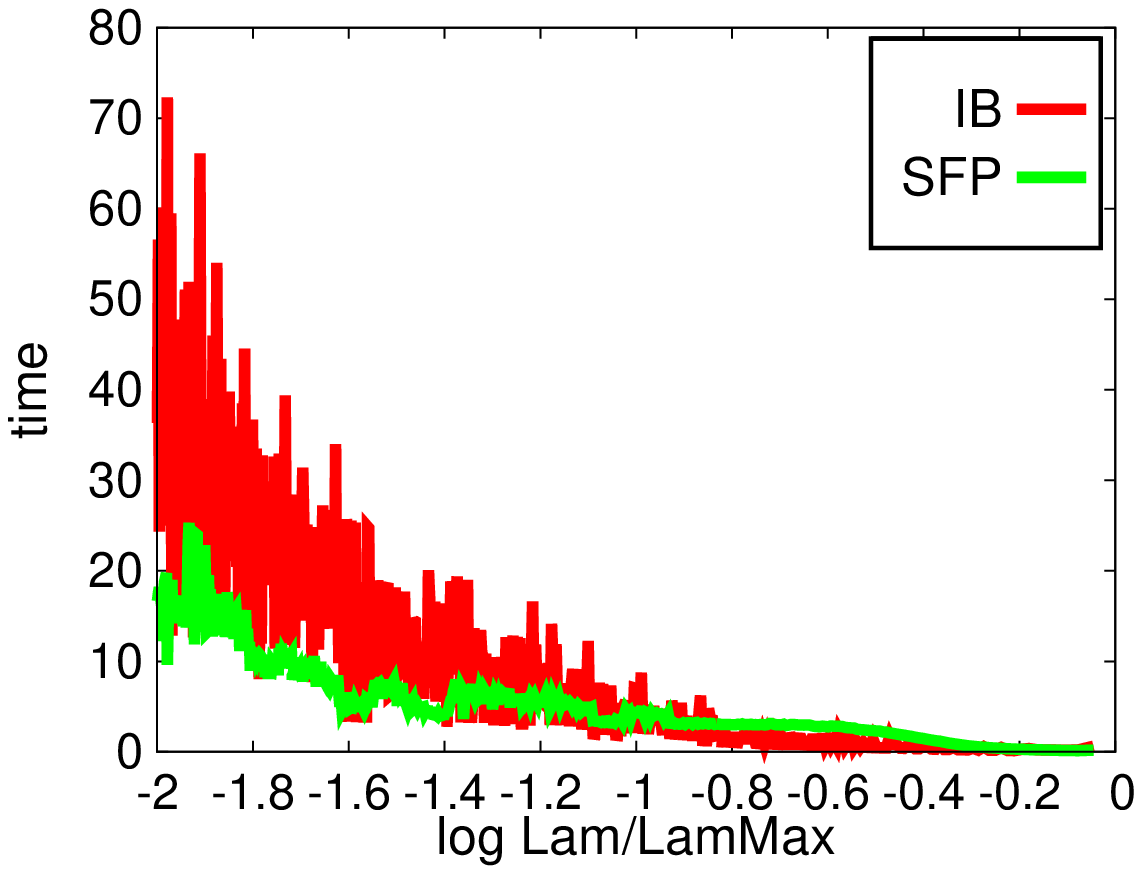}
\\
\multicolumn{2}{c}{(a) Computation time in seconds}
\\
\includegraphics[scale=0.5]{./mnist_trav_15.eps}
&
\includegraphics[scale=0.5]{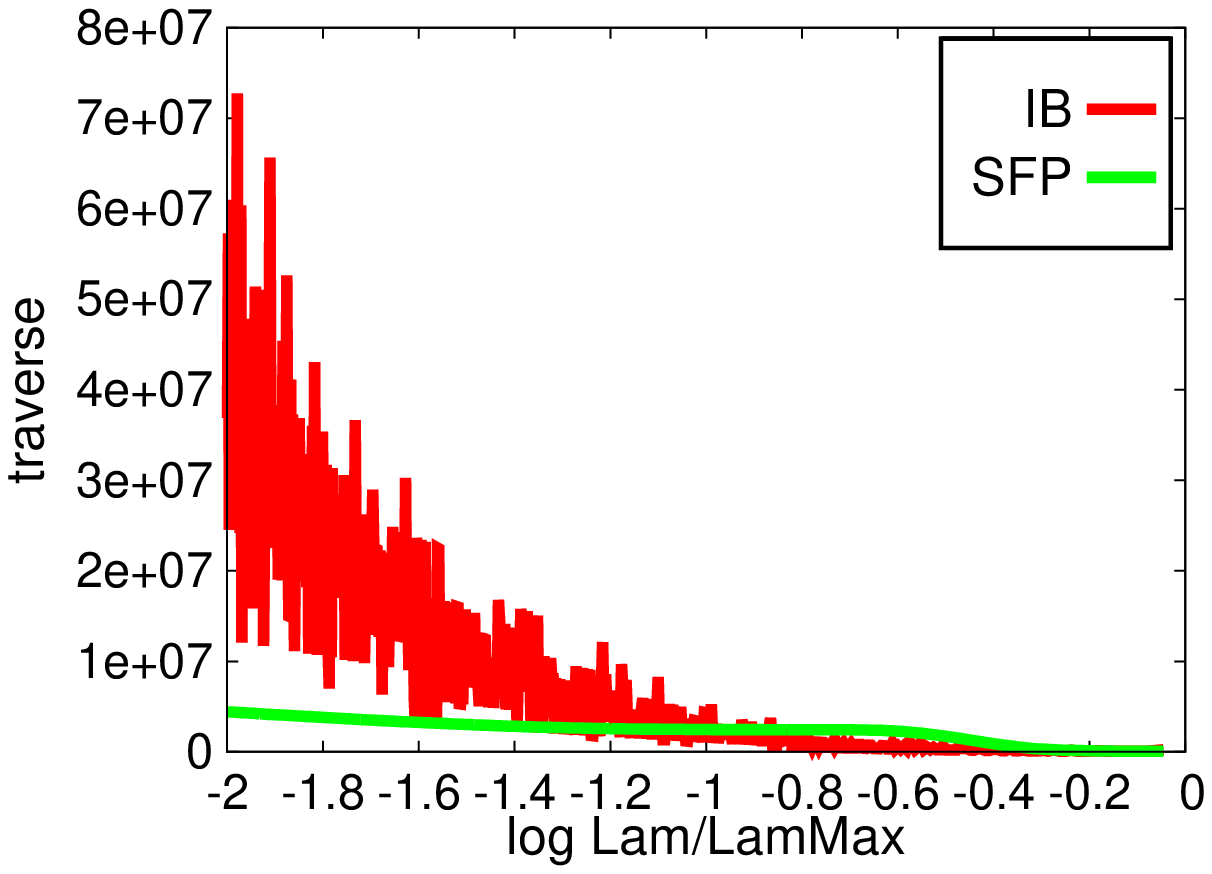}
\\
\multicolumn{2}{c}{(b) The number of traverse nodes}
\\
\includegraphics[scale=0.5]{./mnist_feature_15.eps}
&
\includegraphics[scale=0.5]{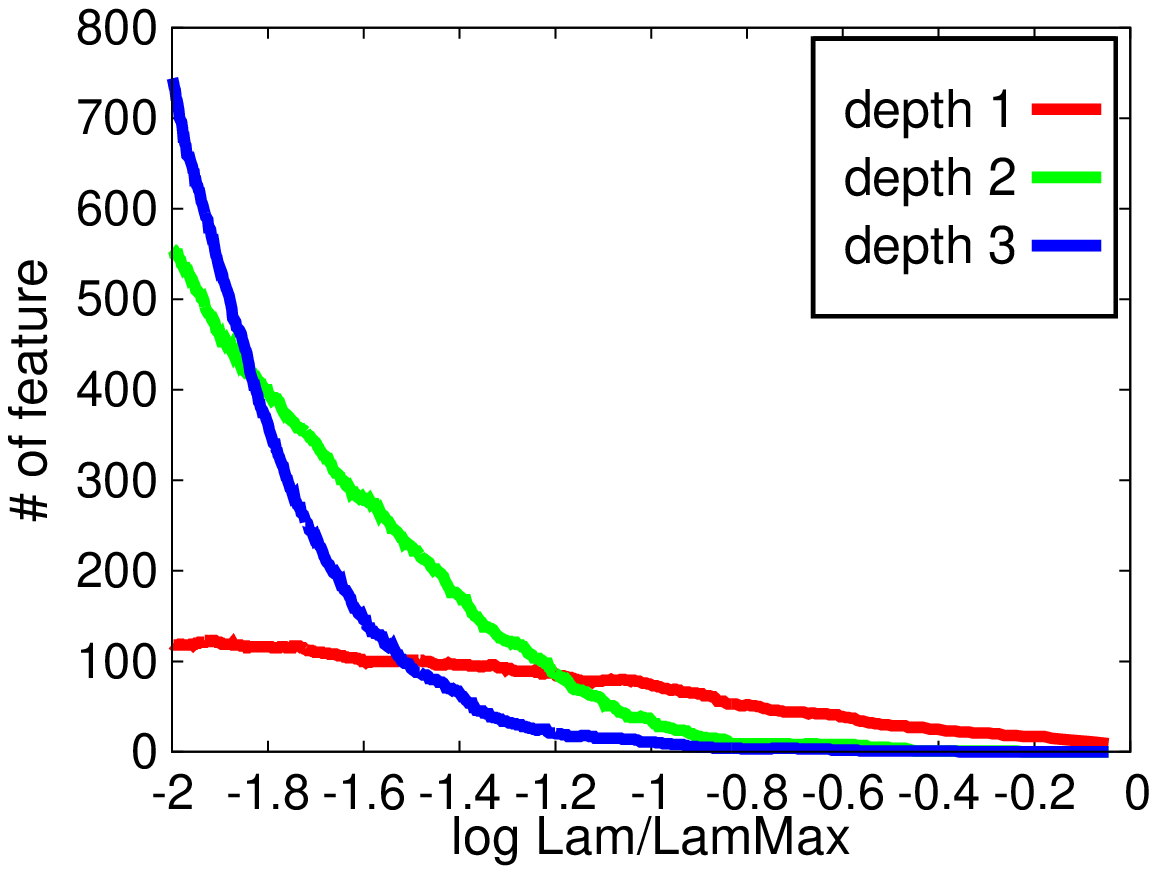}
\\
\multicolumn{2}{c}{(c) The number of active features}
\end{tabular}
\end{center}
\end{figure}

\begin{figure}[ht]
\begin{center}
\begin{tabular}{cc}
\includegraphics[scale=0.5]{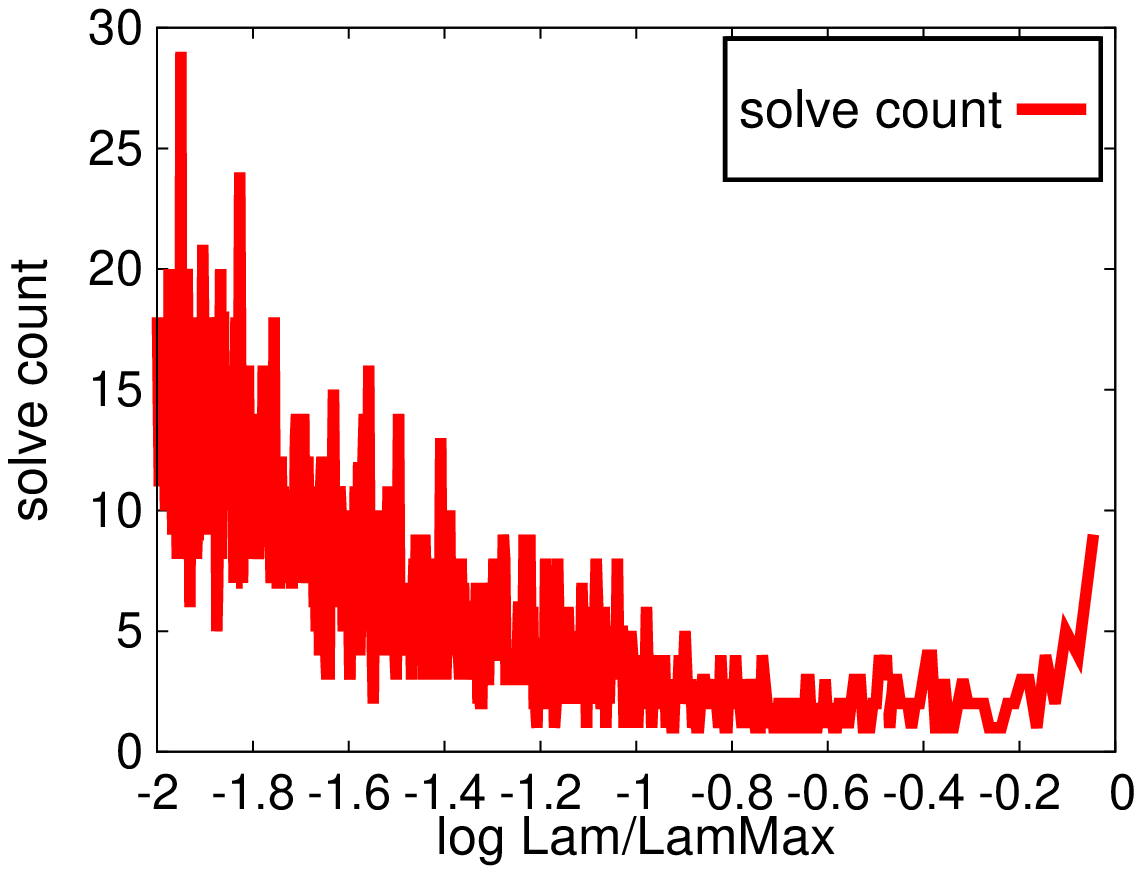}
&
\includegraphics[scale=0.5]{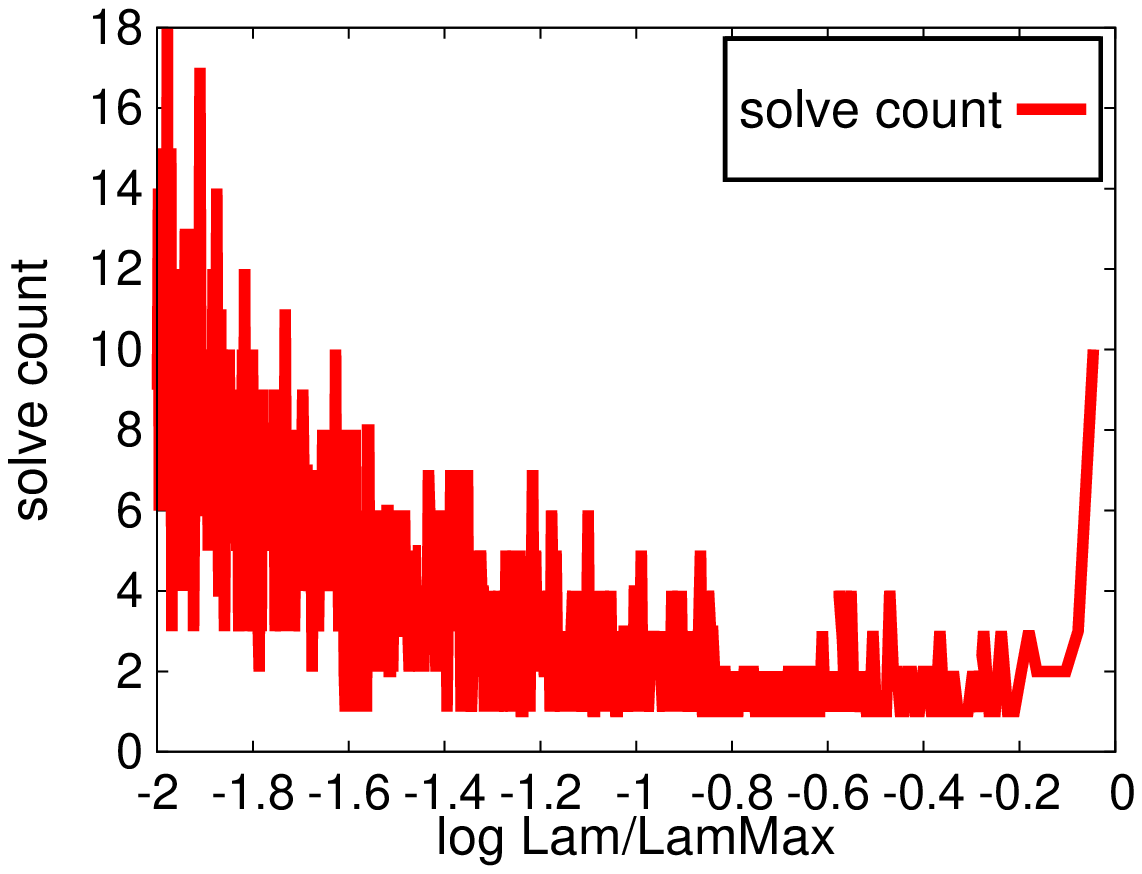}
\\
\multicolumn{2}{c}{(d) The number of solving LASSO in IB}
\\
\includegraphics[scale=0.5]{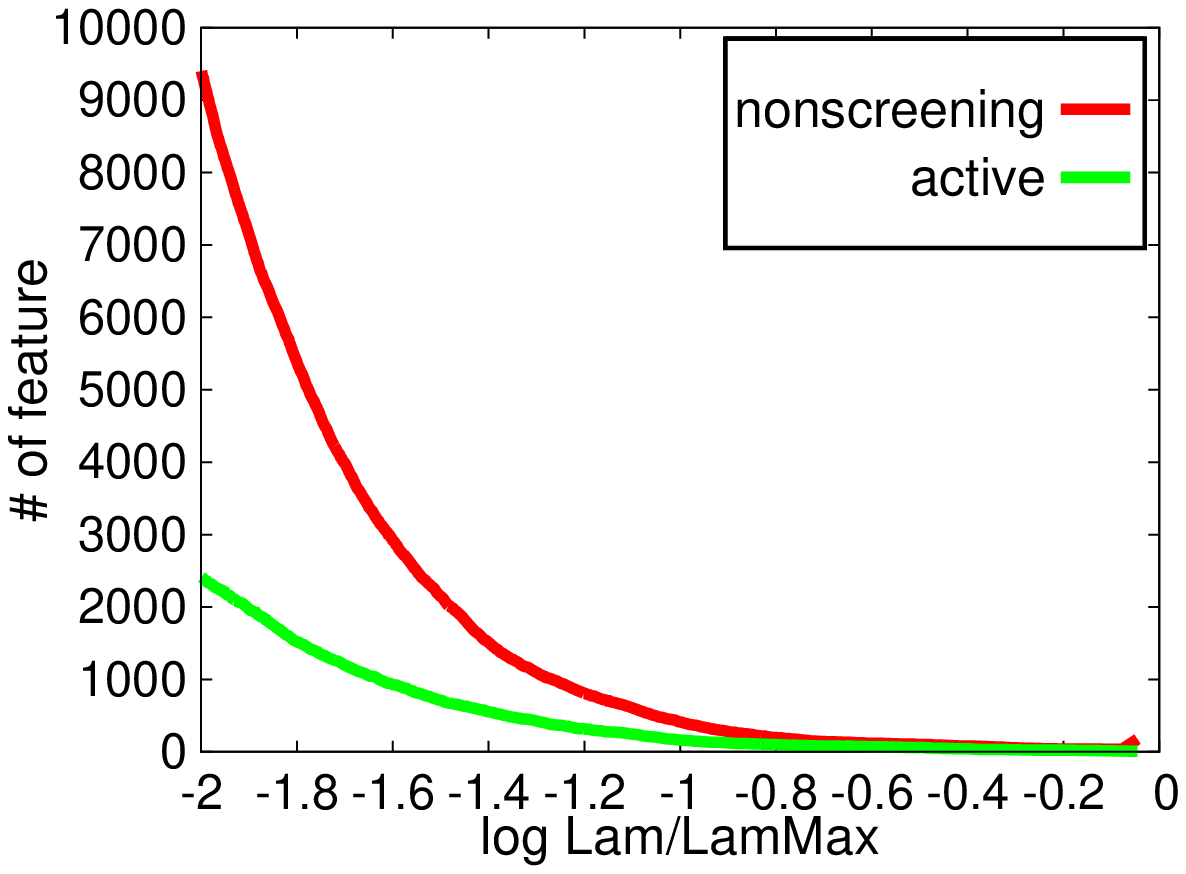}
&
\includegraphics[scale=0.5]{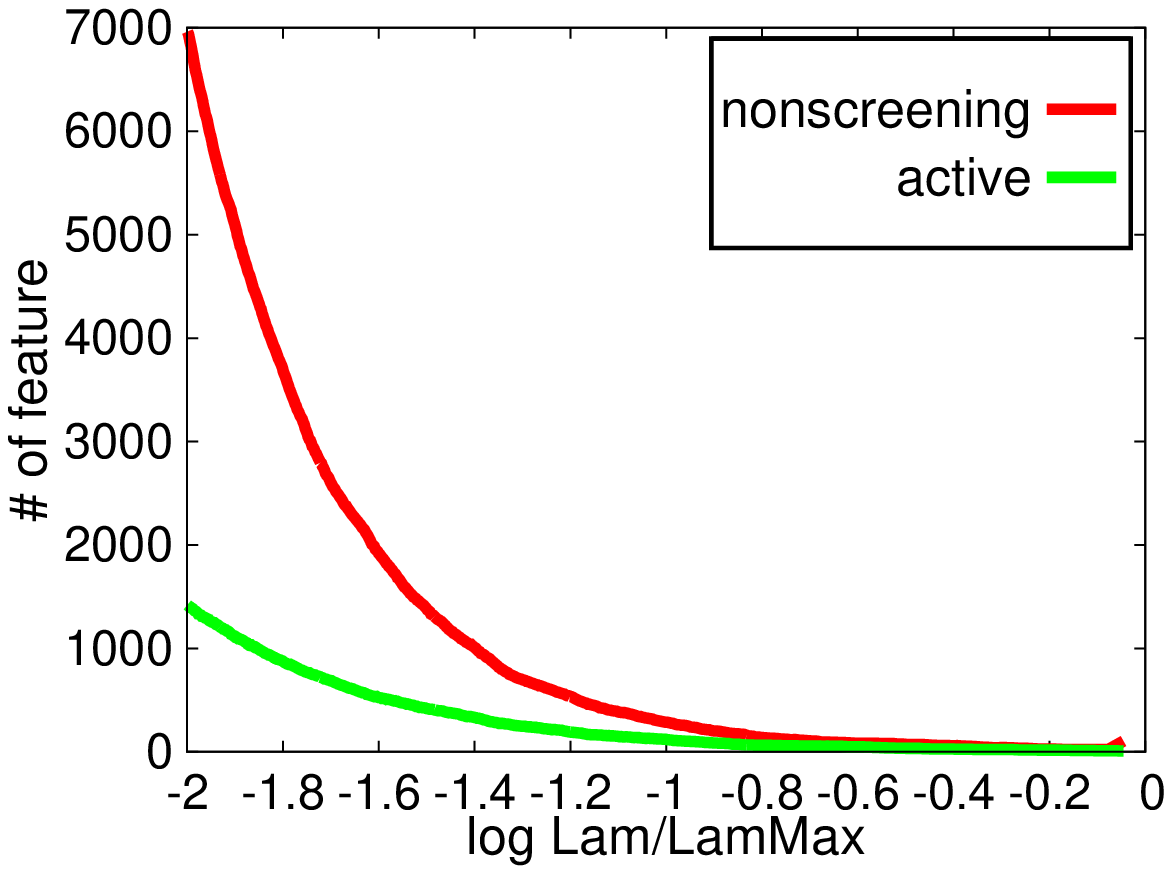}
\\
\multicolumn{2}{c}{(e) The number of non-screened out features and total active features}
\\
\includegraphics[scale=0.5]{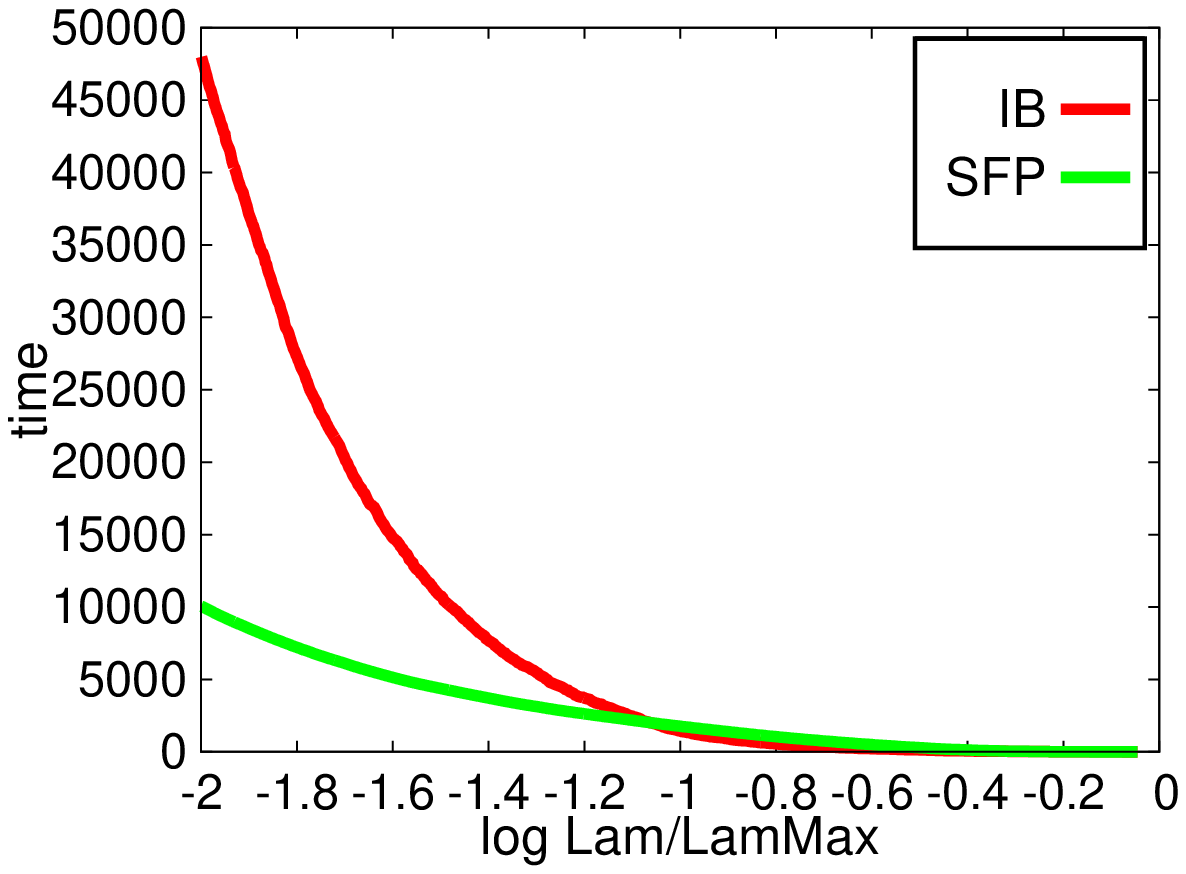}
&
\includegraphics[scale=0.5]{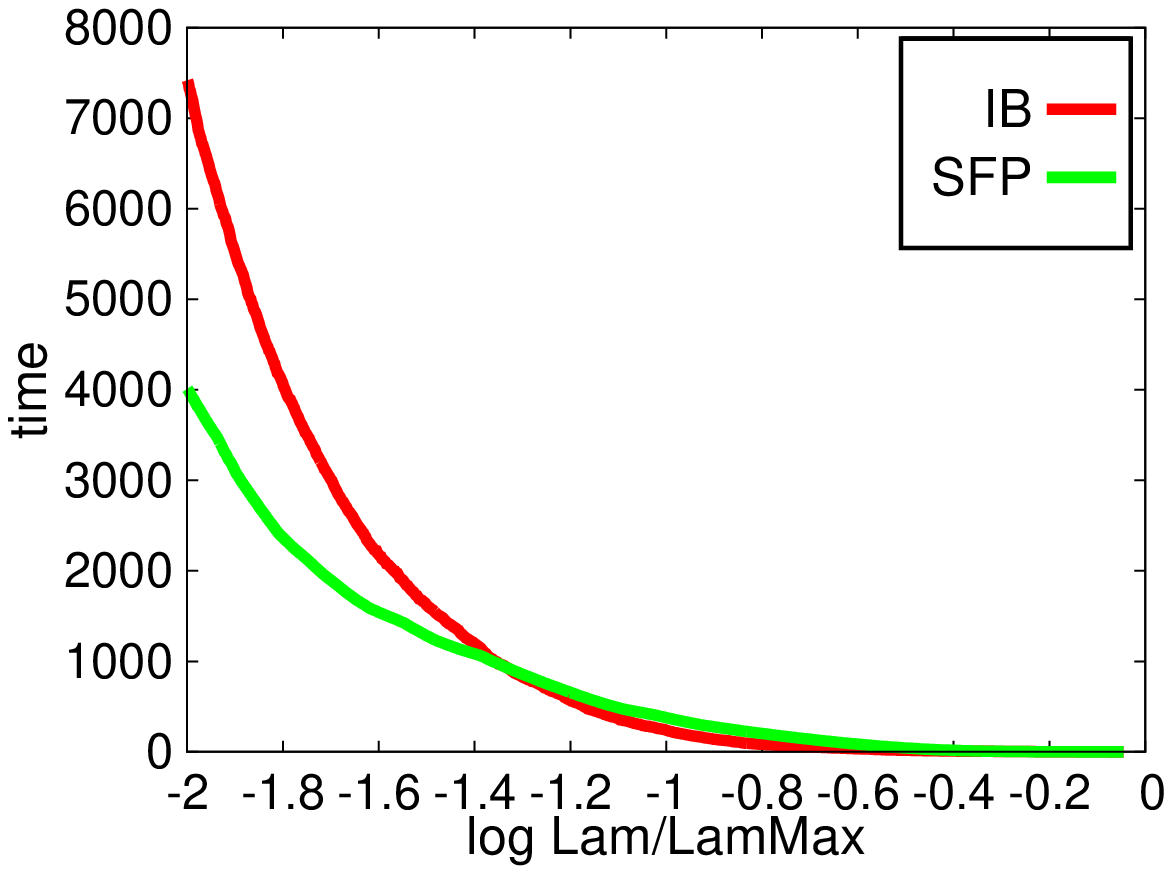}
\\
\multicolumn{2}{c}{(f) Computation total time in seconds}
\end{tabular}
\end{center}
\end{figure}

\clearpage

\subsection{Results on rcv1\_binary}
\begin{figure}[ht]
\begin{center}
\begin{tabular}{cc}
\includegraphics[scale=0.5]{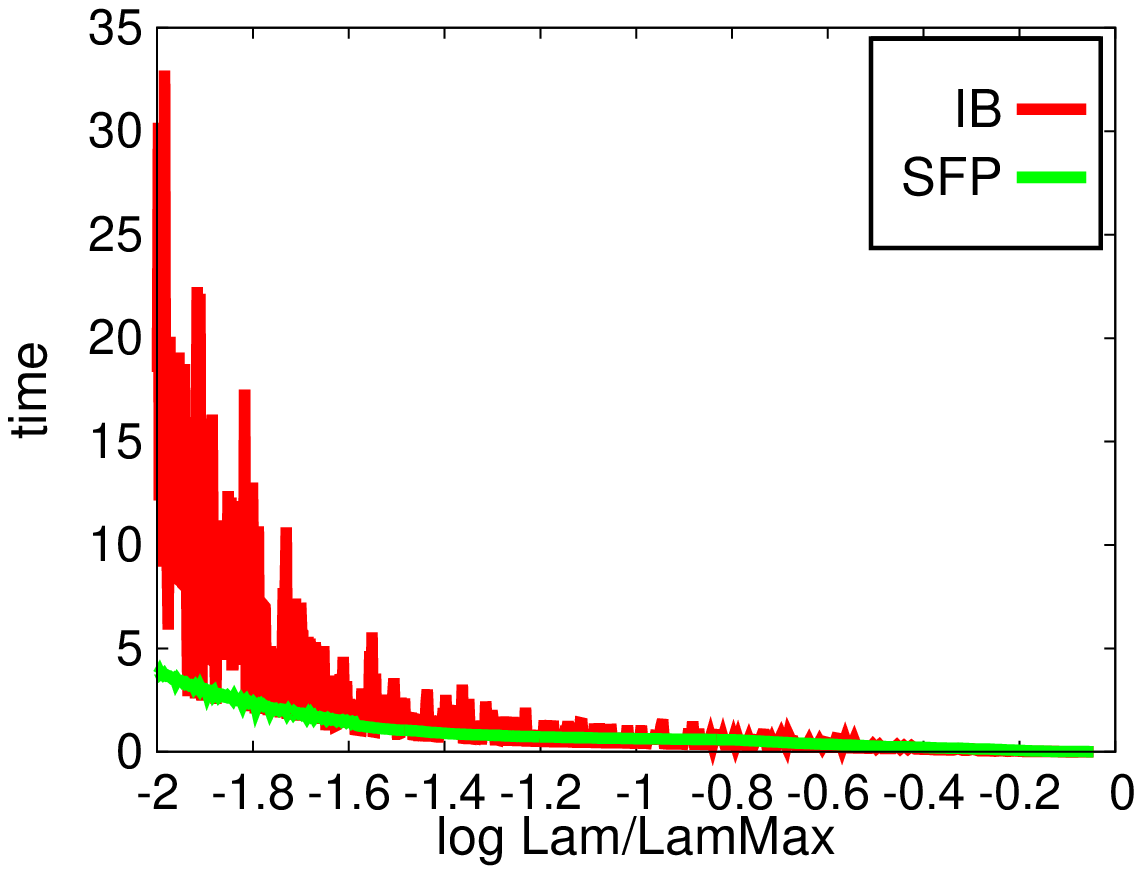}
&
\includegraphics[scale=0.5]{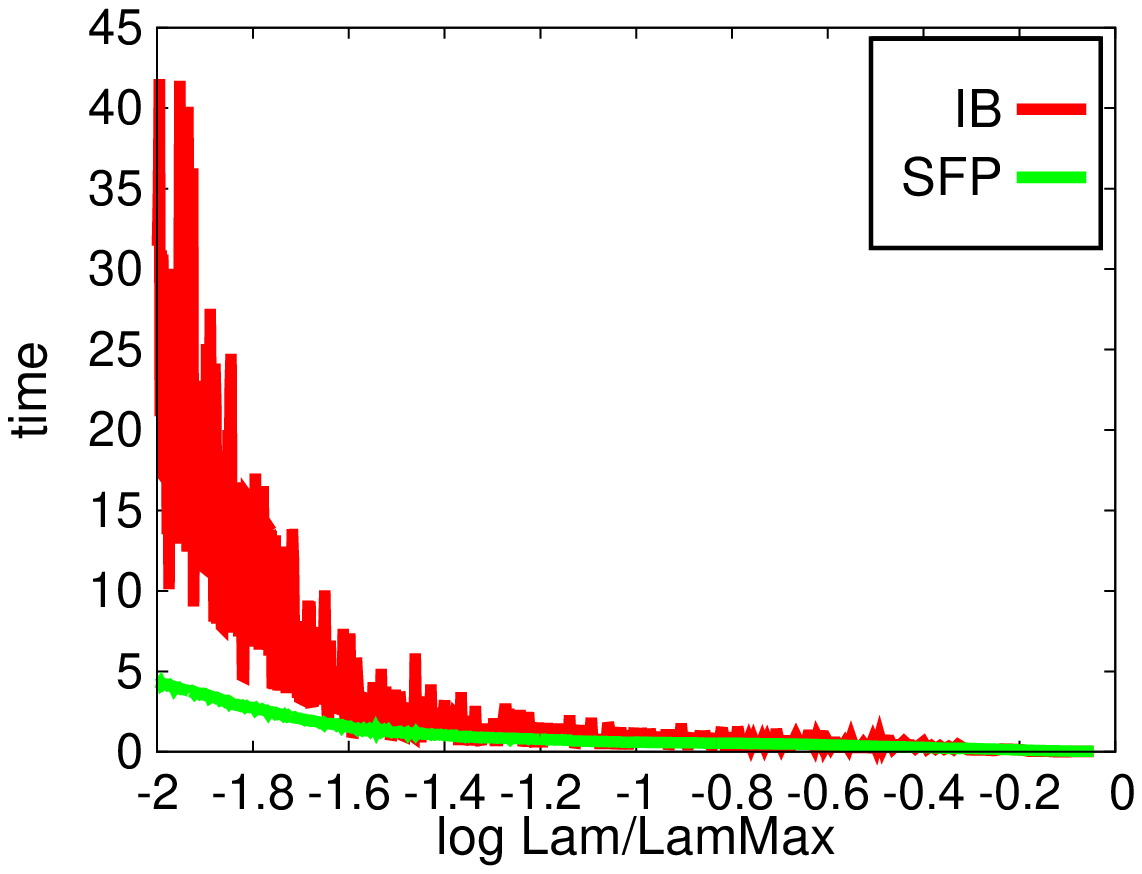}
\\
\multicolumn{2}{c}{(a) Computation time in seconds}
\\
\includegraphics[scale=0.5]{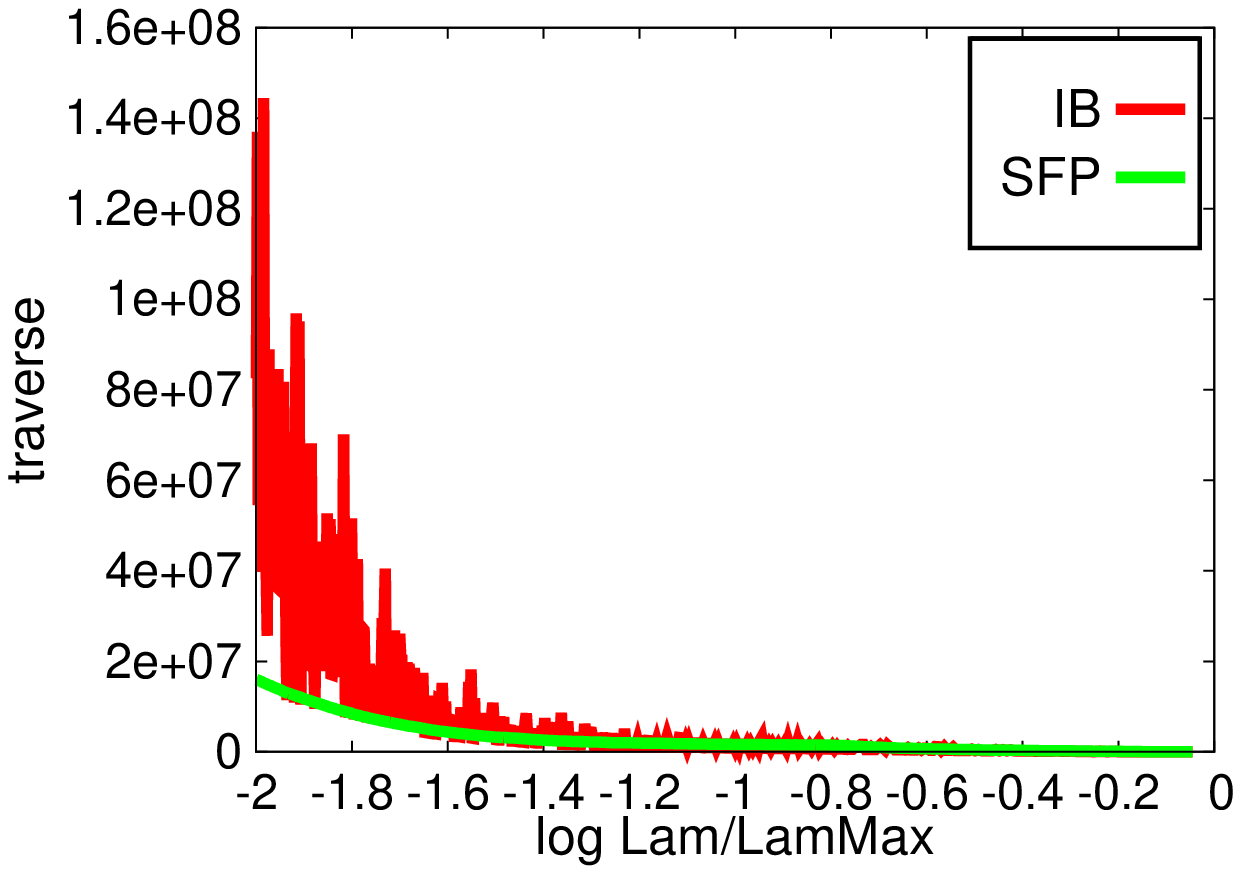}
&
\includegraphics[scale=0.5]{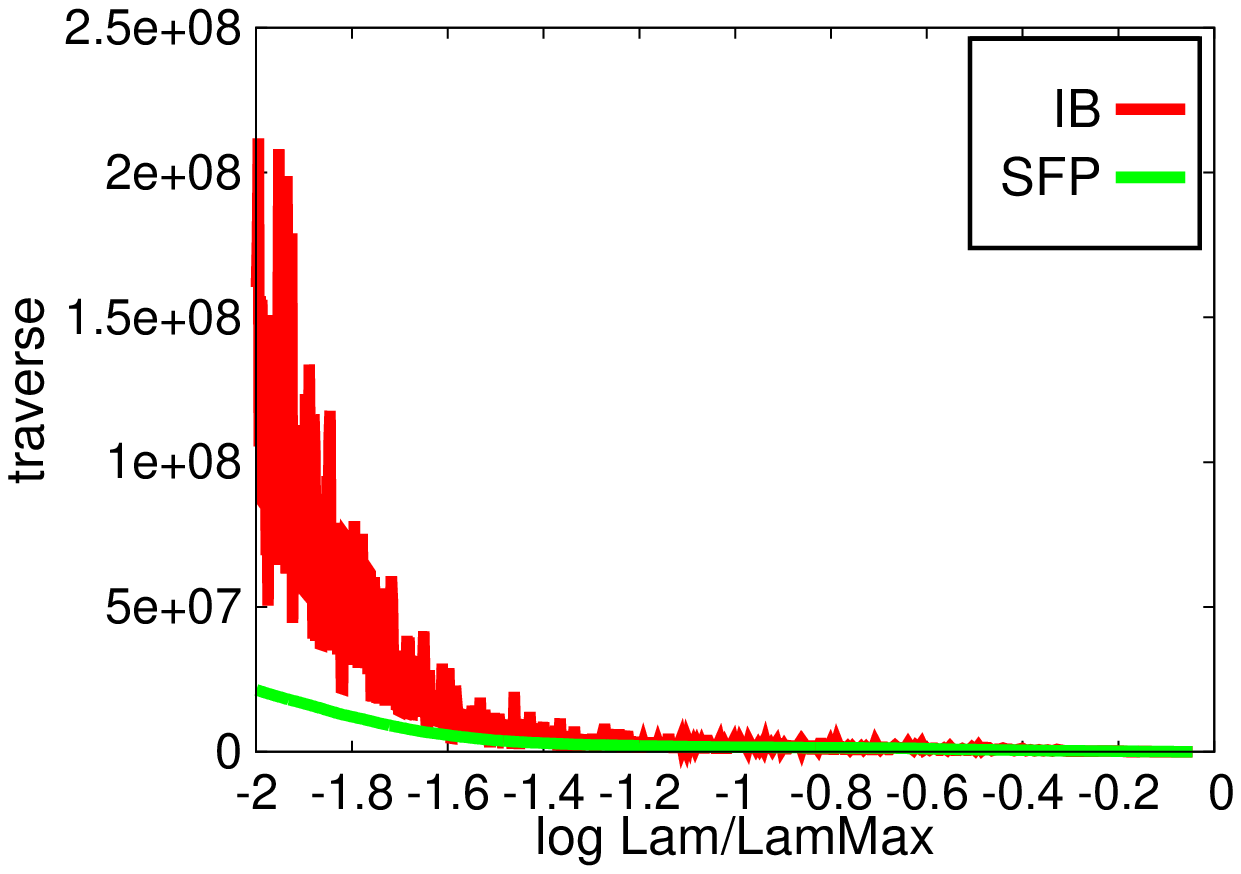}
\\
\multicolumn{2}{c}{(b) The number of traverse nodes}
\\
\includegraphics[scale=0.5]{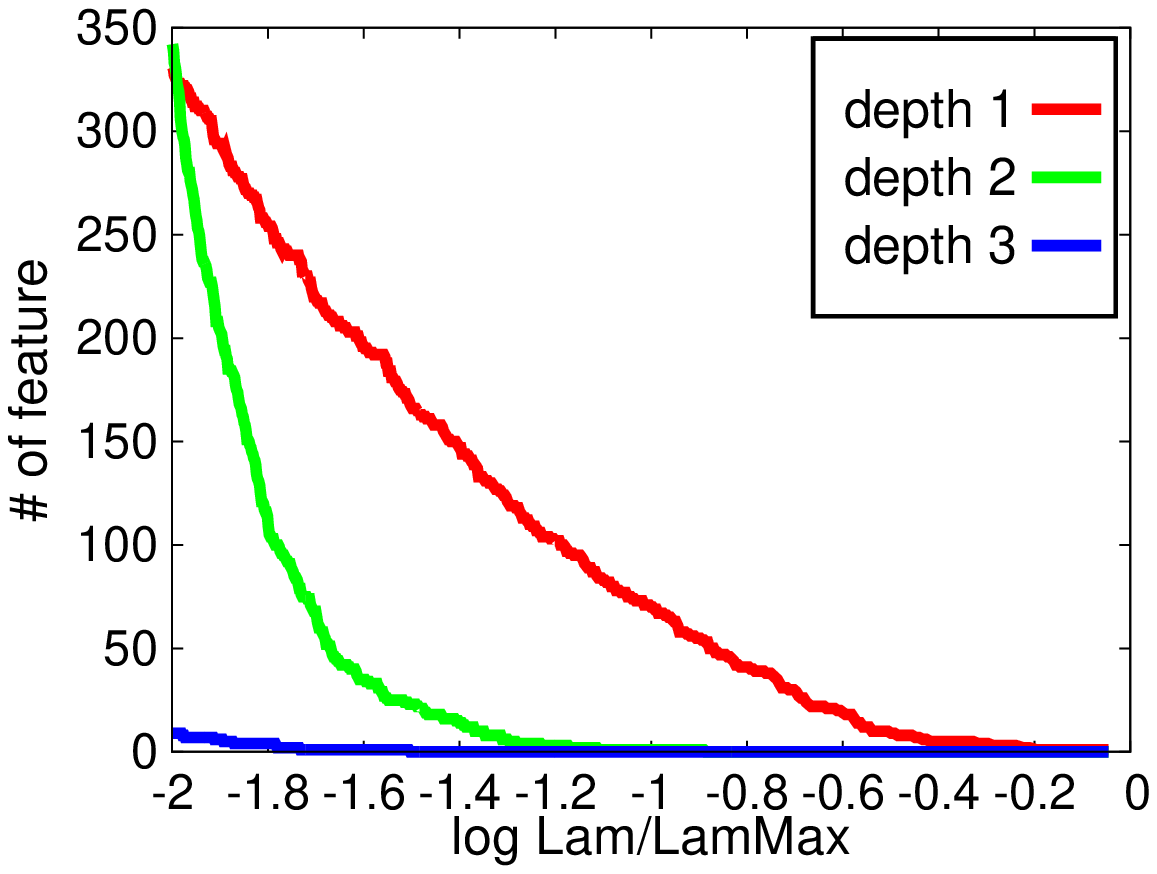}
&
\includegraphics[scale=0.5]{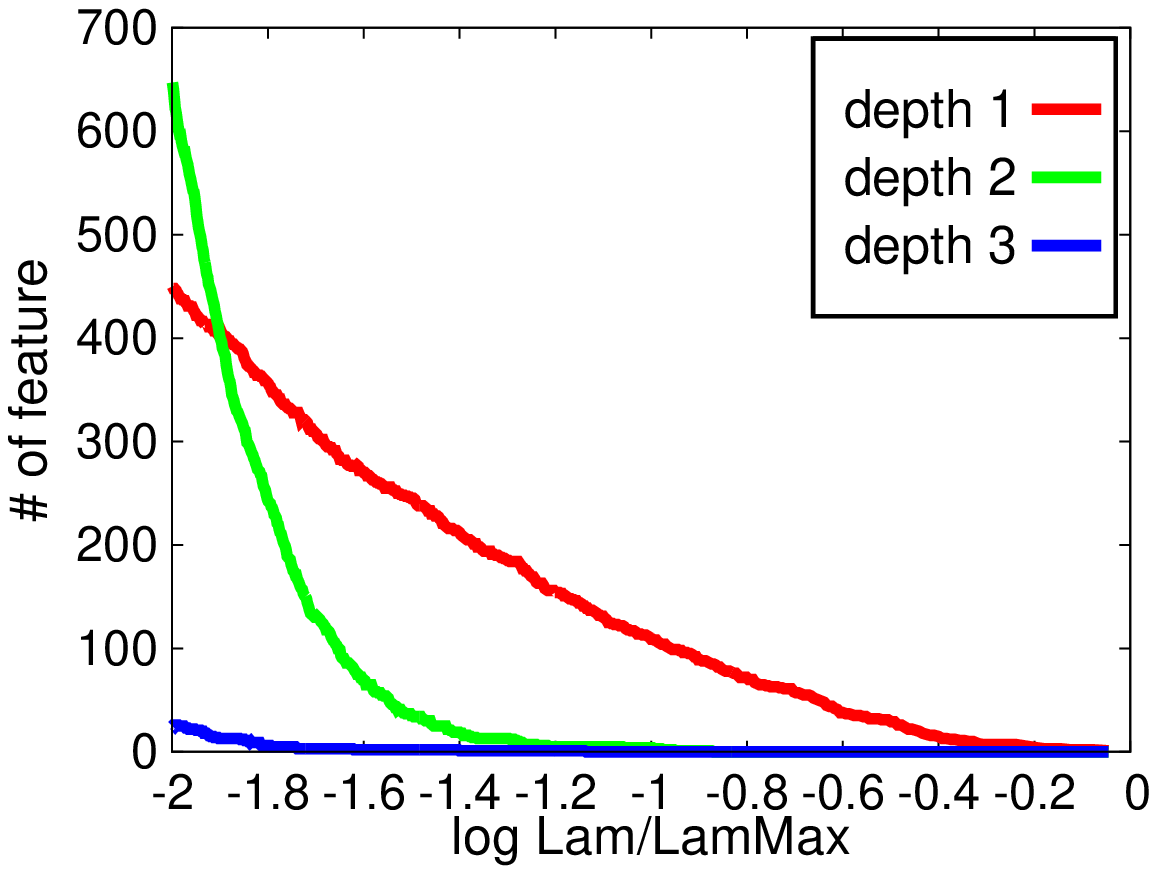}
\\
\multicolumn{2}{c}{(c) The number of active features}
\end{tabular}
\end{center}
\end{figure}

\begin{figure}[ht]
\begin{center}
\begin{tabular}{cc}
\includegraphics[scale=0.5]{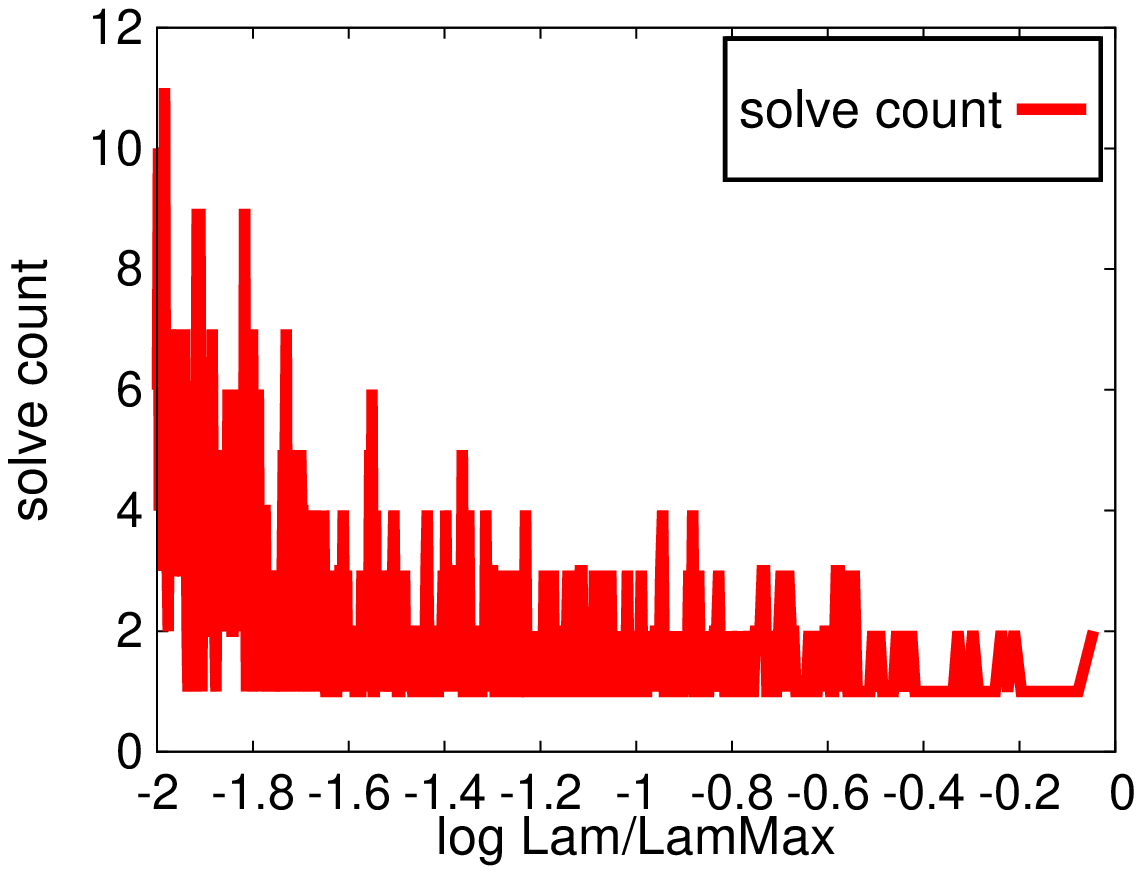}
&
\includegraphics[scale=0.5]{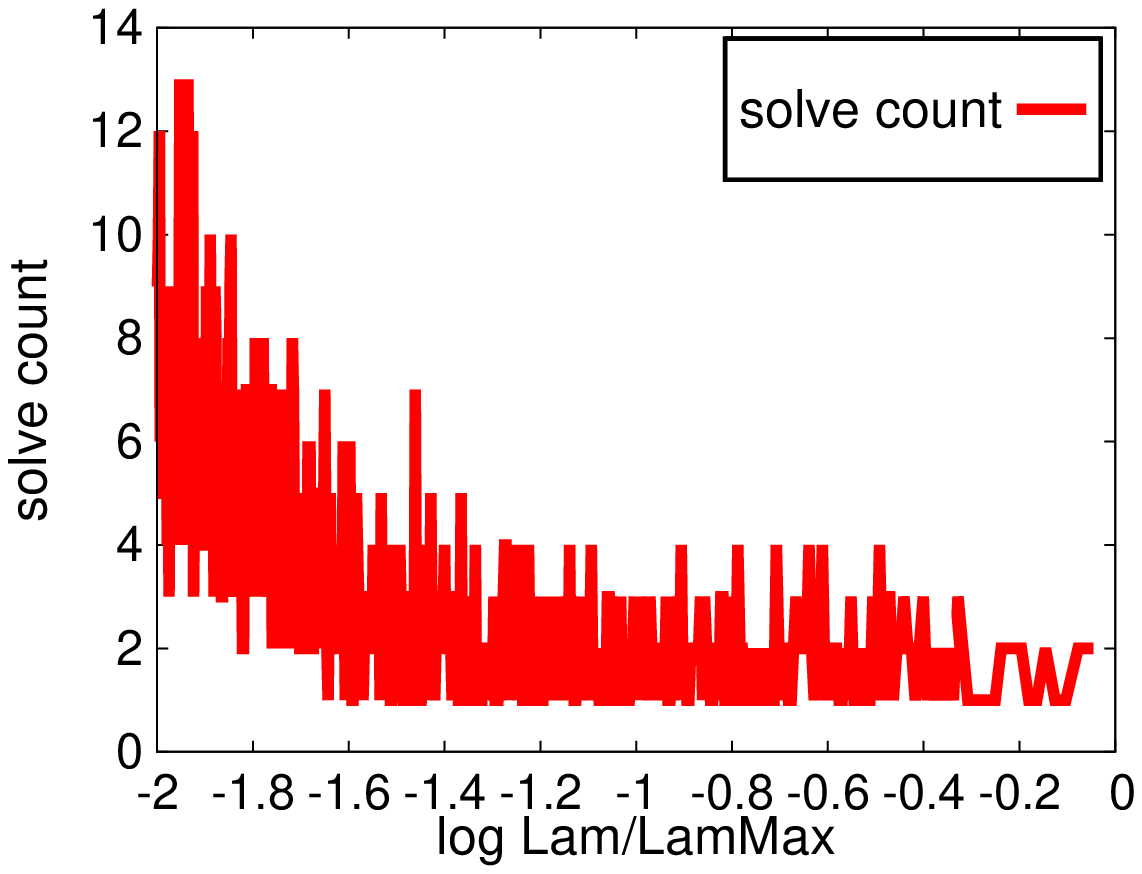}
\\
\multicolumn{2}{c}{(d) The number of solving LASSO in IB}
\\
\includegraphics[scale=0.5]{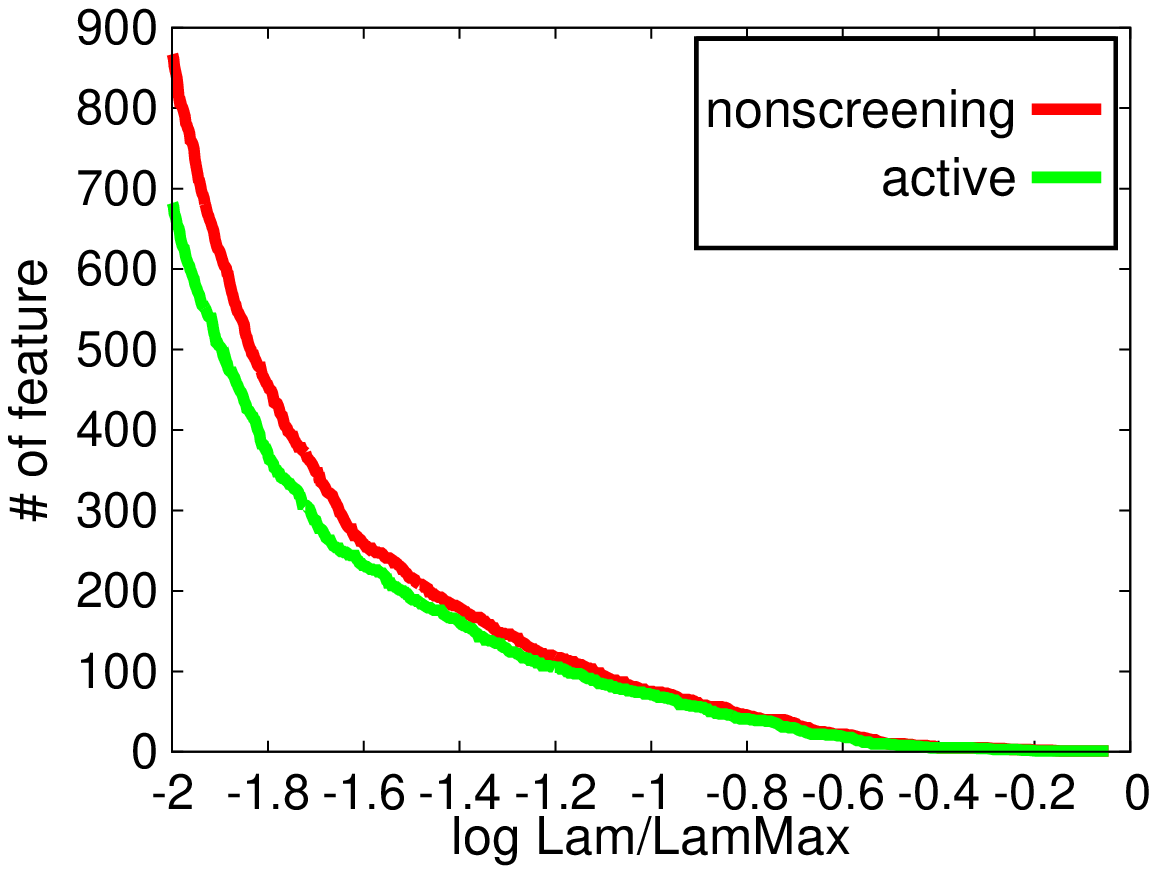}
&
\includegraphics[scale=0.5]{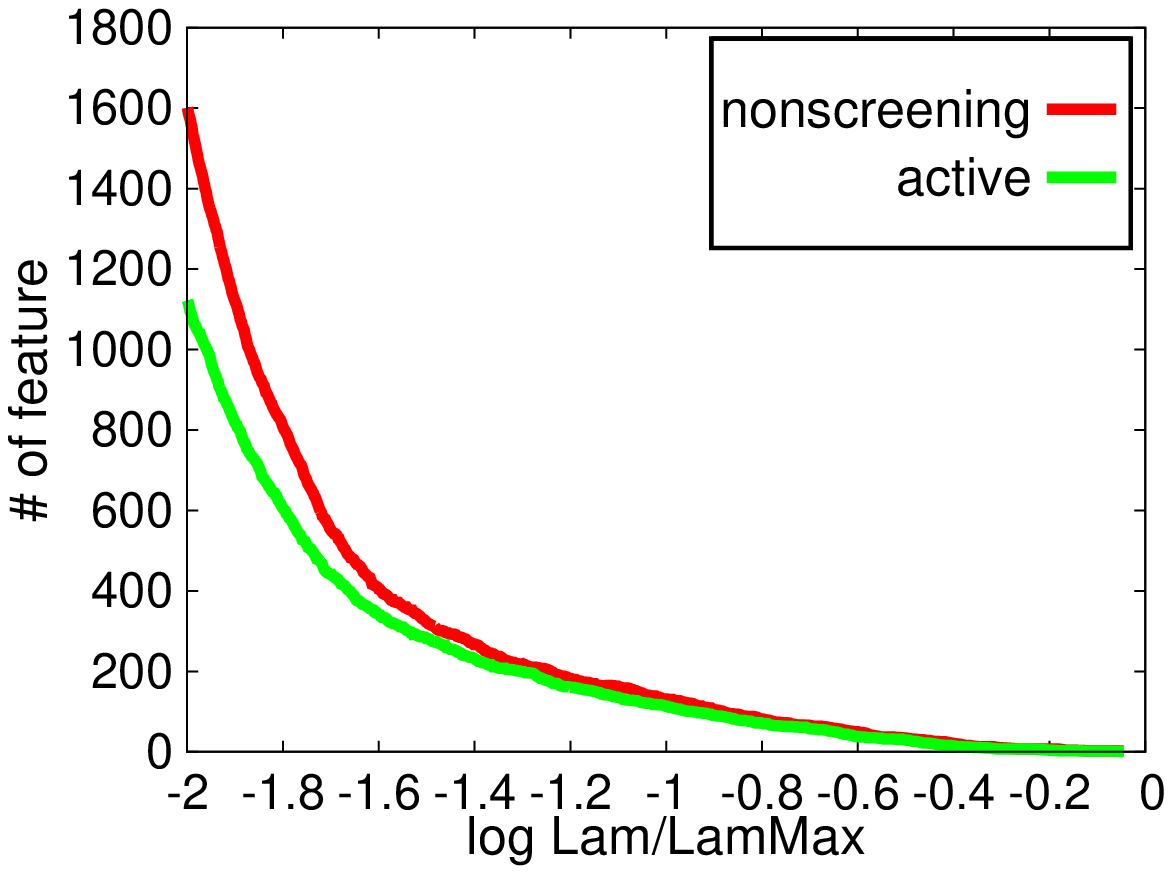}
\\
\multicolumn{2}{c}{(e) The number of non-screened out features and total active features}
\\
\includegraphics[scale=0.5]{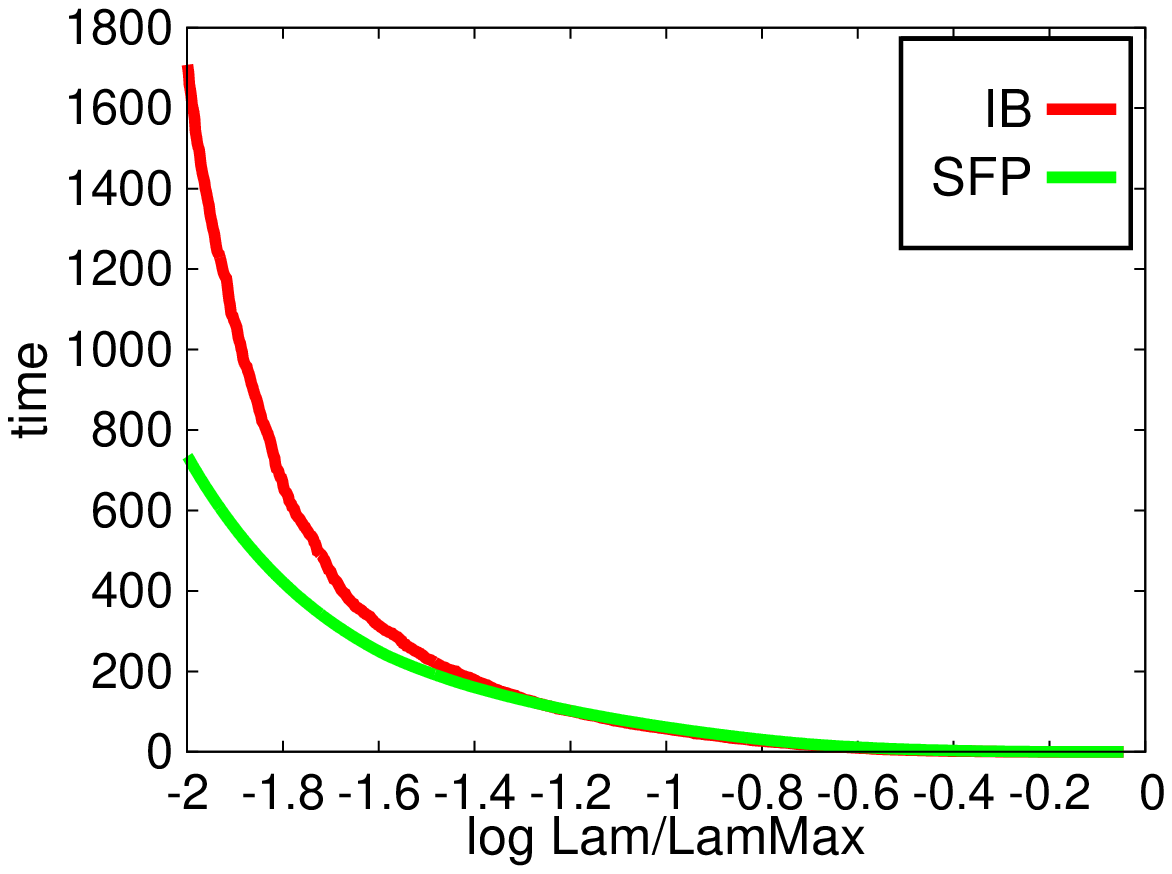}
&
\includegraphics[scale=0.5]{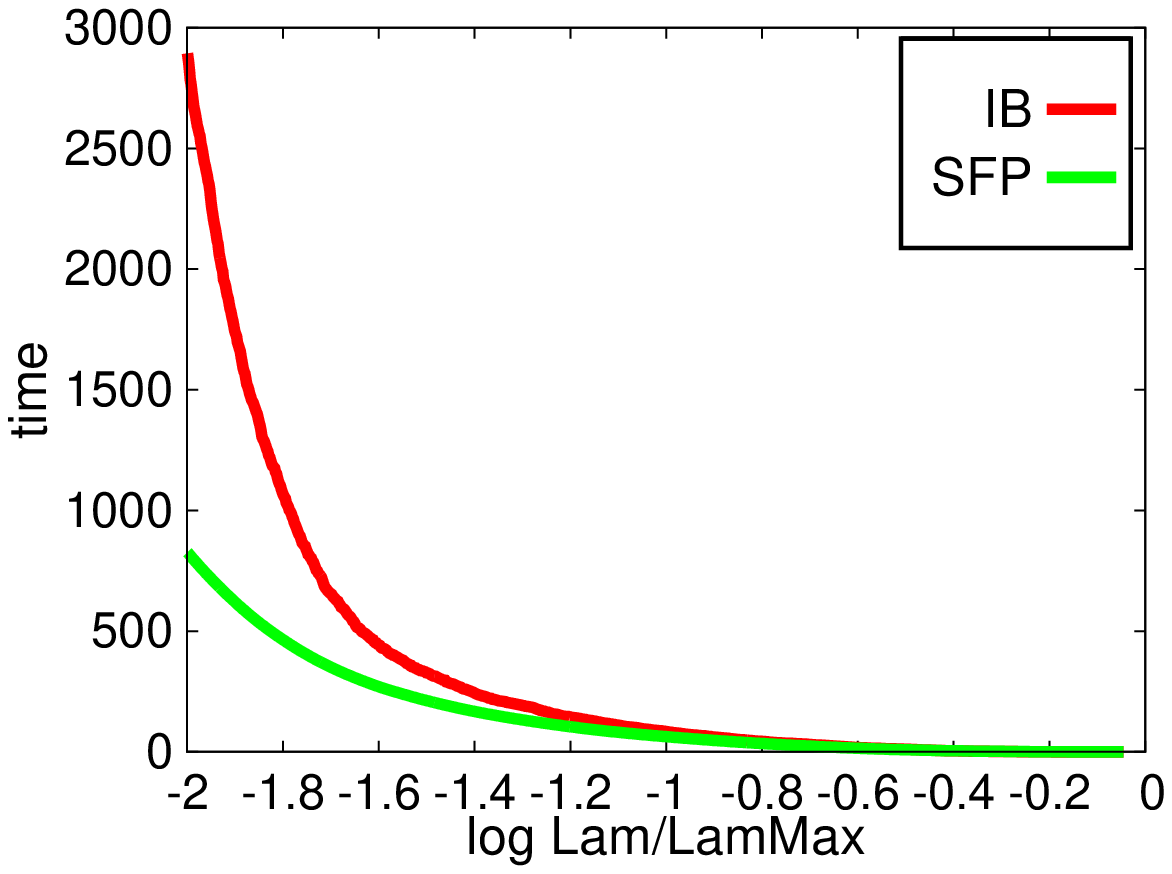}
\\
\multicolumn{2}{c}{(f) Computation total time in seconds}
\end{tabular}
\end{center}
\end{figure}

\clearpage

\subsection{Results on real-sim}
\begin{figure}[ht]
\begin{center}
\begin{tabular}{cc}
\includegraphics[scale=0.5]{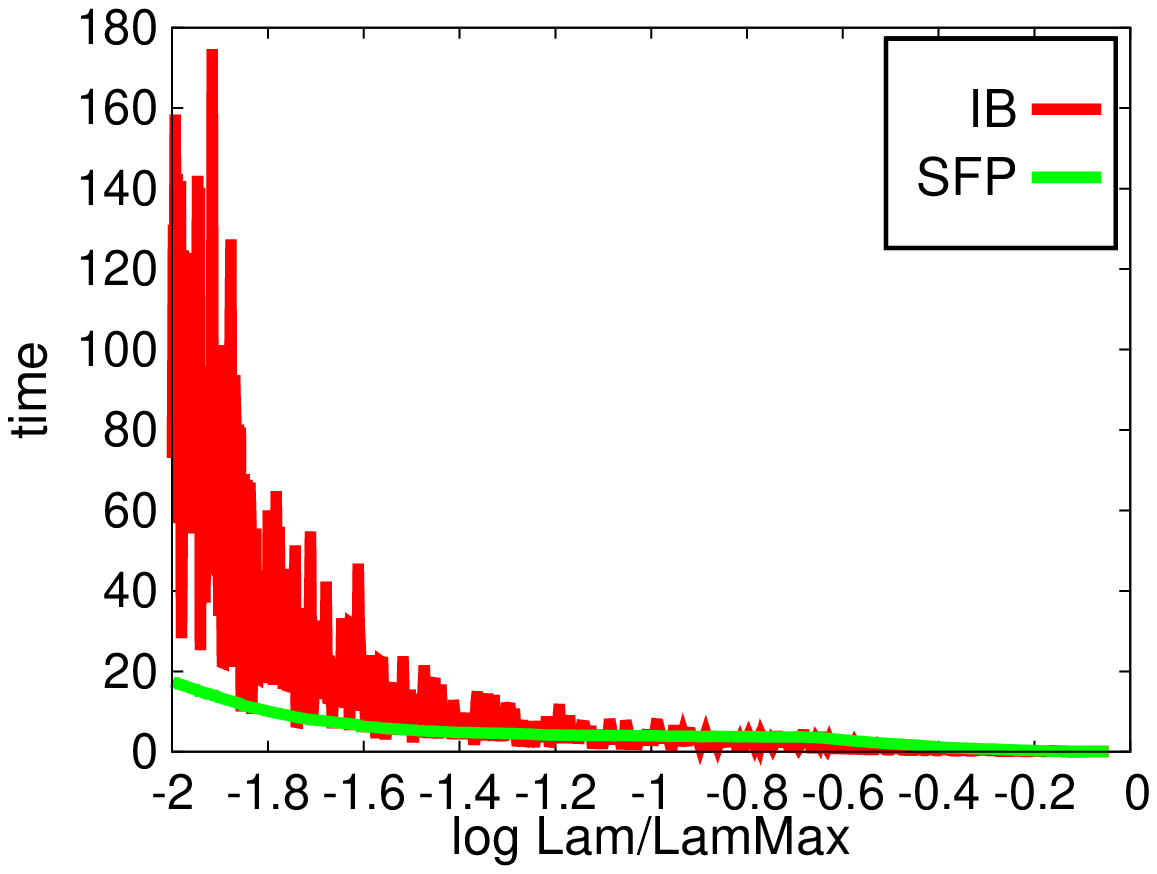}
&
\includegraphics[scale=0.5]{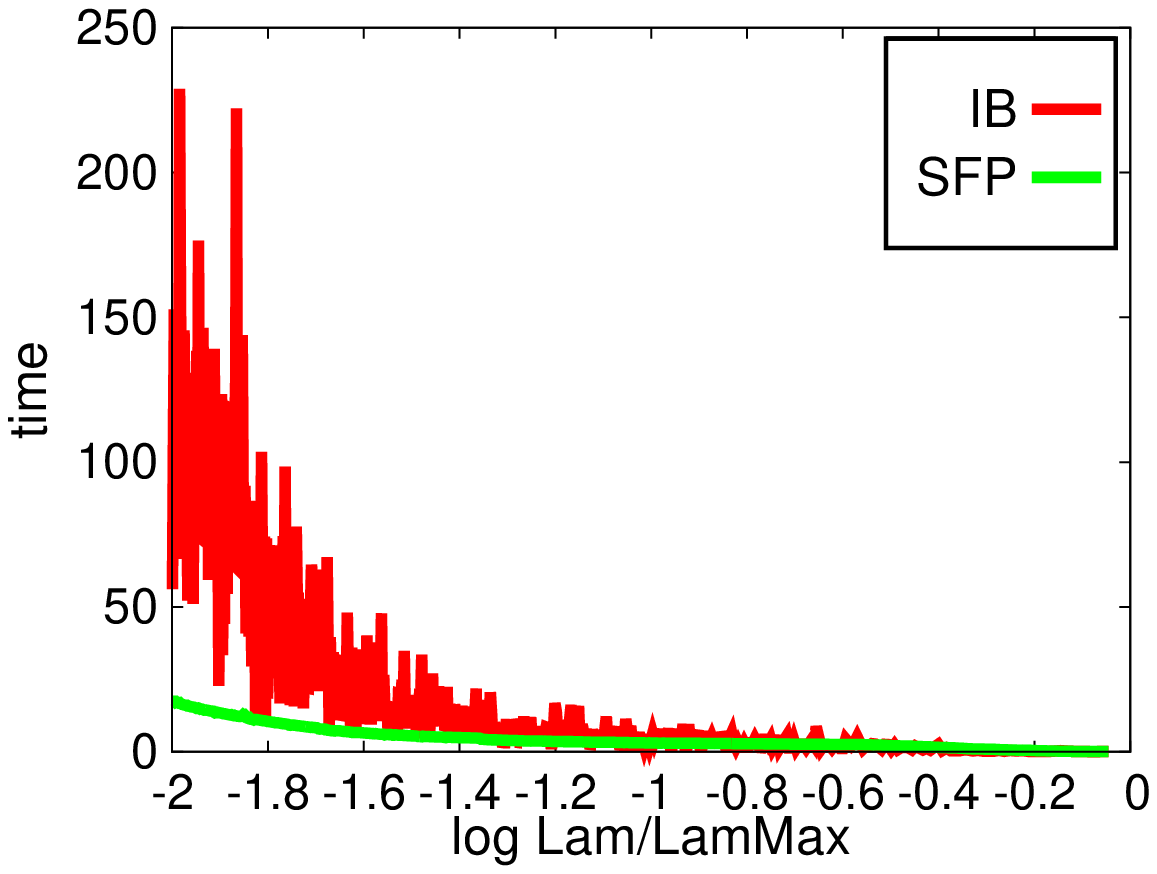}
\\
\multicolumn{2}{c}{(a) Computation time in seconds}
\\
\includegraphics[scale=0.5]{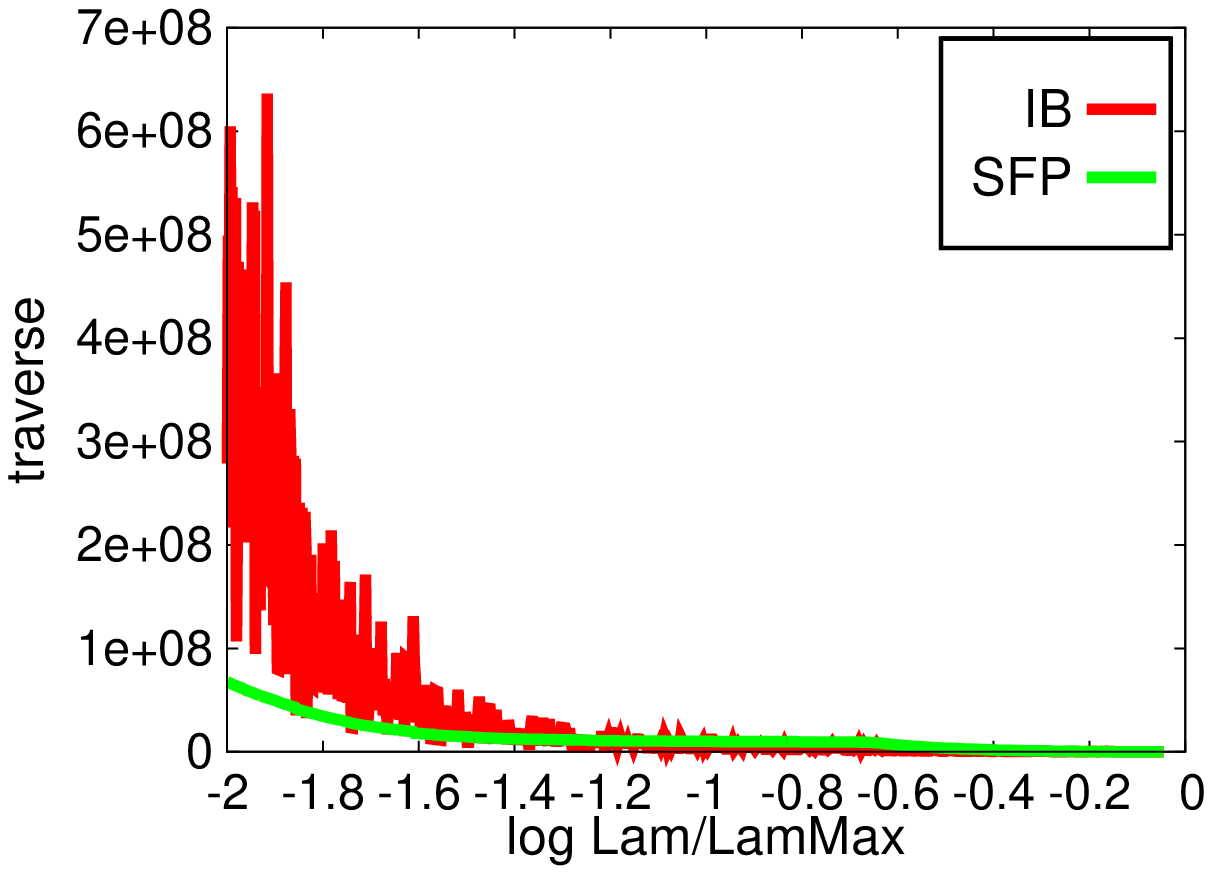}
&
\includegraphics[scale=0.5]{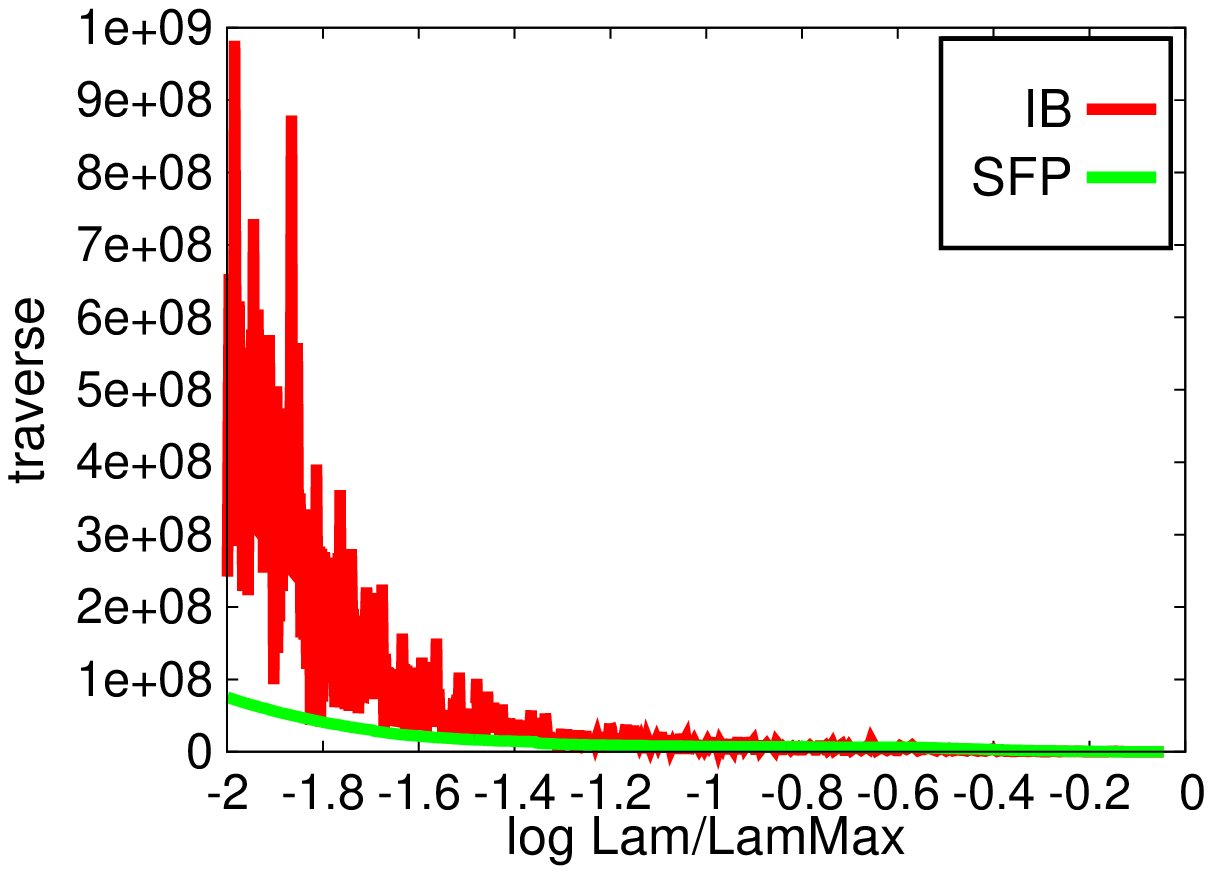}
\\
\multicolumn{2}{c}{(b) The number of traverse nodes}
\\
\includegraphics[scale=0.5]{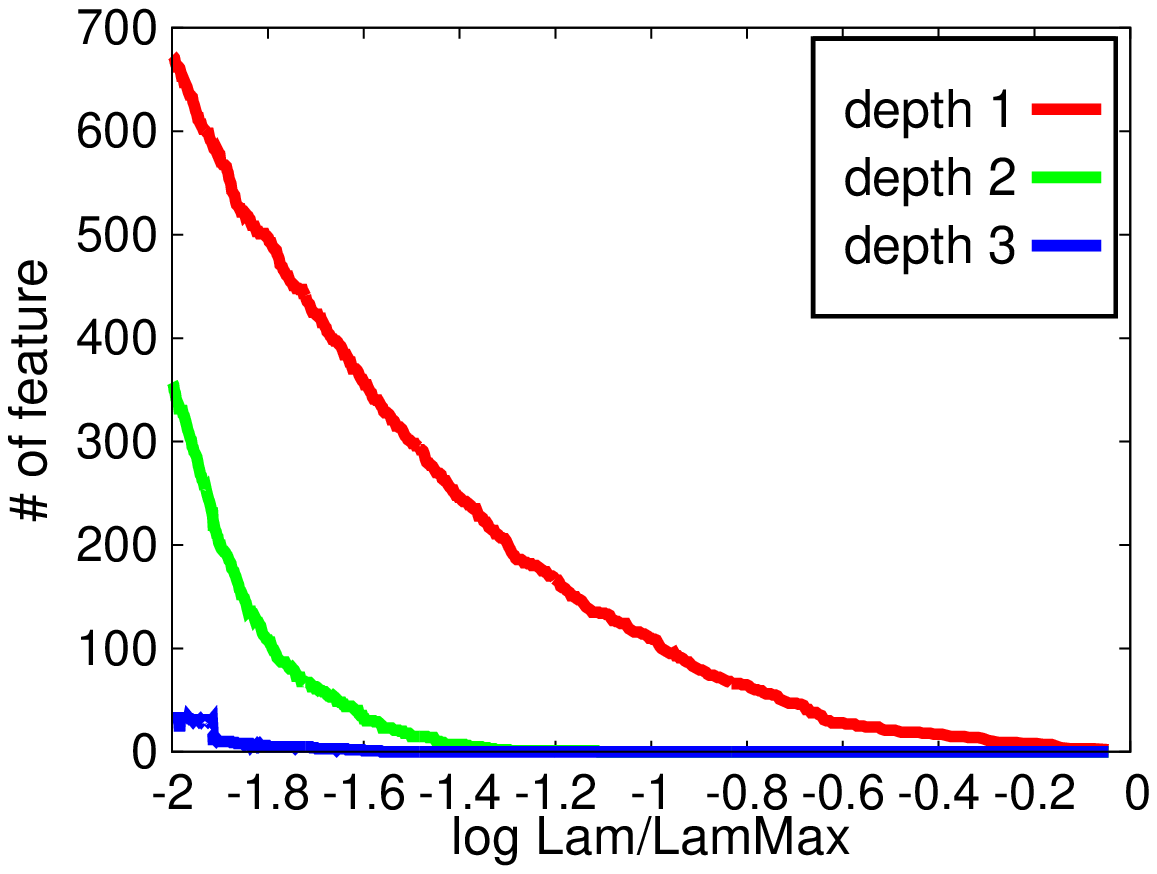}
&
\includegraphics[scale=0.5]{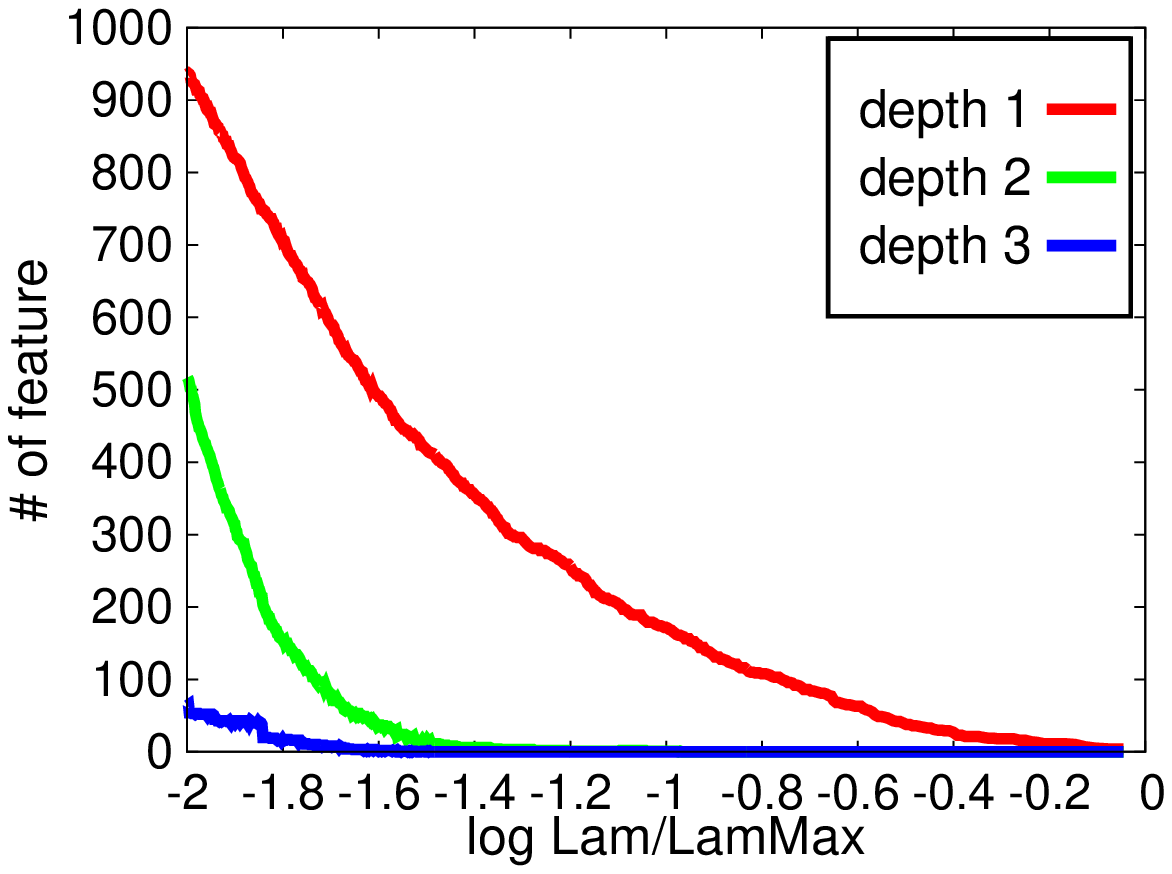}
\\
\multicolumn{2}{c}{(c) The number of active features}
\end{tabular}
\end{center}
\end{figure}

\begin{figure}[ht]
\begin{center}
\begin{tabular}{cc}
\includegraphics[scale=0.5]{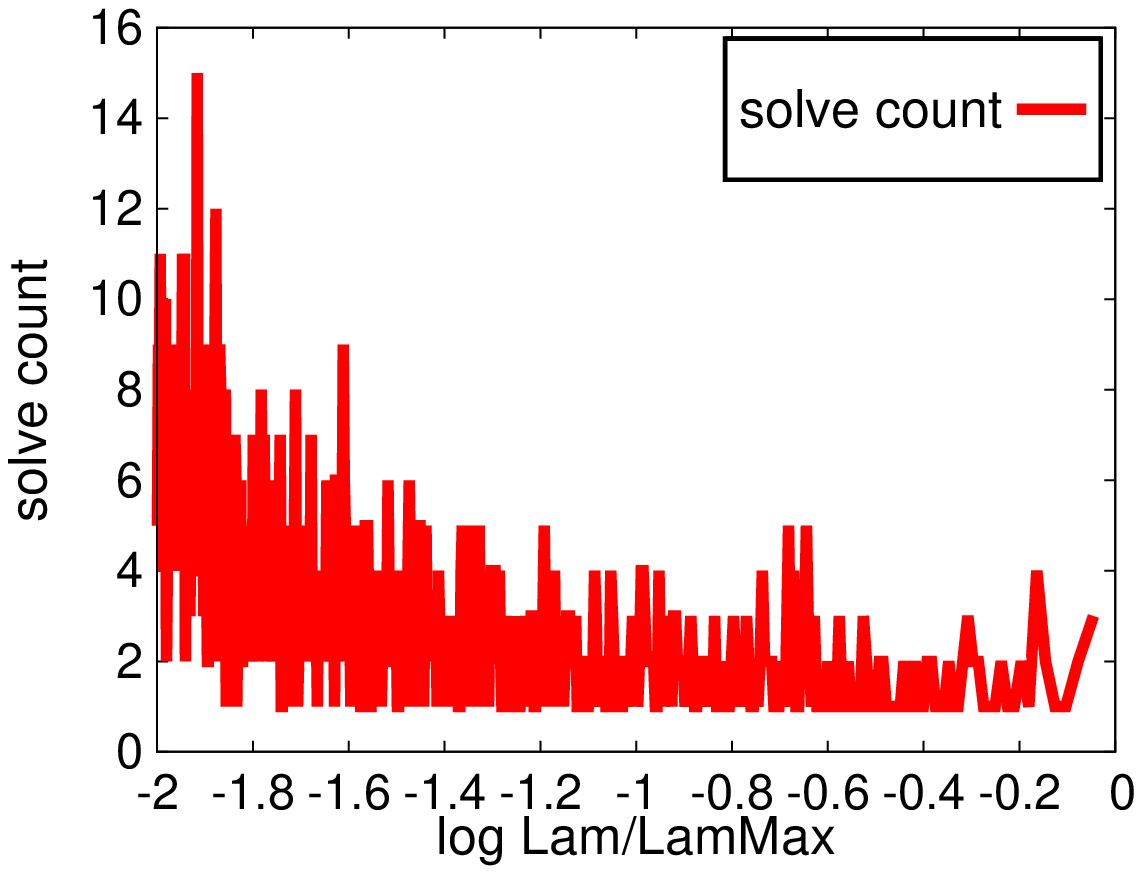}
&
\includegraphics[scale=0.5]{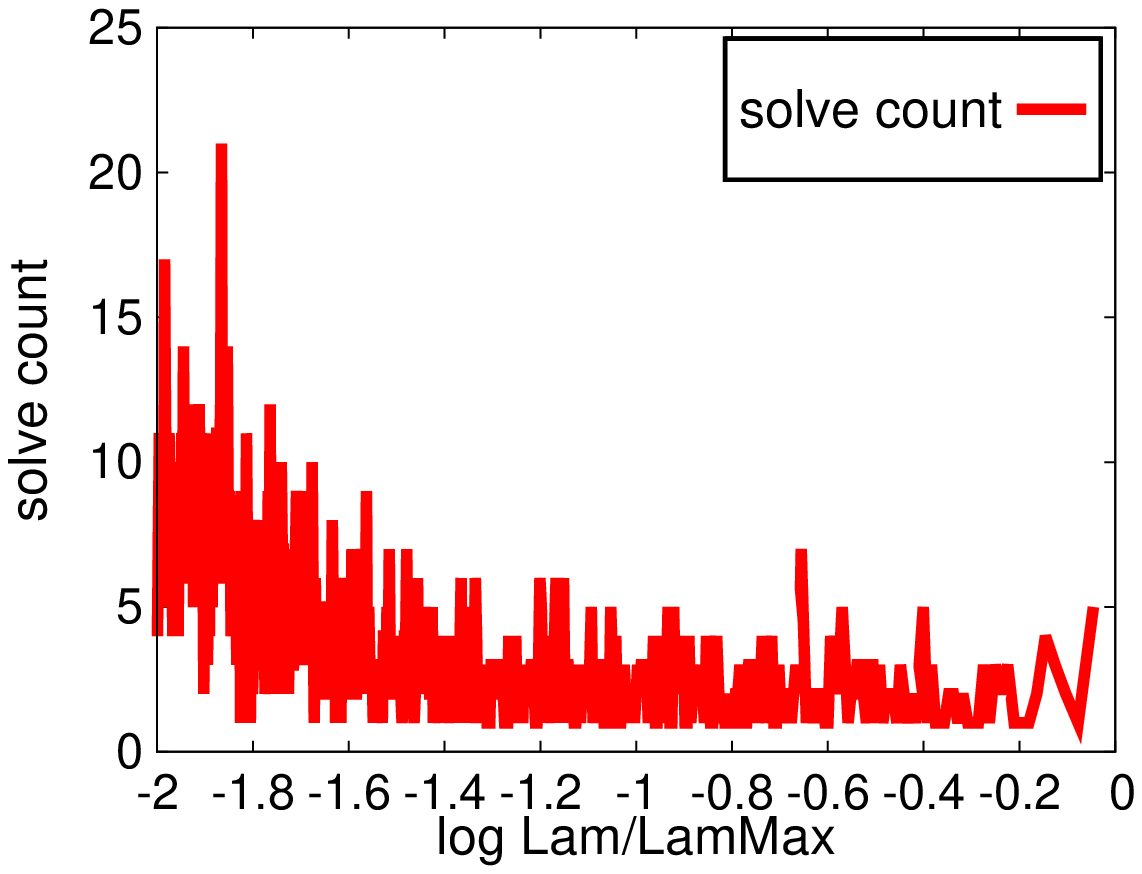}
\\
\multicolumn{2}{c}{(d) The number of solving LASSO in IB}
\\
\includegraphics[scale=0.5]{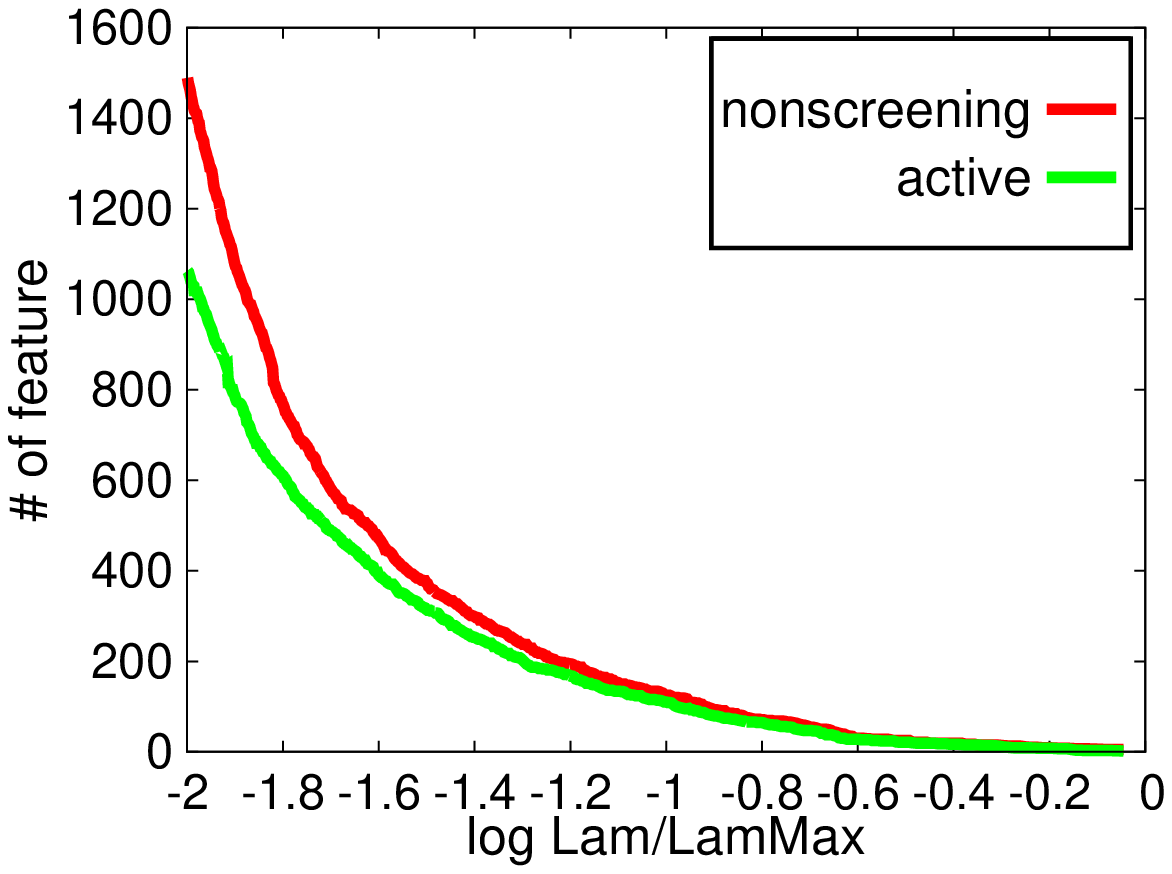}
&
\includegraphics[scale=0.5]{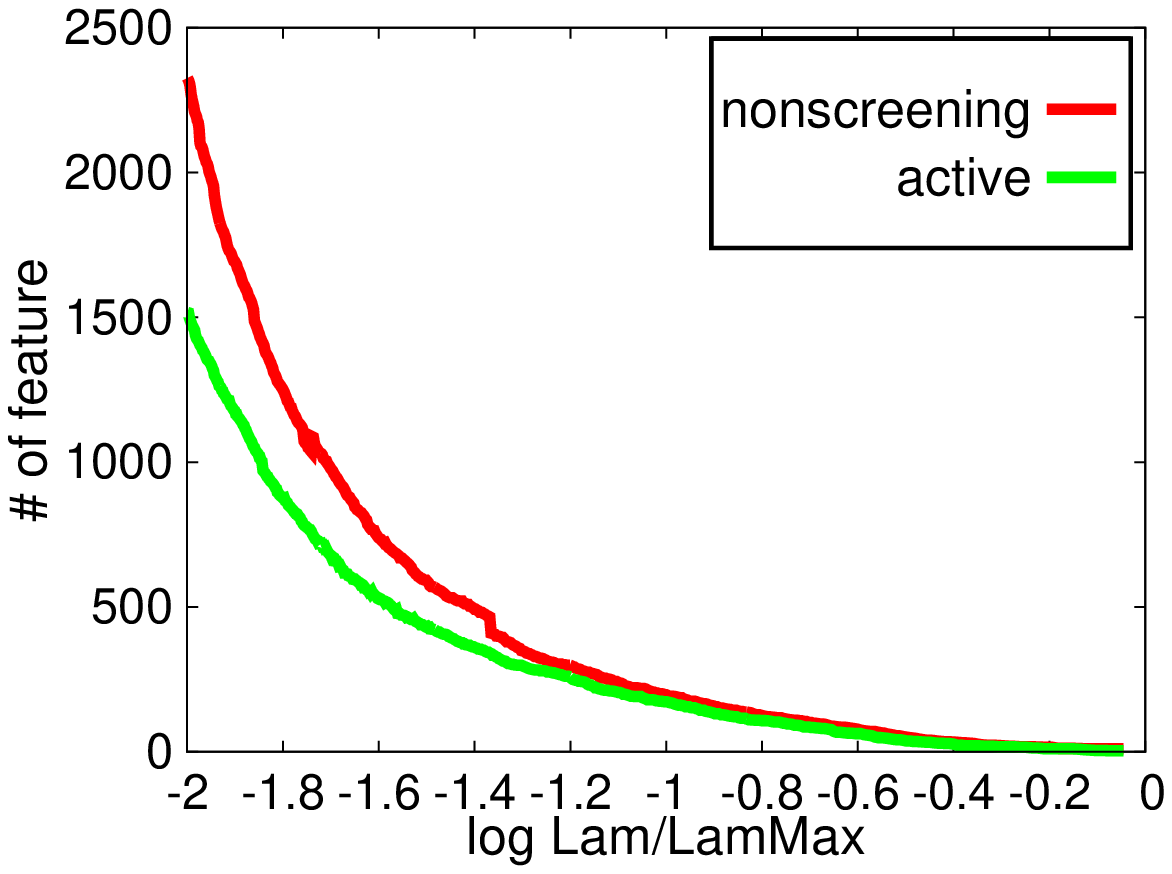}
\\
\multicolumn{2}{c}{(e) The number of non-screened out features and total active features}
\\
\includegraphics[scale=0.5]{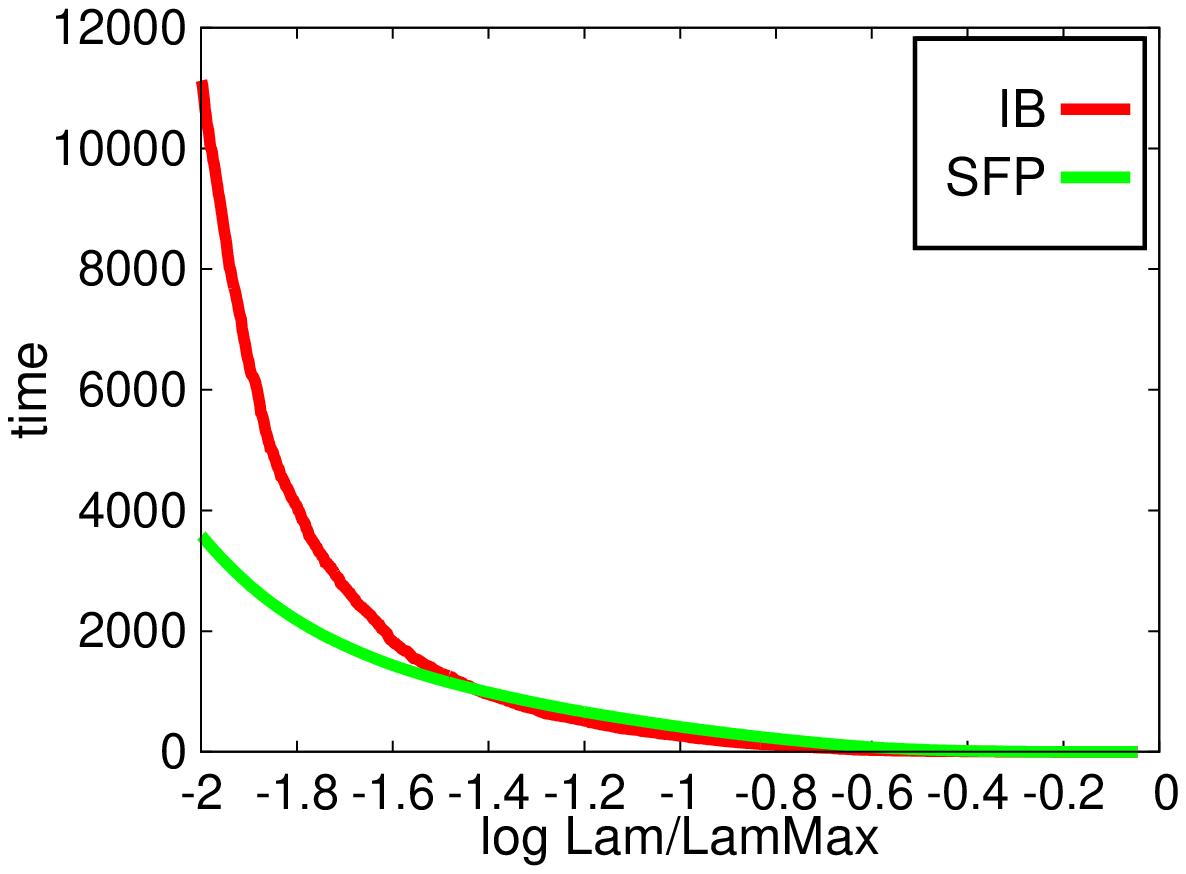}
&
\includegraphics[scale=0.5]{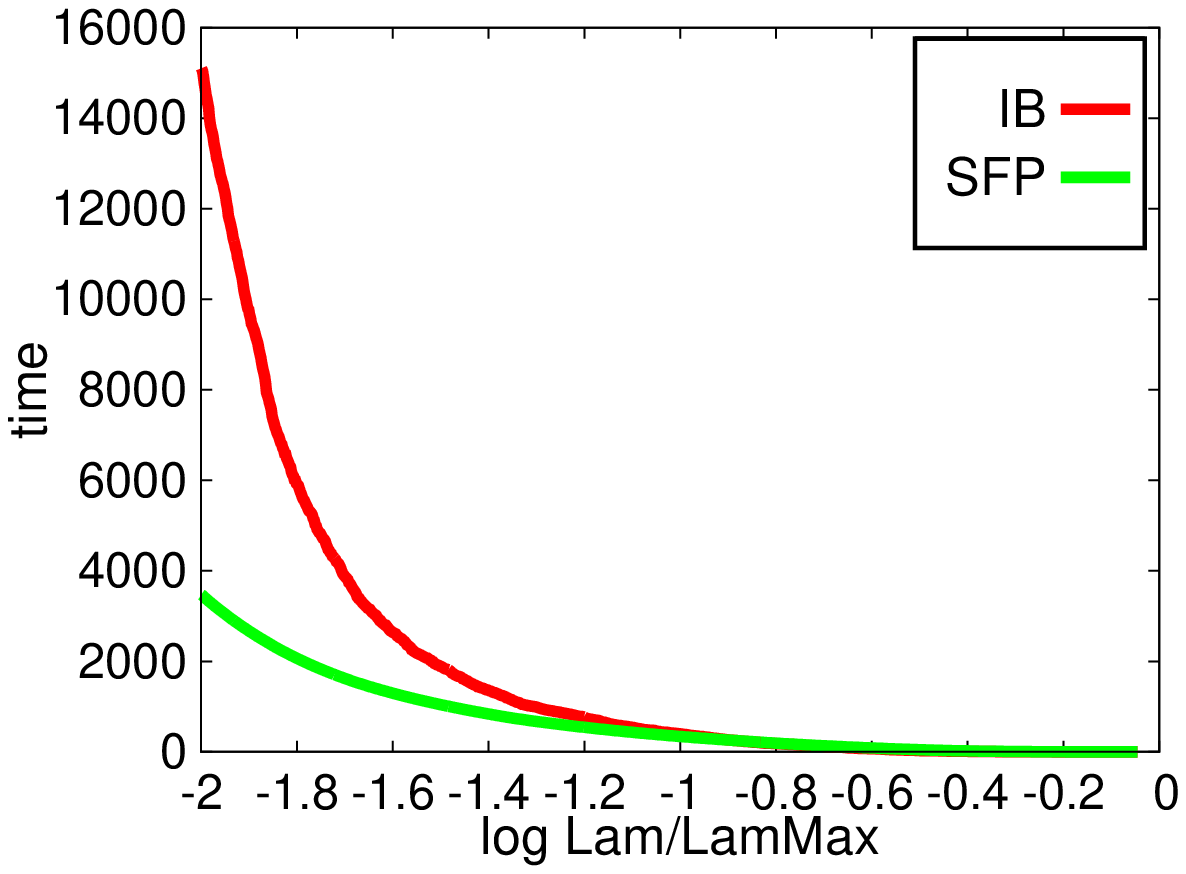}
\\
\multicolumn{2}{c}{(f) Computation total time in seconds}
\end{tabular}
\end{center}
\end{figure}

\clearpage

\subsection{Results on news20}
\begin{figure}[ht]
\begin{center}
\begin{tabular}{cc}
\includegraphics[scale=0.5]{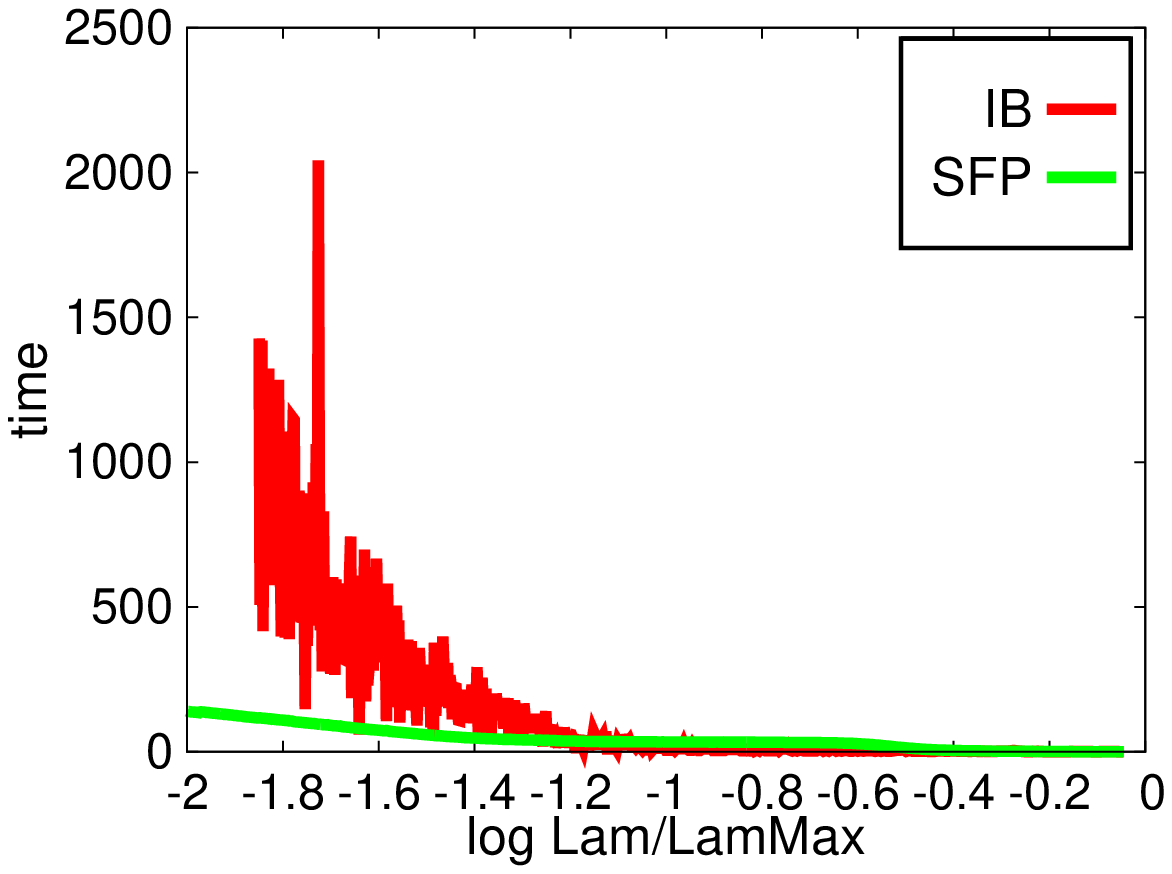}
&
\includegraphics[scale=0.5]{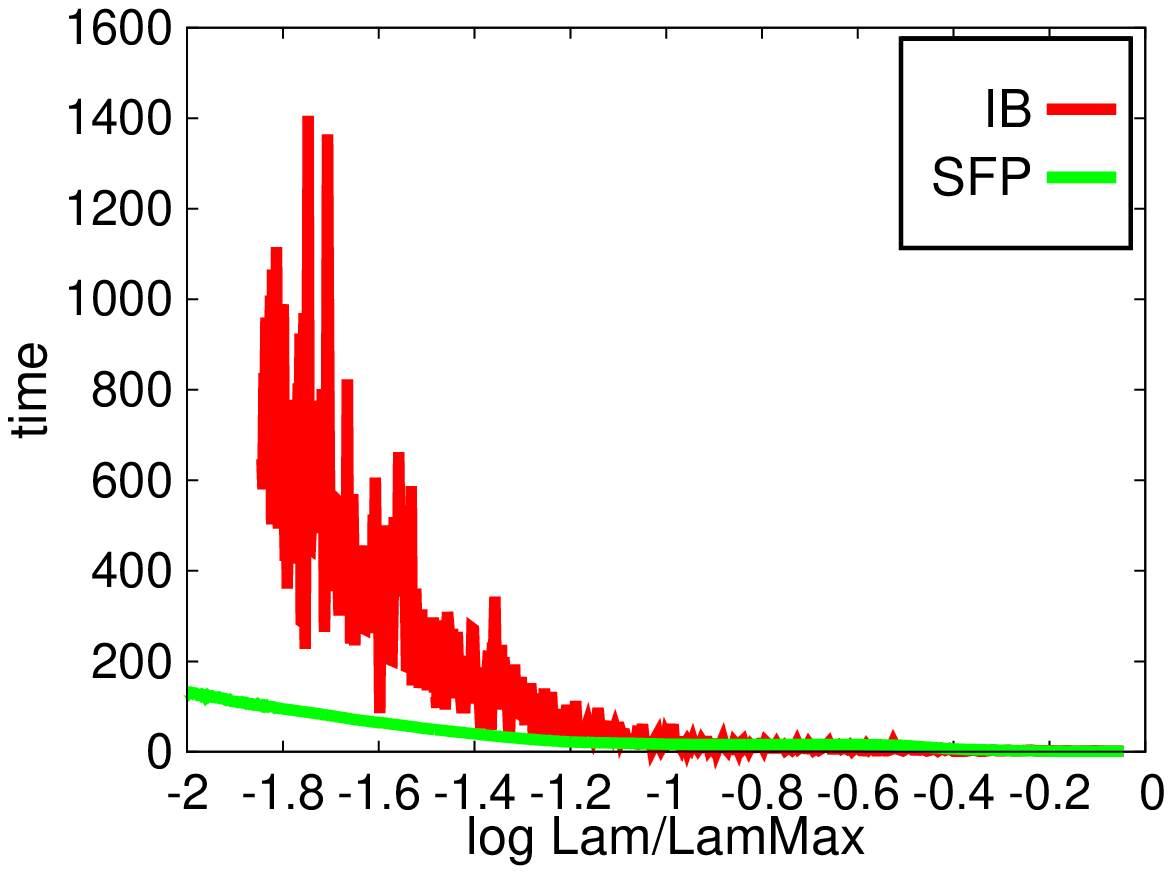}
\\
\multicolumn{2}{c}{(a) Computation time in seconds}
\\
\includegraphics[scale=0.5]{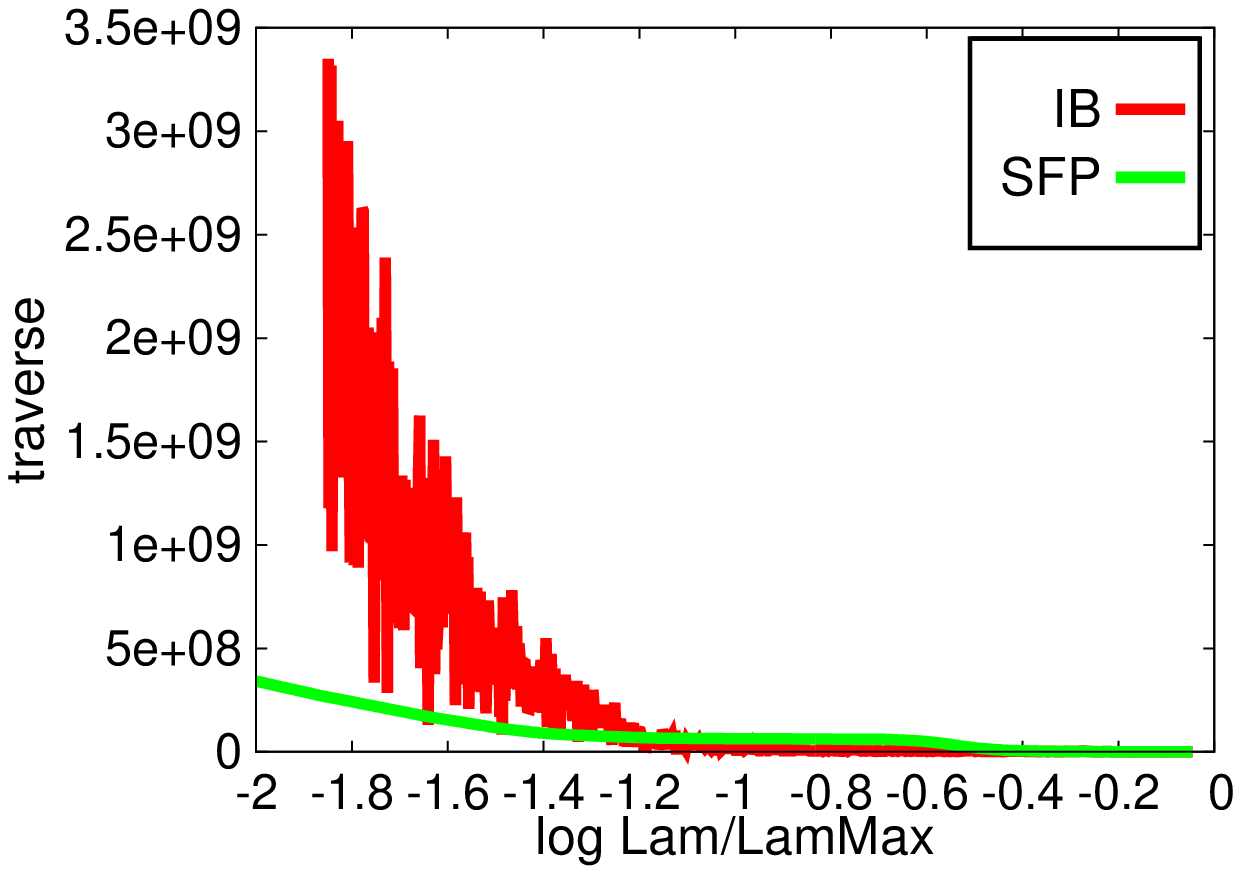}
&
\includegraphics[scale=0.5]{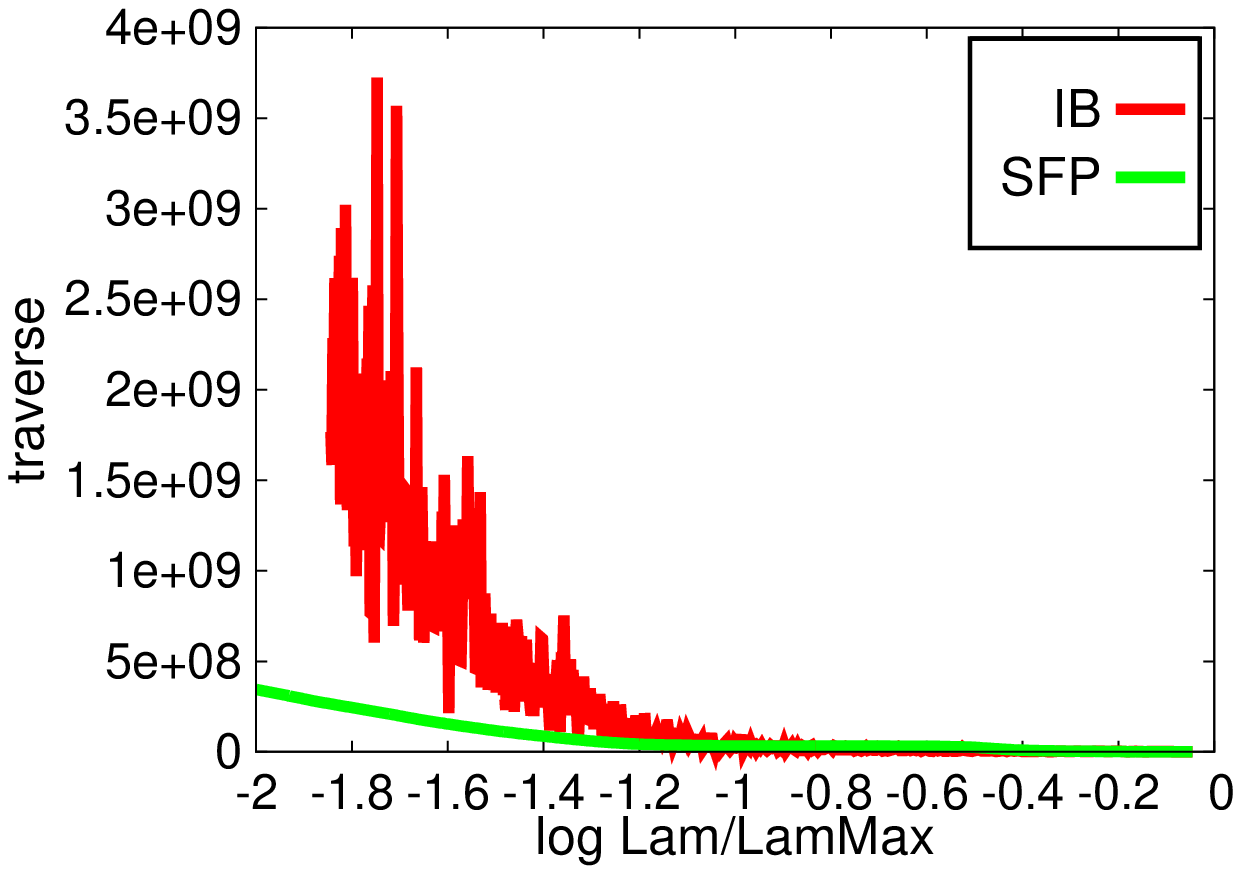}
\\
\multicolumn{2}{c}{(b) The number of traverse nodes}
\\
\includegraphics[scale=0.5]{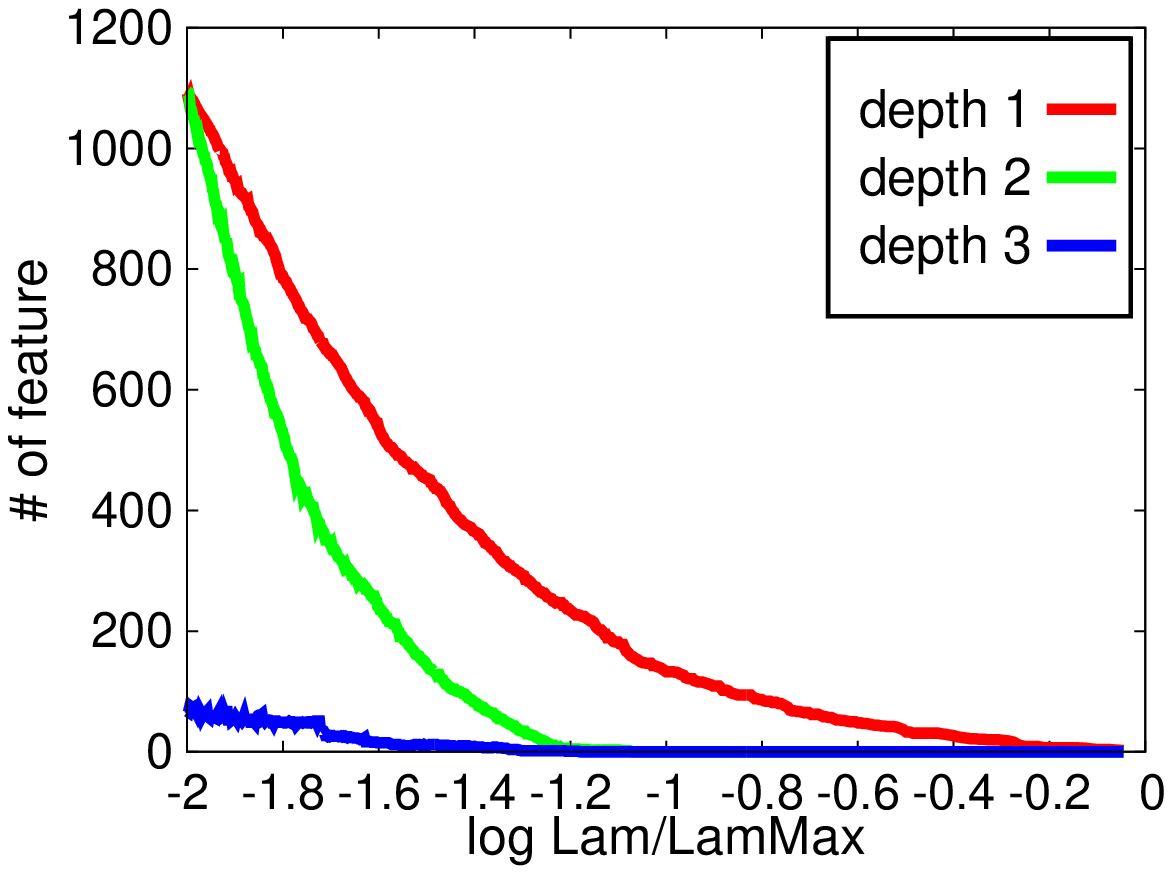}
&
\includegraphics[scale=0.5]{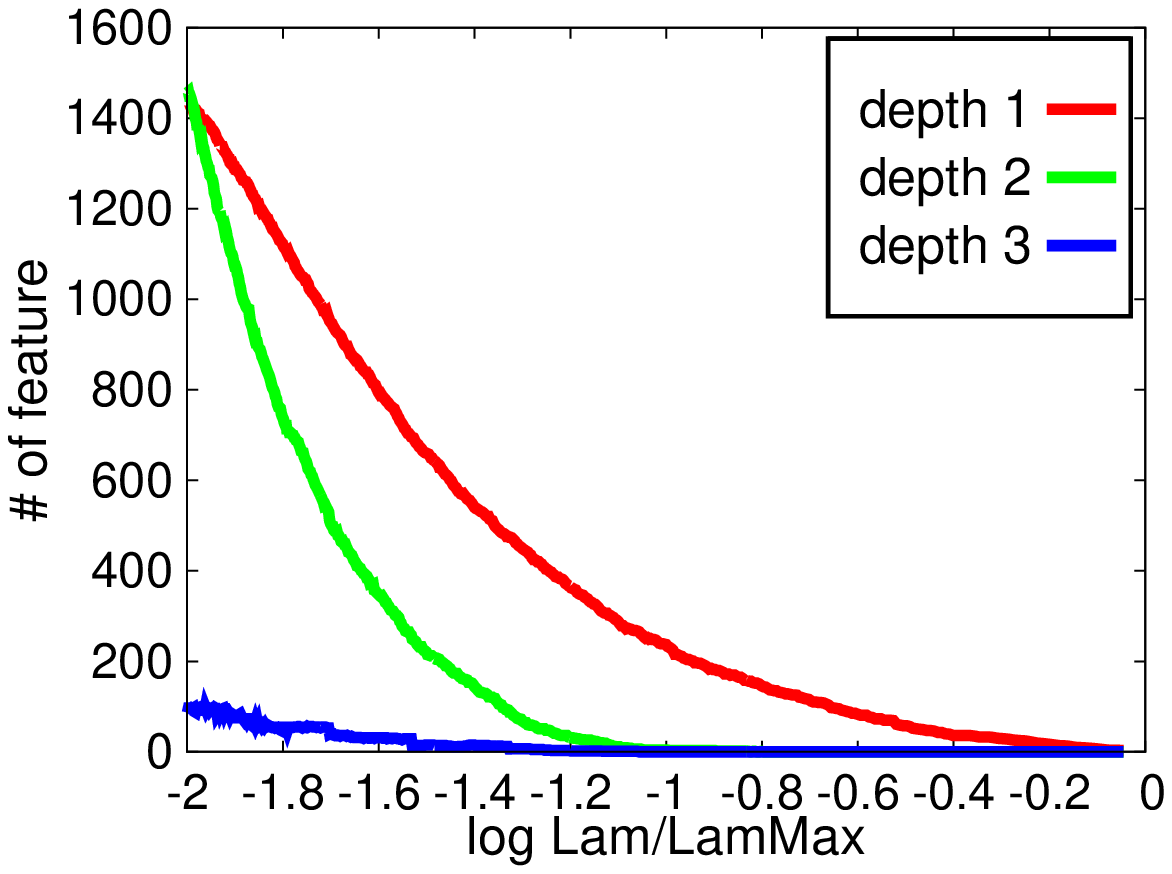}
\\
\multicolumn{2}{c}{(c) The number of active features}
\end{tabular}
\end{center}
\end{figure}

\begin{figure}[ht]
\begin{center}
\begin{tabular}{cc}
\includegraphics[scale=0.5]{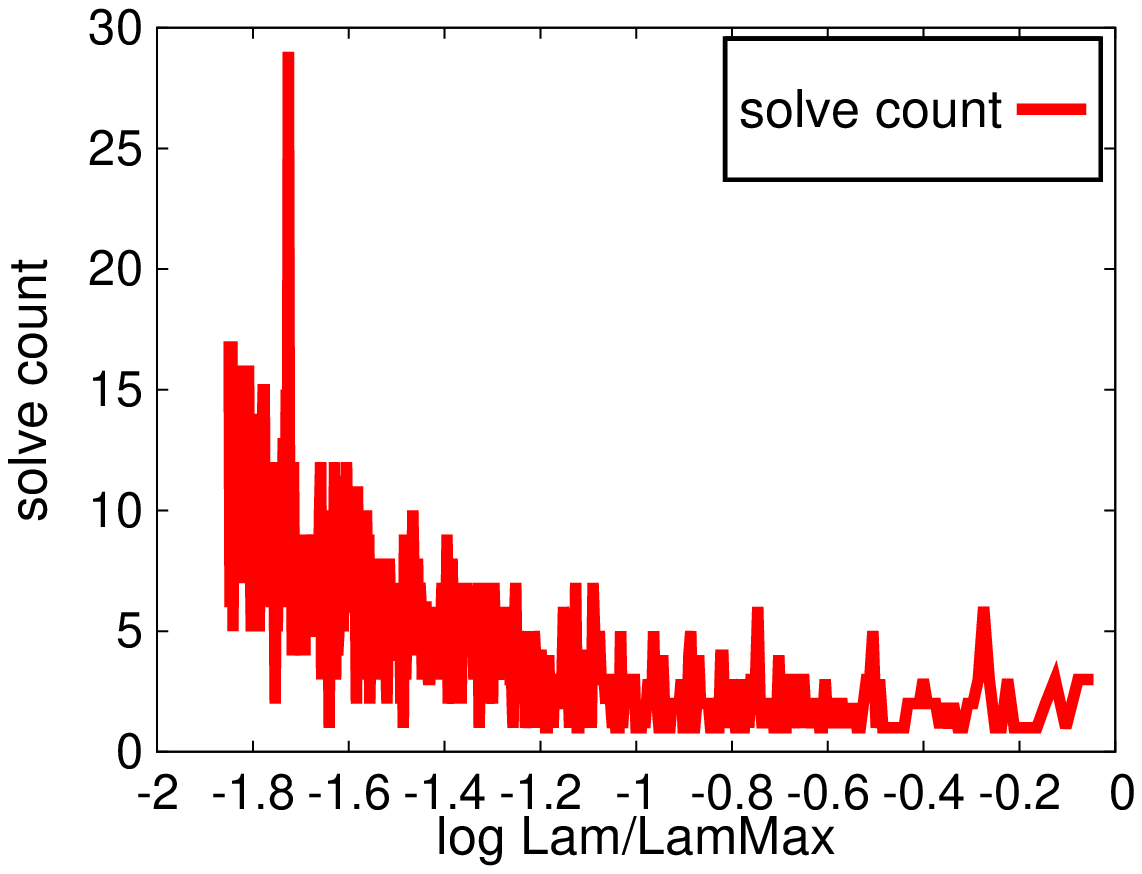}
&
\includegraphics[scale=0.5]{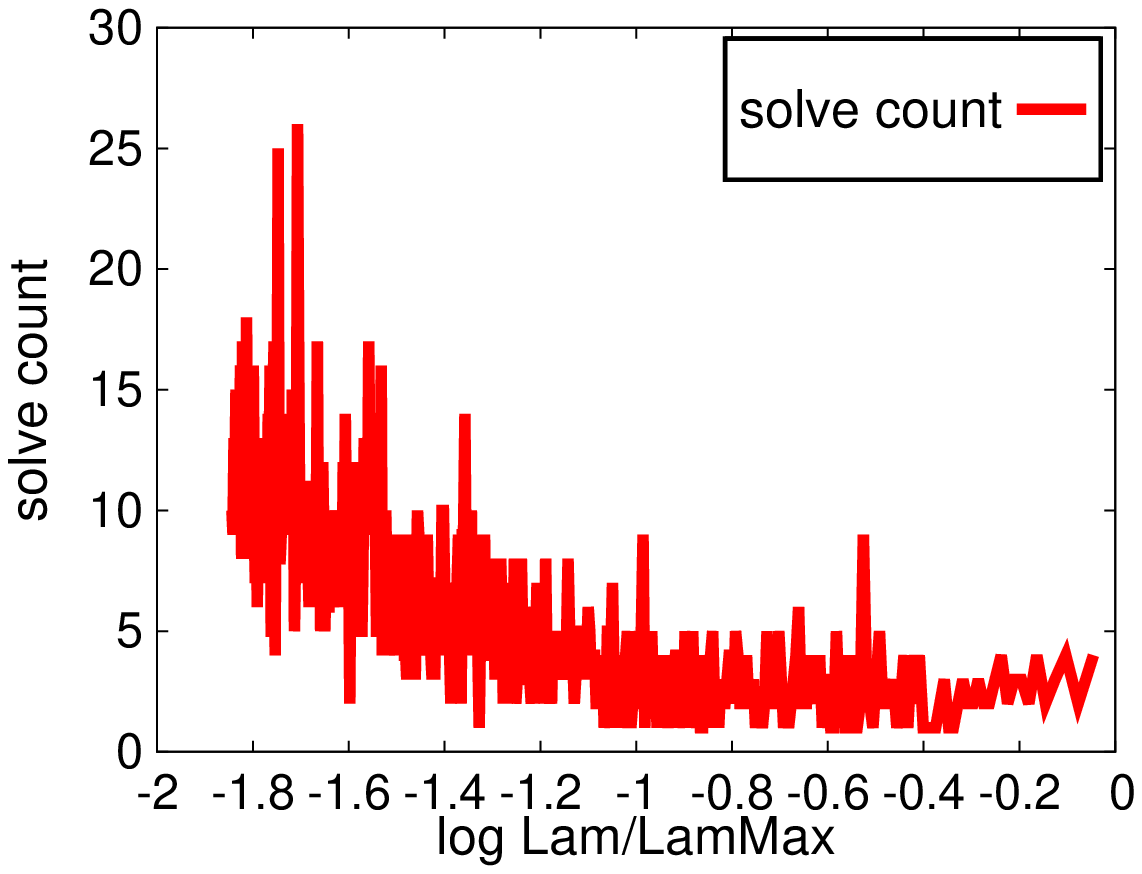}
\\
\multicolumn{2}{c}{(d) The number of solving LASSO in IB}
\\
\includegraphics[scale=0.5]{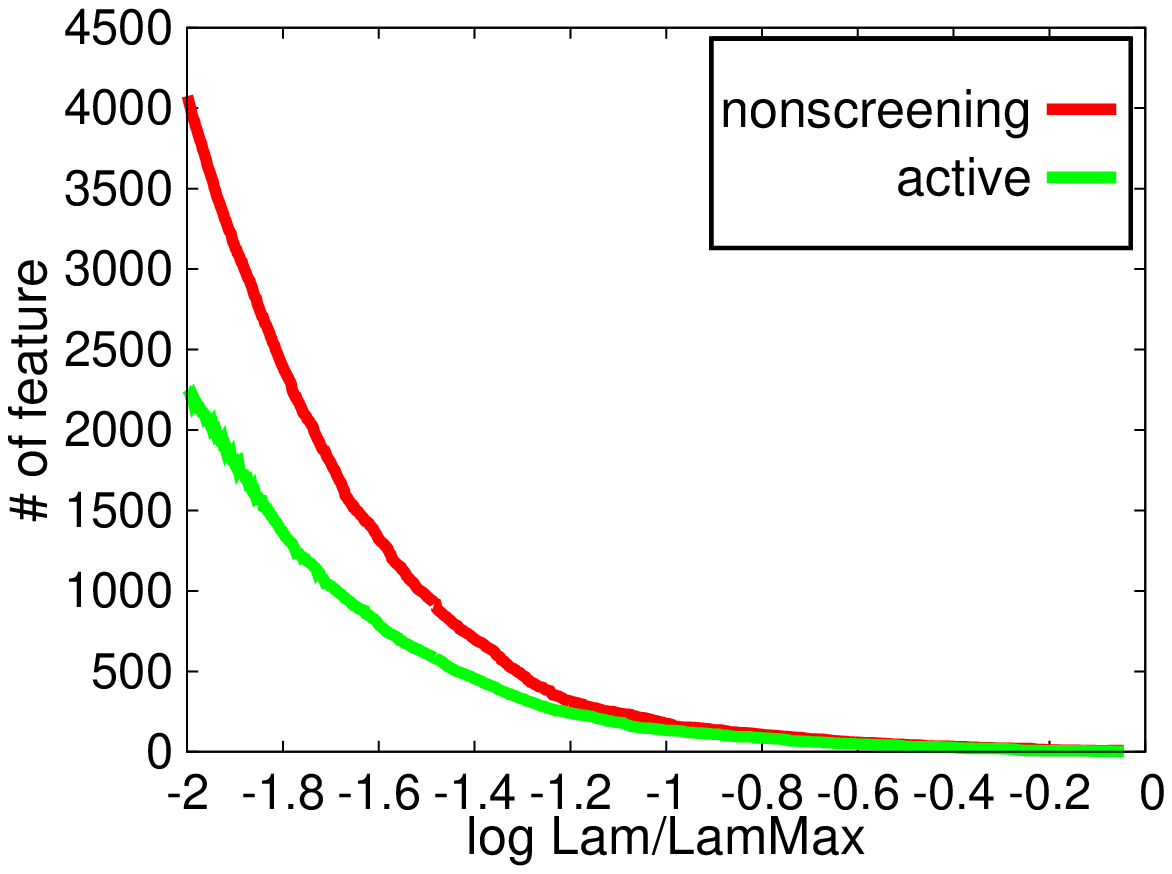}
&
\includegraphics[scale=0.5]{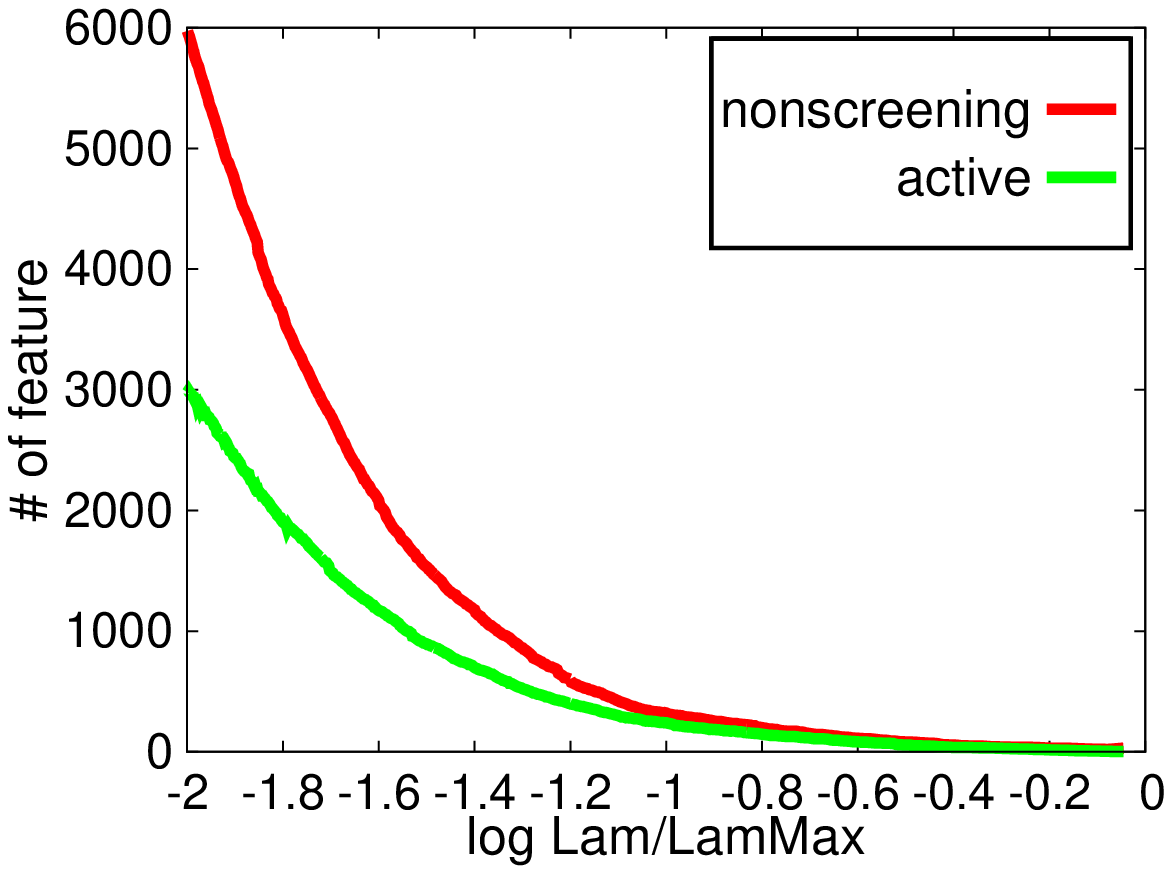}
\\
\multicolumn{2}{c}{(e) The number of non-screened out features and total active features}
\\
\includegraphics[scale=0.5]{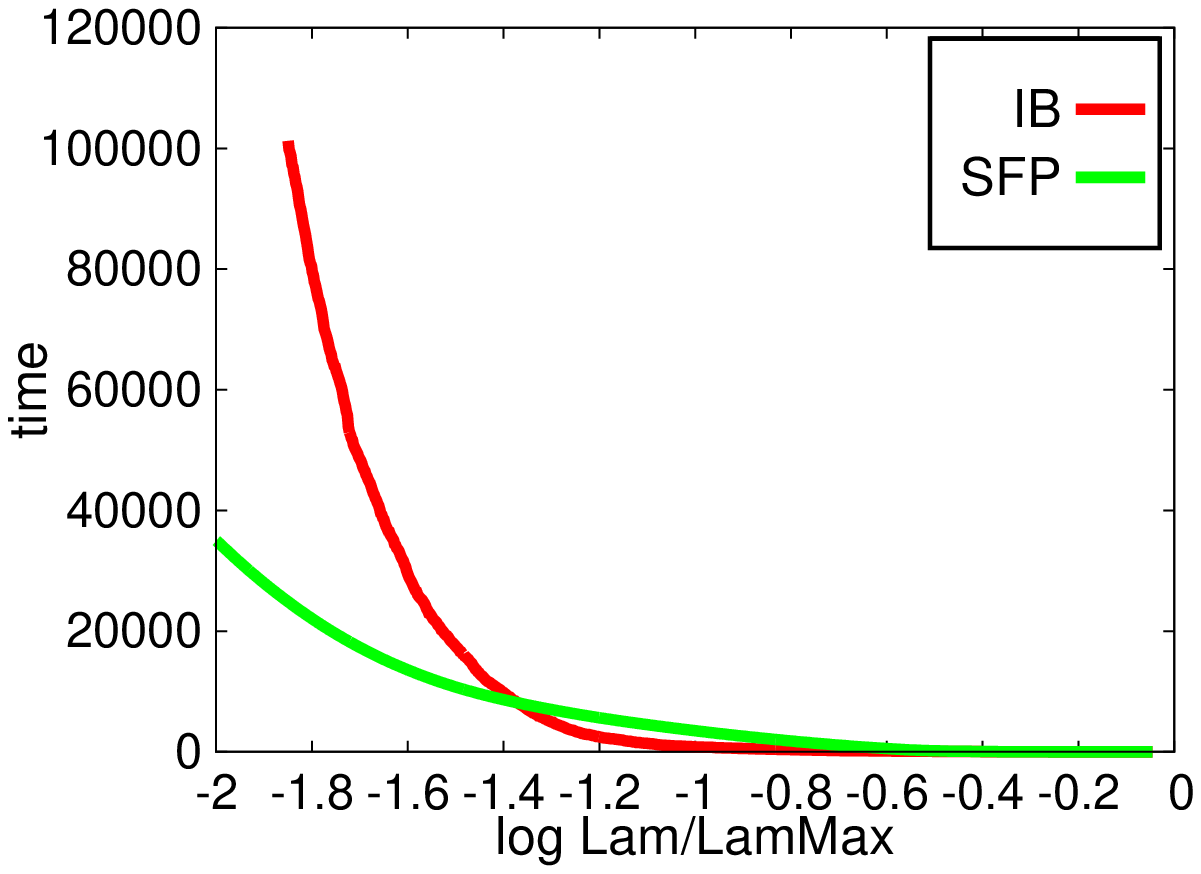}
&
\includegraphics[scale=0.5]{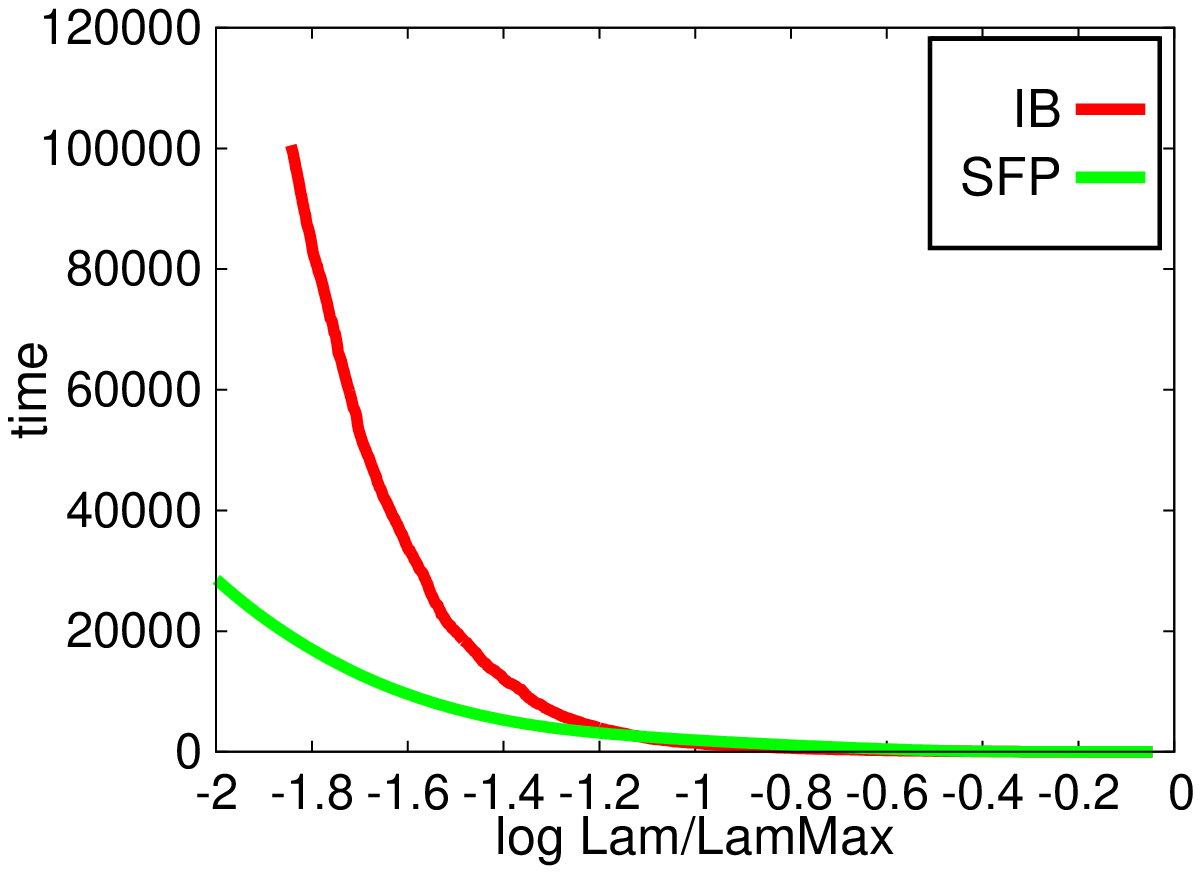}
\\
\multicolumn{2}{c}{(f) Computation total time in seconds}
\end{tabular}
\end{center}
\end{figure}